\documentclass[onecolumn,11pt]{IEEEtran}

\usepackage{fullpage}
\usepackage{amsmath,graphicx,amssymb,amsthm,enumerate,url}
\usepackage{graphicx,bm,float,bbm,caption}
\usepackage{subfig,float}
\usepackage{algorithm,algorithmic,stmaryrd}
\usepackage{mathtools}
\usepackage{xtab}

\def\H{{\cal H}}
\def\argmin{\mathop{\!\arg \min}}
\DeclareMathOperator*{\proj}{{\rm{proj}}}

\def\reals{\mathbb{R}}

\def\deq{{\triangleq}}
\usepackage[usenames]{color}
\definecolor{plum}  {rgb}{.4,0,.4}
\definecolor{forest}  {rgb}{0,.6,0}
\definecolor{midnight}  {rgb}{0,0,.8}
\usepackage[pdftex,plainpages=false,pdfpagelabels,colorlinks=true,urlcolor=midnight,linkcolor=midnight,citecolor=forest]{hyperref}

\newcommand{\trans}{^{\top}}

\newcommand{\tPhi}{\widetilde{\Phi}}

\def\Int{\operatorname{Int}}

\newcommand{\zeros}{\bm{0}}

\def\htheta{{\widehat{\theta}}}
\def\ttheta{{\widetilde{\theta}}}

\def\THETA{\boldsymbol{\theta}}

\def\argmin{\mathop{\!\arg \min}}

\def\deq{{\triangleq}}
\def\grad{\nabla}

\def\reals{\mathbb{R}}

\def\diag{\operatorname{Diag}}

\def\ones{\boldsymbol{1}}

\def\blambda{\boldsymbol{\lambda}}
\def\hlambda{\hat{\lambda}}
\def\tlambda{\tilde{\lambda}}

\def\bx{\bm{x}}

\newcommand{\wh}[1]{\widehat{#1}}
\newcommand{\wt}[1]{\widetilde{#1}}

\newcommand{\ave}[1]{\langle #1 \rangle}

\def\cX{{\mathsf X}}

\newcommand{\ie}[0]{\emph{i.e.,} }

\newcommand{\eg}[0]{\emph{e.g.,} }
\newcommand{\cf}[0]{\emph{cf.} }

\newcommand{\bseq}{\begin{subequations}\begin{align}}
\newcommand{\eseq}{\end{align}\end{subequations}}

\newtheorem{theorem}{Theorem}
\newtheorem{lemma}{Lemma}

\newtheorem{remark}{Remark}

\newtheorem{definition}{Definition}

\def\H{\mathcal{H}}

\title{Tracking Dynamic Point Processes on Networks}
\author{Eric~C.~Hall~and~Rebecca~M.~Willett
  \thanks{E. C. Hall is with the Department of Electrical and Computer
    Engineering, Duke University, Durham, NC, 27708, USA. e-mail:
    eric.hall87@gmail.com.  R. M. Willett is with the Department of
    Electrical and Computer Engineering, University of
    Wisconsin-Madison, Madison, WI 53706, USA. e-mail:
    willett@discovery.wisc.edu.}
 \thanks{This paper was presented in part at the 7th International IEEE EMBS 
 Neural Engineering Conference (2015) and appears in IEEE Transactions on Signal Processing,
 Vol 62, No. 7.}
  \thanks{We gratefully acknowledge the support of the awards
      AFOSR FA9550- 11-1-0028, NSF CCF-1418976, ARO 
      W911NF-09-1-0262, and AFOSR 14-AFOSR-1103.}}

\begin{document}
\maketitle
\begin{abstract}
  Cascading chains of events are a salient feature of many
  real-world social, biological, and financial networks. In social
  networks, social reciprocity accounts for retaliations in gang
  interactions, proxy wars in nation-state conflicts, or Internet
  memes shared via social media. Neuron spikes stimulate or inhibit
  spike activity in other neurons. Stock market shocks can trigger a
  contagion of volatility throughout a financial network. In these and
  other examples, only individual events associated with network nodes
  are observed, usually without knowledge of the underlying dynamic
  relationships between nodes. This paper addresses the challenge of tracking
  how events within such networks stimulate or influence future
  events. The proposed approach is an online learning framework
  well-suited to streaming data, using a multivariate Hawkes point process model to
  encapsulate autoregressive features of observed events within the
  social network. Recent work on online learning in dynamic
  environments is leveraged not only to exploit the dynamics within
  the underlying network, but also to track the network structure as
  it evolves. Regret bounds and experimental results demonstrate  that
  the proposed method performs nearly as well as an oracle or batch algorithm.
  \end{abstract}

\section{Introduction}
\label{sec:intro}
In a variety of settings, we observe a series of events associated
with a group of actors whose interactions trigger future events. The interactions
between these actors can be modeled using a network.
For example:
\begin{itemize}
\item {\bf Social networks:} we observe events such as people meeting,
  corresponding, voting, or sharing information
  \cite{raginsky_OCP,silva:pami,BertozziHawkes,HellerHawkes,zhouZhaSongHawkes};
\item {\bf Biological neural networks:} spiking action potentials can
  trigger or inhibit spikes in neighboring neurons according to
  time-varying functional networks
  \cite{brown2004multiple,colemanConvexPoint,SmithBrownStateSpace,HinneHeskes2012,
    DingSchroeder2011,spikesPillow,spikesMasud}.
\item {\bf Financial networks:} stock market shocks can
  trigger jumps across the global network of financial instruments
  and indices\cite{ait2010modeling, cameron2013regression, engle2001garch};
\item {\bf Epidemiological networks:} as a contagion spreads through a
  community, observations of symptoms in one person are strong
  predictors of future symptoms among that person's neighbors
\cite{kuperman2001small}; and
\item {\bf Seismological networks:} substantial seismic activity is
  often predicated by foreshocks and followed by aftershocks, with the
  epicenter of these shock events determined by the local geography and
  plate tectonics \cite{hawkesEarthquake,ogata1999seismicity, Schoenberg13}.
\end{itemize}

In all the above settings, the interactions between actors are
critical to a fundamental understanding of the underlying network
structure and accurate predictions of likely future events.

We can model these interactions between nodes using a network or
graph, where directed edge weights correspond to the degree to which
one node's activity stimulates activity in another node. For instance,
the network structure may indicate who is influencing whom within a
social network, or the connectivity of neurons. In these
and other contexts the underlying network structure may be changing over time, for instance as
people's relationships evolve or as a function of the activity in which
the brain is engaged. In many cases, we are interested in both the
rates at which different nodes or actors participate in events and
the underlying network structure.
 
Our goal is to filter and track such processes. We present methods and
associated theoretical performance bounds in two settings: (a) where
the underlying network structure is known and (b) where the underlying
network structure is unknown.  Our approach incorporates concepts and
tools from multivariate Hawkes models of point processes
\cite{hawkes1,hawkes2,PointProcesses} and online convex optimization
methods for dynamic environments
\cite{DMD_journal,dynamicMirrorDescent}. In particular, the
multivariate Hawkes process is akin to an autoregressive model for
point processes, where events up to time $t$ dictate the rate at which
events are anticipated after time $t$. 

Estimating the parameters associated with these processes is the
subject of much current research, but existing methods typically
assume that the underlying network parameters are static rather than
changing with time, and require computationally-prohibitive batch
processing algorithms. Specifically, there has been
  substantial work in estimating parameters of the system through
  methods which seek to estimate both the ``parent event" of each
  event, and then use this information to learn the parameters of the
  influence function and/or the network \cite{SimmaJordan10} using
  either EM-type algorithms or Bayesian techniques \cite{Schoenberg13, Mohler13,
    SornetteUtkin09, VeenSchoenberg08,
    ryanAdamsHawkes}. The difficulty with using this approach in the
  online setting is that in order to accurately estimate parent
  events, we need a potentially large buffer of stored event
  times, and do processing that scales poorly with the number of
  previously observed events.  The work closest to ours is
  \cite{LindermanAdams15}, which uses a Bayesian framework to learn the
  parameters of a discretized version of the Hawkes process, which is
  computationally more efficient with regards to the number of events
  observed. However, they still require at least mini-batches and
  having access to data more than once, which is not a truly streaming
  setting. Additionally, all of these methods require the data to
  exactly follow the Hawkes model, and have no guarantees for
  performance in the case of misspecified influence functions or
  generative model, whereas our results both theoretically and
  empirically offer protection against such model mismatch.

In the the cascading point process described above, we face several key challenges:
\begin{itemize}
\item[(a)] the
underlying networks are dynamic, 
\item[(b)] we receive either a large volume of data
or data that is streaming at a high velocity, necessitating sequential
processing, and 
\item[(c)] we seek performance guarantees that are robust to
model mismatch (i.e. perform well even when the data was not truly
generated by the Hawkes model).
\end{itemize}
Our proposed method will simultaneously track the time-varying rates
at which events are expected and the underlying time-varying network
structure. In contrast, most methods assume that the rates are a
known, closed-form function of the observed data and the network
structure, and focus solely on estimating the network; we will see
that this approach is more fragile with respect to modeling errors. 
Additionally, due to the streaming nature of our algorithms, we are in a regime 
where we can easily do forecasting, which is valuable for the financial, epidemiological,
seismological and other networks. Our algorithms create an estimate of the 
process rates at every time before seeing the actual events at that time. Therefore, the online
framework allows us to do one step ahead forecasting. This would be much harder to do
using previous methods which learn networks in Hawkes processes, because to do prediction all the 
previous data would have to be processed, which is computationally intensive, and then projected forward
to new time points. These methods would either have to be run at every time point, which is computationally infeasible, or would only be run a few times and not be able to track short term changes in the network.

The remainder of the paper is organized as follows: Section
  \ref{sec:prob} introduces the basics of the Hawkes process and a
  mathematical description of our learning objective. Section
  \ref{sec:online} covers some of the basics for online learning with
  dynamics in a general setting for generic loss functions.  Section
  \ref{sec:loss} then describes the time discretized loss function and
  dynamical model which corresponds to the Hawkes process which are
  needed for our online learning framework. Section
  \ref{sec:algorithms} introduces our two proposed online algorithms
  for tracking these processes, one which assumes the network is known
  and one which attempts to learn both the time varying-rates as well
  as the network simultaneously. Section \ref{sec:computation} has a
  brief discussion on the computational complexity of the
  methods. Finally Section \ref{sec:experiments} shows how our methods
  perform in practice on synthetic data both when the generative model
  is known and when it is misspecified, as well as experiments
  performed on the Memetracker data set. Proofs and a notation legend
  are placed in the appendix.
\section{Problem Formulation}
\label{sec:prob}
We monitor $p$ actors in a network, and record the identities of the
actor and time of each event. An actor and event may represent a person
``liking'' a photo or article shared by another person in a social
network, a neuron firing in the brain, or the incidence of disease.
That is, we observe a time series of the form $\{(k_n,\tau_n)\}_n$,
where $k_n \in \{1,2,...,p\}$ is the actor involved in the $n^{\rm th}$ event
and $\tau_n \in \reals_+$ is the time at which it occurs. With each
new event, we wish to accurately predict which future events are most
likely in the immediate future and the underlying network of
influence. We define $\tau_0 \deq 0$.

We wish to track the time-varying likelihood of each
of the $p$ actors acting. To do so, we adopt a multivariate Hawkes process
model \cite{hawkes1,hawkes2,PointProcesses} and track the parameters
of this model over time. In particular, for each actor $k$ we have a
point process with time-varying rate function $\mu_k(\tau)$. Let
$N_{k,\tau}$ denote the number of recorded events for actor $k$ up
to and including time $\tau$, and let $N_\tau \deq \sum_{k=1}^p
N_{k,\tau}$ denote the total number of events (across all actors) up
to and including time $\tau$. The likelihood of actor $k$
participating in an event between times $t_1$ and $t_2$ is controlled by
the integral $\int_{\tau_1}^{\tau_2} \mu_{k}(\tau) d\tau$. More
formally, the collection of all observed events up to time $T$ can be
denoted
$$\mathcal{H}^T \deq \{N_{k,\tau}\}_{\substack{k \in \{1,...,p\} \\ \tau \in (0,T]}}.$$
The log likelihood of 
observing $\mathcal{H}^T$ given the $p$ rate functions $\mu_k(t)$ for $k\in\{1,..,,p\}$ is then \cite{patriciaHawkes}
\begin{align}
\log p(\mathcal{H}^T|\mu) = &\sum_{n=1}^{N_T} \log \mu_{k_n}(\tau_n) - \sum_{k=1}^p \int_{0}^T \mu_k(\tau)d\tau.
\label{eq:hawkesLogLike2}
\end{align}

Thus far, everything explained is common to a wide class of point processes. The multivariate Hawkes processes considered in this paper are essentially an autoregressive point process, where each rate function $\mu_k(\tau)$ depends on the history of past events, $H^\tau$. In particular, a multivariate Hawkes process assumes the rate functions can each be expressed as
\begin{equation}
\mu_k(\tau) = \bar{\mu}_k + \sum_{n=1}^{N_t} h_{k,k_n}(\tau-\tau_n).
\label{eq:hawkesdyn}
\end{equation}
Here $\bar{\mu}_k$ is a baseline rate representing the nonzero
likelihood of actor $k$ acting even without having been influenced by
any previous actions, with $\bar{\mu} \triangleq
[\bar{\mu}_1,...,\bar{\mu}_p]\trans$. Furthermore, we have $p^2$
functions of the form $h_{k_1,k_2}(\tau)$ which describe how events
associated with actor $k_2$ will impact the likelihood of events
associated with actor $k_1$.  In order to assure causality we assume
$h_{k_1,k_2}(\tau) = 0$ if $\tau \leq 0$ for all $k_1,k_2 \in
\{1,...,p\}$.  These functions depend on the underlying network
connectivity; if actors $k_1$ and $k_2$ are unconnected, the
corresponding function $h_{k_1,k_2}$ should be identically zero for
all $\tau$. 

One of the main challenges in statistical estimation for multivariate
Hawkes processes is the estimation of these $p^2$ functions. In
general, this problem is highly underdetermined and
challenging. Recent work has attempted to mitigate these challenges
using low-rank and sparse models
\cite{patriciaHawkes,hansen2012lasso,soloHawkesICASSP13}.  In this
paper, we make the common (\cf \cite{zhouZhaSongHawkes,expHawkesSimulation}) simplifying assumption that
these interactions all have the same functional form but different
(and often zero-valued) amplitudes, so that
\begin{equation}
h_{k_1,k_2}(\tau) = W_{k_1,k_2} h(\tau)
\label{eq:Wh}
\end{equation}
where $h(\tau)$ is known but the amplitude matrix $W =
[W_{k_1,k_2}]_{k_1,k_2 \in \{1,..,p\}}$ may be unknown. We will refer
to $h(\tau)$ as the {\em influence function}, as it depicts how an action's influence on an actor will vary in time. The matrix $W$ indicates
the strength of influence between actors; from a graph theory
perspective, $W$ acts like the weighted adjacency matrix of a graph
representation of the network.

{\bf Our goal is to obtain an estimate for $\mu(t)$ as it evolves and
  to infer $W$ {\em online} from streaming network data.} Furthermore,
we seek methods with performance guarantees that hold even when the
observed data is not generated strictly in accordance with the above
Hawkes model. That is, while we use the Hawkes model to measure how
well estimates fit the data, we recognize that the model will never be
perfectly accurate (\eg we may have errors in our estimate of the
influence function $h(\tau)$ or the linear model in \eqref{eq:hawkesdyn} may
not reflect nonlinearities present in real data) and wish performance
guarantees even in the face of these uncertainties. The proposed method is an
application of online optimization in dynamic environments, which requires a loss function and a dynamical model.
On the highest level, the method takes a current estimate of the rate and then slightly adjusts it based on the most recently observed data.
In classical online learning settings, this innovation step is based solely on gradient of the chosen loss function, which will be related to the negative log-likelihood of the Hawkes process. Our approach adds a second main step of the algorithm, which is to then update the rate by incorporating the Hawkes dynamical model that certain nodes in the system will stimulate actions from other nodes. 
\section{Online learning}
\label{sec:online}

As described above, we wish to estimate the rate functions
$\mu_k(\tau)$ for $k= 1,\ldots,p$ and the corresponding likelihood of
future events, based solely on previous events and the (possibly
learned) network structure. In this section, we describe several key
ideas from the field of online learning which we will leverage in our problem. First we 
describe the traditional online learning paradigm, then we describe methods which
incorporate dynamical models into the learning process, which allow one to 
adapt to a time varying environment.

\subsection{Online Learning in non-Dynamic Environments}

Online learning techniques are generally based
on the following paradigm: at every time point $t$ we make a
prediction, receive some data, and then do a few computationally
inexpensive calculations to improve our previous prediction.  In the
setting of autoregressive event tracking, this means we would have an
estimate about each actor's likelihood of acting and then see who
does act. Using the previous prediction, the current action, and
information about the network itself, we update our belief of who is
most likely to act next.  Unlike traditional online learning techniques,
there are strong dynamics involved in the evolution of the system that
must be incorporated. 

More formally, a generic version of an online method proceeds as
follows.  We let $\cX$ denote the domain of our observations, and
$\Lambda$ denote a bounded, closed, convex feasible set of the parameter of
interest.  Given sequentially arriving observations
$\bm{x} \in \cX^\infty$, we wish to construct a sequence of
predictions
$\hat\blambda = (\hlambda_1,\hlambda_2,\ldots) \in \Lambda^\infty$, where
$\hlambda_{t}$ may depend only on the currently available observations
$\bx_{t-1} = (x_1,\ldots,x_{t-1})$. The problem is posed as
  a repeated sequence of predictions given by a Learner and the truth
  being revealed by an oblivious (non-adaptive) Environment.  At time
$t$, the Learner computes a prediction, $\hlambda_{t}$ and the
Environment generates the observation $x_t$.  The Learner then
experiences the loss $\ell_t(\hlambda_t)$, where $\ell_t(\cdot)$ is a
convex cost function measuring the accuracy of the prediction
$\hlambda_t$ with respect to the data $x_t$. The task facing the
Learner is to create a new prediction $\hlambda_{t+1}$ based on the
previous prediction and the new observation, with the goal of
minimizing loss at the next time step.

\sloppypar We characterize the efficacy of $\wh{\blambda}_T \deq (\hlambda_{1},
  \hlambda_{2},\ldots,\hlambda_T) \in \Lambda^T$ relative to a
comparator sequence $\blambda_T\deq (\lambda_1,\lambda_2,\ldots,\lambda_T) \in
\Lambda^T$ as follows:
\begin{definition}[Regret] \label{defn:regret} The {\em regret} of
  $\hat{\blambda}_T$ with respect to a comparator $\blambda_T \in \Lambda^T$
  is
$$  R_T(\blambda_T) \deq  \sum_{t=1}^{T} \ell_{t}(\hlambda_{t}) - \sum_{t=1}^{T}\ell_{t}(\lambda_{t}).$$
\end{definition}

The comparator series can be thought of as the predictions from either an oracle with knowledge of future data or a batch algorithm with access to all the data.  Therefore, the regret characterizes the amount of excess loss suffered from the online algorithm.  Previous work proposed algorithms which yielded regret of
$O(\sqrt{T})$ for {\em static} comparators $\blambda_t$, where $\lambda_t = \lambda$
for all $t$ (\cf \cite{Zin03,BecTeb03,COMD}). Basically, these methods can only perform well if the comparator is a single point or changes either very slowly or very infrequently.   It is this characteristic that causes most existing methods to be poorly-suited to the autoregressive nature of interactions within a network.  

\subsection{Online Learning in Dynamic Environments}
More recent work has explored the impact of dynamical models within the context of online learning (\cf \cite{Rak12,DMD_journal}). In particular, the Dynamic Mirror Descent method proposed in \cite{DMD_journal} incorporates a known dynamical model into online learning, leading to significant improvements in performance in dynamic environments. In the context of multivariate Hawkes processes, a known dynamical model amounts to knowing the exact weighted adjacency structure of the network. In many practical contexts the network structure may be unavailable and will need to be estimated simultaneously with the rates.  This will be discussed further in Section \ref{sec:W_unknown}

Before defining an optimization routine specifically for multivariate Hawkes  data, we briefly describe a simplified version of the Dynamic Mirror Descent (DMD) method \cite{dynamicMirrorDescent}.  Let
$\Phi_t: \Lambda \times \mathcal{W} \mapsto \Lambda$ denote a sequence of known dynamical models that takes as input a value in our decision space and some side information, and set
\begin{subequations}
\label{eq:dmd}
\begin{align}
\tlambda_{t+1} &= {\rm{proj}}_{\Lambda} (\hlambda_t - \eta_t \nabla \ell_t(\hlambda_t)) \label{eq:dmd1}\\
  \hlambda_{t+1} &= \Phi_{t}( \tlambda_{t+1}, W_t) \label{eq:dmd2}
\end{align}
\end{subequations}
where $\eta_t$ is a step size parameter which controls how far we should step in the direction of the new data.
By including $\Phi_{t}$ in the process, we effectively search for a predictor which (a)
attempts to minimize the loss and (b) which is close to $\tlambda_{t+1}$
{\em under the transformation of $\Phi_t$} with side information $W$. In our setting $W$ will correspond to the known or estimated values of the network.
This is similar to a stochastic filter which alternates between using
a dynamical model to update the ``state'', and then uses this state to
perform the filtering action. However, we make no assumptions about $\Phi_t$'s relationship to
the true underlying parameters.  It has been shown, under mild conditions on the sequence $\{\Phi_t\}_{t>0}$, that the regret of this algorithm obeys the following:
$$ R_T(\blambda_T) \leq C \sqrt{T} \left(1+ \sum_{t=1}^{T-1}\|\lambda_{t+1}-\Phi_t(\lambda_t, W)\|\right)$$
for some $C>0$ independent of $T$.  This bound scales with the
comparator sequence's deviation from the sequence of dynamical models
$\{\Phi_t\}_{t>0}$ -- a stark contrast to previous tracking regret
bounds which are only sublinear in $T$ for comparators which change
slowly with time or at a small number of distinct time instances.
In order to use this framework to learn the rates and
  network of a
  Hawkes process we need to derive two key ingredients, the loss
  function and the dynamical model to be used in Equations \ref{eq:dmd1} and \ref{eq:dmd2}. These ingredients will take us from
  the general setup presented in this section, to the specific
  application being studied. These functions will be derived in the
  next section. Once these have been derived, we can use and expand
  upon the existing theory for online learning, and finally present a method to learn the rates and the underlying network simultaneously.

\section{Loss Function and Dynamic Model}
\label{sec:loss}
In order to analyze and make estimates of our point process network
data, we use the Hawkes model described in Section~\ref{sec:prob}
to define a loss function, dynamical models, and
other ingredients of the online learning framework described in
Section~\ref{sec:online}. 

\subsection{Time discretized loss function}
We discretize time into intervals of length $\delta>0$, where $\delta$ is  small enough that it is very infrequent that the same actor acts multiple times in the same time window. (For simplicity, we assume the total sensing time, $T$, is selected such that $T/\delta$ is an integer.) We let $t = 1,2,\ldots, T/\delta$ index these intervals, and note $N_{k,t\delta}$ is the number of events observed in the $k^{\rm th}$ process (i.e. by the $k^{\rm{th}}$ actor) up to the end of the $t^{\rm th}$ interval, $((t-1)\delta,t\delta]$, with $N_{k,0} \deq 0$.

The value $x_{t,k}=N_{k,t\delta}- N_{k,t(\delta -1)}$ denotes how many times actor $k$ acted during the $t^{\rm th}$ interval, which will mostly be either zero or one for an appropriately chosen $\delta$.  The vector $x_t\triangleq[x_{t,1},...,x_{t,p}]\trans$ will be our data vector at each time point. Using the negative log likelihood of the Hawkes process up to time $\delta t$, we can formulate appropriate loss functions to apply to an online setting.  We introduce an approximation of the time varying rate function in the Hawkes process, $\lambda_t=[\lambda_{t,1},...,\lambda_{t,p}]\trans\in\reals_+^p$.  To do this we define a new set of times $\{\bar{\tau}_n\}_n$ which are the ends of the discrete time intervals that the events occur.  These times are defined by $\bar{\tau}_n = \lceil \frac{\tau_n}{\delta}\rceil \delta$.  Here and for the rest of the paper, we denote the summation over a set of events $\{n:\bar{\tau}_n < \delta t\}$ by simply saying we sum over $\bar{\tau}_n < \delta t$.
\begin{align}
\lambda_{t,k}= \bar{\mu}_k + \sum_{\bar{\tau}_n < \delta t}W_{k,k_n} h\left(\delta t - \tau_n\right) \label{eq:timediscrete}
\end{align}
Equation \ref{eq:timediscrete} acts the same way as the original Hawkes process, but we do not immediately update the rates with the events as they occur but instead push them to integer multiples of $\delta$.  Notice that although we wait until $\bar{\tau}_n$ to include the event in the rate, we update it with the full knowledge of when the event actually occurred i.e. we use $h(\tau-\tau_n)$ instead of $h(\tau - \bar{\tau}_n)$.  Equation \ref{eq:timediscrete} suggests a loss function with analogous changes to the original, continuous time log likelihood.  We define $L_T(\mu)$ to be the negative log likelihood of the Hawkes process at time $T$, and $L_T^{(\delta)}(\lambda)$ to be the discrete time equivalent.  
\begin{align}
L_T(\mu) &\triangleq \sum_{k=1}^p\int_{0}^T \mu_k(\tau) d\tau-\sum_{n=1}^{N_T} \log \mu_{k_n}(\tau_n) \nonumber \\
\approx &\sum_{k=1}^p \left(\sum_{t=1}^{T/\delta} \delta \lambda_{t,k}- \sum_{t=1}^{T/\delta}  x_{t,k} \log \lambda_{t,k}\right) \triangleq L^{(\delta)}_T(\lambda) \label{eq:HawkesLogLikeDisc}
\end{align}
This new loss function is based on replacing the integral term with a summation and replacing $\mu_{k_n}(\tau_n)$ with $\lambda_{k_n,\bar{\tau}_n/\delta}$.  Both of these substitutions become closer to the truth as $\delta \rightarrow 0$.  In Lemma \ref{lem:disc_loss2} we characterize the difference between the two functions.

\begin{lemma}\label{lem:disc_loss2}
Given the influence function $h(\tau)=\alpha^\tau\ones_{[\tau>0]}$ and data with a maximum activity rate of $x_{\max}$ events per actor per unit time, the negative log likelihood of the true Hawkes Process, given in Equation \ref{eq:hawkesLogLike2}, and the approximate negative log likelihood for the discrete time rate, given in Equation \ref{eq:HawkesLogLikeDisc}, both generated by the same matrix $W$ and vector $\bar{\mu}$ with all elements $0\leq W_{i,j} \leq W_{\max}$ and $0 \leq \mu_{\min} \leq \mu_i < \infty$ respectively, there exists a constant $C>0$ depending on $x_{\max},p,W_{\max}, \mu_{\min}$ and $\alpha$ such that
$$|L_T(\mu) - L^{(\delta)}_T(\lambda)| \leq C N_T \delta.$$
{\bf{Remark:}} For a general influence function $h(\tau)$ which is Lipschitz on $(0,T]$, a similar proof gives a slightly higher bound of $C(T N_T \delta + N_T \log(1+N_T \delta))$. As we focus mostly on $h(\tau)=\alpha^\tau \ones_{[\tau>0]}$, we show the bound for this specific function.
\end{lemma}

This Lemma says that if $\delta$ is set small enough, the discrete approximation can be used in a learning algorithm, without many errors coming from the discretization approximation.  However, the smaller $\delta$ is the more frequently the rates will have to be updated, leading to a higher computational burden.

Thus the approximation justifies the proposed method using the instantaneous loss function:
\begin{align}
\ell_t(\lambda_t) \triangleq - \langle \log(\delta \lambda_t),x_t \rangle + \langle \delta \lambda_t, \ones \rangle. \label{eq:losses1}
\end{align}
Here and for the remainder of the paper, $\log$ and $\exp$ of a vector are assumed to be taken element-wise.
Notice that $L_T^{(\delta)}(\lambda) = \sum_{t=1}^{T/\delta} \ell_t(\lambda_t) + \langle x_t, \log \delta \rangle$, and thus the total cumulative loss is summation of instantaneous losses and a term which is independent of the rate estimate.    

\subsection{Dynamical models}
In order to use the DMD framework, we model the autoregressive nature of the Hawkes process using a series of data-dependent dynamical models, $\Phi_t$, which update a rate parameter $\lambda$ given a weighted adjacency matrix $W$ and the previously observed data $\H^t$. We can model this dependence using the dynamical model
$$\Phi_t(\lambda,W) = A_t \lambda + W y_t + c_t,$$
 for $\lambda, y_t, c_t \in \reals^p$, $A_t \in \reals^{p \times p}$, and $W \in \reals_+^{p \times p}$. 
If we let $A_t = \beta I$ for some $\beta \in (0,1)$, where $I$ is the identity matrix, it suggests that our dynamical model causes the rates in $\lambda$ to decay at a rate depending on $\beta$ in the absence of other effects.  The term $W y_t$ allows us to model autoregressive effects. In particular, the matrix $W$ could correspond to a weighted adjacency matrix associated with the network of interest, and $y_t$ could contain information about previous events as specified below. More generally, we might replace the term $W y_t$ with $\sum_{r=0}^{m-1} W  y_{t-r}$ for an $m^{\rm th}$-order process if we thought there should be some latency in the response times for pairs of actors.

Recall that the influence functions $h(\tau)$ describe how the causal influence between actors varies over time.  Dynamical models for various forms of $h(\tau)$ can be developed for the time discretized multivariate Hawkes process described in Equation \ref{eq:timediscrete}. First, a general function $h(\tau)$ is considered, with some mild assumptions, and then is used to derive models for some specific choices of $h(\tau)$.

\begin{itemize}
\item 
  \noindent{\bf General influence functions:} We assume $h(\tau)$ is a
  continuous, non-negative function on $\tau>0$.  Additionally, we assume there is a $B$ such that $h(\tau) > 0 $ for $0 < \tau \leq B$ 
  and $h(\tau) = 0 $ otherwise. Finally,
  let $e_{k_n}\in \reals^p$ be a vector of all zeros, with a single 1
  in the $k_n^{\rm{th}}$ entry indicating the actor involved in the $n^{th}$ action.  Then we derive the following dynamical model:
\begin{align*}
\lambda_{t+1} =& \bar{\mu} + \sum_{\bar{\tau}_n < \delta (t+1)} W e_{k_n} h(\delta (t+1) - \tau_n)\\
=& \bar{\mu} + \sum_{\bar{\tau}_n < \delta t} a_{t,n}W e_{k_n} h(\delta t-\tau_n)\\
& + \sum_{\bar{\tau}_n = \delta t} W e_{k_n} h(\delta (t+1) - \tau_n)\\
=& \bar{\mu} + A_{t} \sum_{\bar{\tau}_n<\delta t} W e_{k_n} h(\delta t-\tau_n) + W y_{t}\\
=& A_{t}\lambda_{t} + W y_{t} + (I-A_{t})\bar{\mu}
\end{align*}
In the above we have used the following values $a_{t,n}, A_t$ and $y_{t}$:
{\allowdisplaybreaks
\begin{align*}
a_{t,n} \triangleq& \begin{cases}
1, & h(\delta t - \tau_n) = 0\\
\frac{h(\delta (t+1)-\tau_n)}{h(\delta t - \tau_n)}, & {\rm{else}}
\end{cases}\\
A_{t,k} \triangleq & \begin{cases}
1/2 ,\hspace{.15 in} {\text{if}}\hspace{.15in}\displaystyle \sum_{\bar{\tau}_n<\delta t} W_{k,k_n} h(\delta t - \tau_n) =& 0\\
\frac{\displaystyle\sum_{\bar{\tau}_n<\delta t} a_{t,n}W_{k,k_n}h(\delta
  t-\tau_n)}{\displaystyle\sum_{\bar{\tau}_n < \delta t}
  W_{k,k_n} h(\delta t - \tau_n)}, &  {\rm{else}}
\end{cases}\\
A_t =& \diag(A_{t,1},A_{t,2},...,A_{t,p})\\
y_{t} \triangleq & \sum_{\bar{\tau}_n = \delta t} e_{k_n} h(\delta (t+1) - \tau_n)
\end{align*}}
Thus, we have the dynamics in the desired form.
$$ \lambda_{t+1}=\Phi_t(\lambda_t,W)= A_t \lambda_t + W y_t + (I-A_t)\bar{\mu}$$
Notice that in general $A_t$ may be a function of $W$.
\item
  \noindent{\bf Rectangular influence functions:} Using the above
  framework, dynamical models can be worked out for the
  specific instance when $h(\tau) = \ones_{[0<\tau<B]}$ for some
  positive $B>\delta$.  We first show the values $a_{t,n}$.
\begin{align*}
a_{t,n}=&\begin{cases}
1, \hspace{.94 in} \ones_{[0<\delta t - \tau_n<B]}=&0\\
\ones_{[0<\delta(t+1) - \tau_n < B]} / \ones_{[0<\delta t - \tau_n < B]},  &{\rm{else}} \end{cases}\\
=&\ones_{[\tau_n \leq \delta t - B]} + \ones_{[\tau_n > \delta (t +1) - B]}
\end{align*}
This leads to the following form of $A_{t,k}$.
\begin{align*}
A_{t,k}=\begin{cases} 1,  \displaystyle \sum_{\substack{\bar{\tau}_n<\delta t \\ \delta t - \tau_n <B}} W_{k,k_n} =0\\
\displaystyle\sum_{\substack{\bar{\tau}_n<\delta t \\ \delta(t+1) - B < \tau_n}} W_{k,k_n}/\displaystyle \sum_{\substack{\bar{\tau}_n<\delta t \\ \delta t - B <\tau_n}} W_{k,k_n}, & {\rm{else}} \end{cases}
\end{align*}
Notice that the elements of $A_t$, are the weighted ratio of how many events influence the current rate compared to how many events influence the rate at the previous time point.  Importantly, notice that $A_{t,k}\leq 1$.
\item 
\noindent{\bf Exponential influence functions:} Here we consider influence functions of the form $$h(\tau) =  \alpha^{\tau}\ones_{[\tau>0]}$$ for $\alpha \in (0,1)$. 
We then have
\begin{align*}
\lambda_{t+1} =& \bar{\mu} + \sum_{\bar{\tau}_n<\delta (t+1)} W e_{k_n} h(\delta (t+1) - \tau_n)\\
=& \bar{\mu} + \sum_{\bar{\tau}_n<\delta (t+1)}W e_{k_n}  \alpha^{\delta (t+1)-\tau_n}\\
=& \bar{\mu} + \alpha^\delta \sum_{\bar{\tau}_n<\delta t}W e_{k_n}\alpha^{\delta t-\tau_n} + \sum_{\bar{\tau}_n=\delta t}W e_{k_n} \alpha^{\delta (t+1) - \tau_n}\\
=&  (1-\alpha^{\delta})\bar{\mu} + \alpha^{\delta} \lambda_{t} + W y_{t} 
\end{align*}
yielding the dynamical model
$$\Phi_t(\lambda,W) = \alpha^{\delta}\lambda + W y_{t} +  (1-\alpha^\delta)\bar{\mu}.$$

\item 
  \noindent{\bf Delayed exponential influence functions:} The exponential decay might be a reasonable influence function,
  however reactions might not always be able to take place
  immediately.  To model this we use $h(\tau)=\alpha^{\tau-D}
  \ones_{[\tau > D]}$ for some positive delay $D\geq\delta$. In this
  scenario, a similar dynamical model can be derived as with the exponential decay,
  but with slight change in the additive $W y_t$ term:
\begin{align*}
\lambda_{t+1} = &\bar{\mu} + \sum_{\bar{\tau}_n < \delta (t+1)} W e_{k_n} \alpha^{\delta (t+1) - \tau_n - D} \ones_{[\delta(t+1)-\tau_n > D ]}\\
=& \bar{\mu}+ \sum_{{\tau}_n < \delta (t+1) - D} W e_{k_n} \alpha^{\delta (t+1)-\tau_n-D}\\
=&\bar{\mu} + \alpha^\delta \sum_{{\tau}_n + D <\delta t} W e_{k_n} \alpha^{\delta t - \tau_n - D}\\
&+ \sum_{\delta t \leq \tau_n + D < \delta(t+1)} W e_{k_n} \alpha^{\delta (t+1) - \tau_n - D}\\
=& \alpha^\delta \lambda_t + W y_{t}' + (1-\alpha^\delta)\bar{\mu}\\
y'_{t} \triangleq& \sum_{\delta t - D \leq \tau_n  < \delta(t+1) - D} e_{k_n} \alpha^{\delta (t+1) - \tau_n - D}
\end{align*}
This takes the same basic form as the non-delayed exponential, but with a slightly different term $y_{t}'$ instead of $y_t$.  Also, notice that when $D$ is equal to $\delta$ these equations become equivalent to the non-delayed version, suggesting that the time discretization is creating estimation error on the order of a slight delay in the influence function.
\end{itemize}

In the general setting, the dynamical model is written in the form

\begin{equation}
\Phi_t(\lambda,W) = A_t\lambda + W y_t + c_t
\label{eq:dyn}
\end{equation}
for some linear operator $A_t$, a vector $y_t$ which is a known function of $h$ and previously observed data, and a known constant $c_t$. In the following, we assume a generic dynamical model of the form \eqref{eq:dyn}. 

The first method we present will depend on the ease of
  computation of the matrix $A_t$, and if many observations need to be
  held in memory to compute $A_t$, then the method could be quite
  slow. However, if the influence function is the exponential decay or
  the delayed exponential decay, then $A_t$ is constant in time thus
  we only need to compute it once.  Another important feature of both
the exponential decay and delayed exponential decay functions is that
the linear operator $A_t$ does not depend on the values of the matrix
$W$. It is this separation that will allow simultaneously estimation
of the rates, $\lambda$ and the values of $W$.

\section{Proposed Algorithms}\label{sec:algorithms}

Our main contribution is to propose two algorithms, depending on whether or not the weighted adjacency matrix $W$ is known, and we show relevant regret bounds for both.  

\subsection{Proposed algorithm - $W$ Known}
We first present Algorithm \ref{alg:W_Known}, a method for tracking the rate vector $\lambda_t$ from streaming observations $x_t$ for $t=1,2,...T/\delta$.  The basic idea is the following: at time $t$ we start with the current rate estimate $\wh{\lambda}_t$. We then observe $x_t$ and incur the loss $\ell_t(\hlambda_t)$.  Based on this incurred loss, we update our previous prediction in the value $\tlambda_{t+1}$, which can be thought of as an {\em{a posteriori}} estimate of the rate at time $t$, given all the data up to and including $t$.  From here, we make our prediction of the rate at the next time by applying our dynamic model, $\Phi_t$.  

\begin{algorithm}[t]
\caption{MV Hawkes Tracking - $W$ Known}
\label{alg:W_Known}
\begin{algorithmic}[1] 
\STATE Initialize $\hat{\lambda}_1=\bar{\mu}$
\FOR {$t=1,...,T/\delta$}
\STATE Observe $x_t$ and incur loss $\ell_t(\hat{\lambda}_t) = \langle \ones,\delta \hat{\lambda}_t \rangle -  \langle x_t,\log(\delta\hat{\lambda}_t) \rangle$
\STATE Set $\tilde{\lambda}_{t+1} = \proj_{\Lambda}(\hlambda_t - \eta_t \grad \ell_t(\hlambda_t)) = \proj_{\Lambda} ((1-\eta_t) \hlambda_t - \eta_t x_t /\delta)$
\STATE Set $\wh{\lambda}_{t+1} = \Phi_t(\tlambda_{t+1}, W)$
\ENDFOR
\end{algorithmic}
\end{algorithm}

Algorithm \ref{alg:W_Known} admits the following result, which bounds the
amount of excess error of the output sequence generated by the
algorithm compared to any comparator sequence. The proof of Theorem \ref{thm:W_known} assumes that the decision space $\Lambda \triangleq [\lambda_{\min}, \lambda_{\max} ]^p$ for some $\lambda_{\min} > 0$ and $\lambda_{\max} < \infty$ and that there's a maximum amount of times any actor can act per unit time, denoted by $x_{\max}$ and therefore every element of the data vector takes values in the range $[0, \delta x_{\max}]$. We also assume the sequence of dynamical models is contractive with respect to a given Bregman divergence. We say that a dynamical model, $\Phi_t$ is contractive with respect to the Bregman divergence $D^*$ on the set $\Lambda$ if for any $\lambda_1,\lambda_2 \in \Lambda$ we have:
$$D^*(\delta\Phi_t(\lambda_1,W)\|\delta\Phi_t(\lambda_2,W)) - D^*(\delta\lambda_1\|\delta\lambda_2) \leq 0.$$
This is a condition which works to ensure some amount of stability in the output sequence by preventing small estimation errors at any one time step from getting worse and worse as the algorithm continues.  Lemma \ref{lem:contractive} proves sufficient conditions on the function $\Phi_t$ to ensure that it is contractive with respect to the needed Bregman divergence.

\begin{lemma}\label{lem:contractive}
If the dynamical model $\Phi_t(\lambda,W) = A_t \lambda + b_t$, where $A_t$ is a diagonal matrix with all elements in the range $[0,1]$ for all $t$ and $b_t\succeq 0$, then $\Phi_t$ is contractive with respect to the Bregman divergence induced by the function $\langle \lambda, \log \lambda \rangle - \langle \ones, \lambda \rangle$ on $\Lambda = [\lambda_{\min},\lambda_{\max}]^p$.
\end{lemma}
All the dynamical models we have proposed satisfy the conditions that $A_t$ is diagonal.  Additionally, $b_t \succeq0$ as long as all elements of $W$ and $\bar{\mu}$ are non-negative and $h(t)\geq 0$.  The most restrictive assumption that this lemma makes is that the elements of $A_t$ are upper bounded by one, which is true if $h(t)$ is non-increasing after the initial impulse.

\begin{theorem}[Tracking regret of Algorithm~\ref{alg:W_Known}]
  \label{thm:W_known} \sloppypar Using a sequence of contractive dynamical models $\Phi_t(\lambda,W)$ for all $t$, if we choose $\eta_t$ proportional to either $1/\sqrt{t}$ or $1/\sqrt{T/\delta}$, then there exists a constant $C>0$ depending on $\delta, p, x_{max}, \lambda_{\max}$ and $\lambda_{\min}$ such that the regret of $\wh{\lambda}_1,\wh{\lambda}_2,...,\wh{\lambda}_{T/\delta}$ generated by Algorithm \ref{alg:W_Known} for any data sequence $x_1,x_2,...,x_{T/\delta} \in [0, x_{\max}]^p$ with respect to a comparator sequence $\lambda_1,...,\lambda_{T/\delta} \in [\lambda_{\max},\lambda_{\min}]^p$  is bounded by: 
\begin{align*}\sum_{t=1}^{T/\delta} {\ell}_t(\wh{\lambda}_t) - {\ell}_t(\lambda_t)
 \leq C\left(1+ \sum_{t=1}^{T/\delta} \|\lambda_{t+1}-{\Phi}_t(\lambda_t) \|_2\right)\sqrt{T}.
\end{align*}
\end{theorem}

This algorithm takes as input known parameters $h(\tau), W$ and $\bar{\mu}$. With these known parameters and the data stream, one could simply calculate the rate at any given time directly by using Equation \ref{eq:timediscrete}. This would be equivalent to Algorithm \ref{alg:W_Known} with parameter $\eta_t = 0$. However, this strategy would be very fragile and susceptible to model mismatch. For instance, if the true influence function $h(\tau)$ has a shorter support in time than the estimate we use for direct calculations, then the predicted rates will depend on events too far in the past and will consistently over estimate the likelihood of events happening. In contrast, by adapting our estimate of $\lambda_t$ with a non-zero $\eta_t$ and dynamical models, we can mitigate this effect, thus incurring lower overall loss. Therefore we have gained robustness to model mismatch by not simply using a direct calculation method. 
These effects are demonstrated by a few important details of the regret bound.  The first is that if the complexity measure of the comparator sequence relative to the dynamics ${\Phi}_t$ is low, then the algorithm has $\sqrt{T}$ regret, which is sublinear as desired.  Secondly, no assumptions have been made about how the comparator sequence $\lambda$ was actually generated.  Instead we simply measure how well the comparator is approximated by a Hawkes process with dynamics dictated by ${\Phi}_t$.  

Therefore, if the true process acts like a Hawkes process, there will be low regret, but if the sequence is not generated this way or is generated as a Hawkes process with different parameters such as a different $W$ matrix, or a different influence function, we have an understanding about how much this will influence the performance of the algorithm. 

\subsection{Proposed algorithm - $W$ Unknown}\label{sec:W_unknown}

When the influence function $h(t)$ was a decaying exponential function, the dynamical model used had the form $\Phi_t(\lambda,W) = A_t \lambda + W y_t + (I-A_t) \bar{\mu}$, where $A_t=\alpha^\delta I$ was independent of the value of $W$.  This fact paired with the additional assumption that the solution to line 4 of Algorithm \ref{alg:W_Known} is a point on the interior of $\Lambda$, allows for a method of tracking both the rates $\lambda_1,..,\lambda_T$ as well as the matrix $W$.  We denote $\lambda_{t}^{W}$ as the estimate at time $t$ of Algorithm \ref{alg:W_Known} using matrix $W$ in line 5.  When the solution $\tlambda_{t+1}$ is on the interior of the set $\Lambda$, the value of $\wh{\lambda}^{W}_{t+1}$ takes the form:
 \begin{align}
\wh{\lambda}_{t+1}^{W}=(1-\eta_t) \alpha^\delta \wh{\lambda}^{W}_t + \eta_t \alpha^\delta \frac{x_t}{\delta} + W y_t + (1-\alpha^\delta)\bar{\mu}\label{eq:update}
 \end{align}
 It is this closed form solution that leads to Lemma \ref{lem:track_W}. It would seem that the assumption that $\tlambda_{t+1} \in \Int (\Lambda)$ would be very restrictive.  However, under very mild assumptions this will be true. For instance, since we are already assuming that there is a maximum amount of times any actor can act per unit time of $x_{\max}$, setting $\lambda_{\max} \geq x_{\max}$, insures the condition $\tlambda_{t+1} \leq \lambda_{\max}$.  Therefore, our space $\Lambda$ would be a bounded region, but the solution of line 4 would always be on the interior of this feasible set under the same assumptions as Theorem \ref{thm:W_known}. 
 
 \begin{lemma}\label{lem:track_W}
If Algorithm \ref{alg:W_Known} is run separately for $W_1$ and $W_2$ producing estimates $\wh{\lambda}_t^{W_1}$ and $\wh{\lambda}_t^{W_2}$ respectively at time $t$, with the dynamical model $\Phi_t(\lambda,W) = \alpha^\delta \lambda + W y_t + (1-\alpha^\delta)\bar{\mu}$, and assuming that the value $\tlambda_{t+1}$ is always in the interior of $\Lambda$, then at any given point in time the predictions of the algorithms corresponding to $W_1$ and $W_2$ will be related in the following way:
 $$\wh{\lambda}_{t}^{W_1} = \wh{\lambda}_{t}^{W_2} + (W_1 - W_2) K_t$$
 with
 $$K_{t+1} = (1-\eta_t)\alpha^\delta K_t + y_t, \hspace{.3 in}K_1=\boldsymbol{0}.$$
 \end{lemma}
 \begin{remark}
 In this section, we assume we still have knowledge of $\bar{\mu}$ and are trying to learn the time varying rates, $\lambda_t$, and the network structure, $W$. However, the exact same algorithm and analysis could be used to learn $\bar{\mu}$ using the following technique. Consider appending $\bar{\mu}$ as an extra column of the matrix $W$ and also appending $1 - \alpha^\delta$ to the $y$ vector. This would have a corresponding change in the dynamical model $\Phi_t(\lambda, W) = \alpha^\delta \lambda +  [W \bar{\mu}] [y_t\trans  1- \alpha^\delta]\trans$. Using this form and the technique of Lemma \ref{lem:track_W} we could simultaneously learn $W$ and $\bar{\mu}$, but for clarity of exposition we focus solely on learning $W$.
 \end{remark}
 Using this lemma, the losses that would have been incurred
 with a different weighted adjacency matrix $W$ can be calculated and used
 to update $\wh{W}_t$ using gradient descent, yielding $\wh{W}_{t+1}$,
 as described in Algorithm \ref{alg:hawkes}. To do this,
   a convex feasible set of influence matrices, denoted
   $\mathcal{W}$, must be defined. For instance, we might consider families of sparse
   $W$
$$\mathcal{W} = \left\{W \in \reals_+^{p \times p}: \|W\|_1 \leq c
\right\},$$
or low-rank $W$
$$\mathcal{W} = \left\{W \in \reals_+^{p \times p}: \|W\|_* \leq c
\right\},$$ or even $W$ with partially known support (\ie prior
knowledge of a subset of the elements of $W$ that are zero-valued).
First, the prediction $\wh{\lambda}_{t+1}$ is updated using the previous
estimate of the network, $\wh{W}_t$.  Then the estimate of $W$ is updated, and the transformation described in Lemma \ref{lem:track_W} is applied. 

\begin{algorithm}
\caption{MV Hawkes Tracking - $W$ Unknown}
\label{alg:hawkes}
\begin{algorithmic}[1] 
\STATE Initialize $\wh{W}_1 = W_0$, $K_1 = \zeros$, $\wh{\lambda}_{1} = \bar{\mu}$
\FOR {$t=1,...,T/\delta$}
\STATE Observe $x_t$ and incur loss $\ell_t(\wh{\lambda}_t) = \langle \ones, \delta \wh{\lambda}_t \rangle - \langle x_t, \log \delta \wh{\lambda}_t \rangle$
\STATE Set $\tilde{\lambda}_{t+1} = (1-\eta_t) \wh{\lambda}_t + \eta_t x_t/\delta$
\STATE Define $y_t \triangleq \displaystyle \sum_{\bar{\tau}_n = \delta t} e_{k_n} h(\delta (t+1) - \tau_n)$
\STATE Set $g_t(W) = {\ell}_t(\wh{\lambda}_t^{W} ) =\langle\ones,\delta \wh{\lambda}_t^{W} \rangle - \langle x_t,\log \delta \wh{\lambda}_t^W\rangle$
\STATE Set $\grad g_t(W) = \delta \ones K_t \trans - \diag(\wh{\lambda}_t^{\wh{W}_t})^{-1}x_t K_t\trans$
\STATE Set 
$\wh{W}_{t+1} = \proj_{\mathcal{W}} \left(\wh{W}_{t} - \rho_t \grad g_t(\wh{W}_t)\right)$

\STATE Set $K_{t+1} = (1-\eta_t)\alpha^\delta K_{t} + y_t$
\STATE Set $\wh{\lambda}_{t+1} = \Phi_t(\wt{\lambda}_{t+1},\wh{W}_t) + (\wh{W}_{t+1}-\wh{W}_{t})K_{t+1}$
\ENDFOR
\end{algorithmic}
\end{algorithm}

The next result establishes tracking regret bounds for Algorithm~\ref{alg:hawkes}:
\begin{theorem}[Tracking regret of Algorithm~\ref{alg:hawkes}]
  \label{thm:main} \sloppypar Let $\Phi_t(\lambda,W)=\alpha^\delta \lambda + W y_t + (1-\alpha^\delta)\bar{\mu}$ with $0<\alpha<1$ for all $W$ and $t = 1, 2, \ldots, T/\delta$. Additionally, let the sequence $\wh{\lambda}_1,\wh{\lambda}_2,...,\wh{\lambda}_T$ be the output of Algorithm~\ref{alg:hawkes}, and let $\lambda_1,\lambda_2,...,\lambda_T$ be an arbitrary sequence in
  $[\lambda_{\min},\lambda_{\max}]^p$. If we set $\eta_{t}$ and $\rho_t$ both proportional to either $1/\sqrt{t}$ or $1/\sqrt{T/\delta}$, then for any data sequence $x_1,...,x_{T/\delta}$ in $[0, \delta x_{\max}]^p$ with $x_{\max} < \lambda_{\max}$, there exists a constant $C>0$ which depends on $\delta, p, x_{max}, \lambda_{\max}$ and $\lambda_{\min}$ such that
\begin{align*}
\sum_{t=1}^{T/\delta} \ell_t (\wh{\lambda}_t) - &\sum_{t=1}^{T/\delta}\ell_t(\lambda_t)\\
\leq C\Bigg(1&+\min_{W \in \mathcal{W}} \sum_{t=1}^{T/\delta} \|\lambda_{t+1}-\Phi_t(\lambda_t,W)\|_2\Bigg)\sqrt{T}.
\end{align*}

\end{theorem}

This theorem is proved in Appendix~\ref{app:pf_main}. This bound says that using Algorithm~\ref{alg:hawkes} achieves an average per-round loss which is nearly as low as what would have been achieved with access to all data to choose the optimal time-varying rate vectors with a batch method. The gap between the losses of the proposed method and the losses accrued with a batch method scale with how closely the batch output (\ie comparator sequence) follows the dynamical model associated with the {\em best} estimate of the network structure as encapsulated by $W$. For instance, imagine that there existed a true, fixed $W$ representing a network, and an oracle used this $W$ to estimate a sequence of rate vectors which followed the model in \eqref{eq:dyn} exactly and, subject to that constraint, minimized the sum of losses. For that oracle, the variation $\sum_{t=1}^{T-1}\|\lambda_{t+1}-\Phi_t(\lambda_{t},W)\|_2= 0$. Clearly such an estimator is not practical because we do not know $W$ and are operating in an online setting. Despite these disadvantages, the difference of the average per-round losses of Algorithm~\ref{alg:hawkes} and the oracle estimator scales like $1/\sqrt{T}$, so that as $T \rightarrow \infty$, the performance gap between the two methods vanishes.

These bounds do not rely on any assumptions about the data actually being generated by a multivariate Hawkes process or even being stochastic (which would be a fallacy in any real-world application). Rather, the ideas underlying the multivariate Hawkes model are used to generate a loss function and dynamical model. These values are used to characterize how well our methods perform tracking in an online setting relative to how well any other method might perform {\em on the same set of observations}. Further, the comparator sequence against which performance is measured might be computed in batch  (rather than online) or using significantly more computational and memory resources than are required by Algorithm~\ref{alg:hawkes}.

The intuition behind why this method is robust to model mismatch can be seen in the algorithm itself and in the regret bound. Notice in line 4 of the algorithm, we are directly adjusting our estimate of the rate, prior to adjusting the network weights. Because we are adjusting not only our estimate of the network weights and directly calculating the resulting rate, our sequence of estimates is allowed to deviate from a pure Hawkes process. This amount of deviation can allow us to be more flexible to combat the errors due to model mismatch. Additionally, the form of the regret bound tells us that the method will perform competitively against any set of comparators that nearly follows the dynamical model. Therefore if the generative model is similar, although not exactly a Hawkes process, then this variation term will still be low, and result in low overall loss. To see how Algorithm \ref{alg:hawkes} adds robustness to the estimation of $W$, consider two contrasting approaches. In the first, which is detailed in Appendix \ref{app:OGD} and considered in our experimental results, we estimate $W$ by performing online gradient descent on $W$ with a loss function corresponding to the Hawkes negative log likelihood (which may contain incorrectly estimated model parameters). In the second approach, corresponding to Algorithm 2, we again estimate $W$ via online gradient descent, but this time with a loss function based the accuracy of predictions from Algorithm \ref{alg:W_Known}. While Algorithm \ref{alg:W_Known} may also depend on incorrectly estimated model parameters, it is much more robust to model mismatch than simply using the Hawkes generative model as in the first approach. This robustness is thus inherited by Algorithm \ref{alg:W_Known}. 

\section{Computational Complexity}
\label{sec:computation}
One important feature of the proposed method is the low computational cost per iteration.  Algorithm \ref{alg:hawkes} performs the tasks of estimating both the current intensity $\wh{\lambda}_t \in \reals^p$ and the network relationships $\wh{W}_t \in \reals^{p \times p}_+$.  A brief examination of lines 3 - 6 of the algorithm shows mostly vector operations on length $p$ vectors, requiring $O(p)$ operations.  The main computational burden of the algorithm comes in line 7, with the matrix multiplications requiring $O(p^2)$ operations and in line 8 projecting onto the space $\mathcal{W}$.  Without the projection step, this leaves the algorithm at an overall complexity of $O(p^2)$ to estimate $p^2 + p$ values.  Depending on the space $\mathcal{W}$, the algorithm may be slower.  For instance, a reasonable space would be the space of matrices with a bounded nuclear norm, which requires computing a singular value decomposition at each step requiring $O(p^3)$ operations.  Projecting onto other spaces, such as an $\ell_1$ ball with some radius, would only require $O(p^2)$ operations, maintaining our baseline complexity.

\section{Experimental Results}\label{sec:experiments}

\begin{figure*}[t!]
\centering
\subfloat[Mean cumulative loss averaged over 100 iterations. Error bars show one standard deviation above and below mean.]{\includegraphics[height=1.75in]{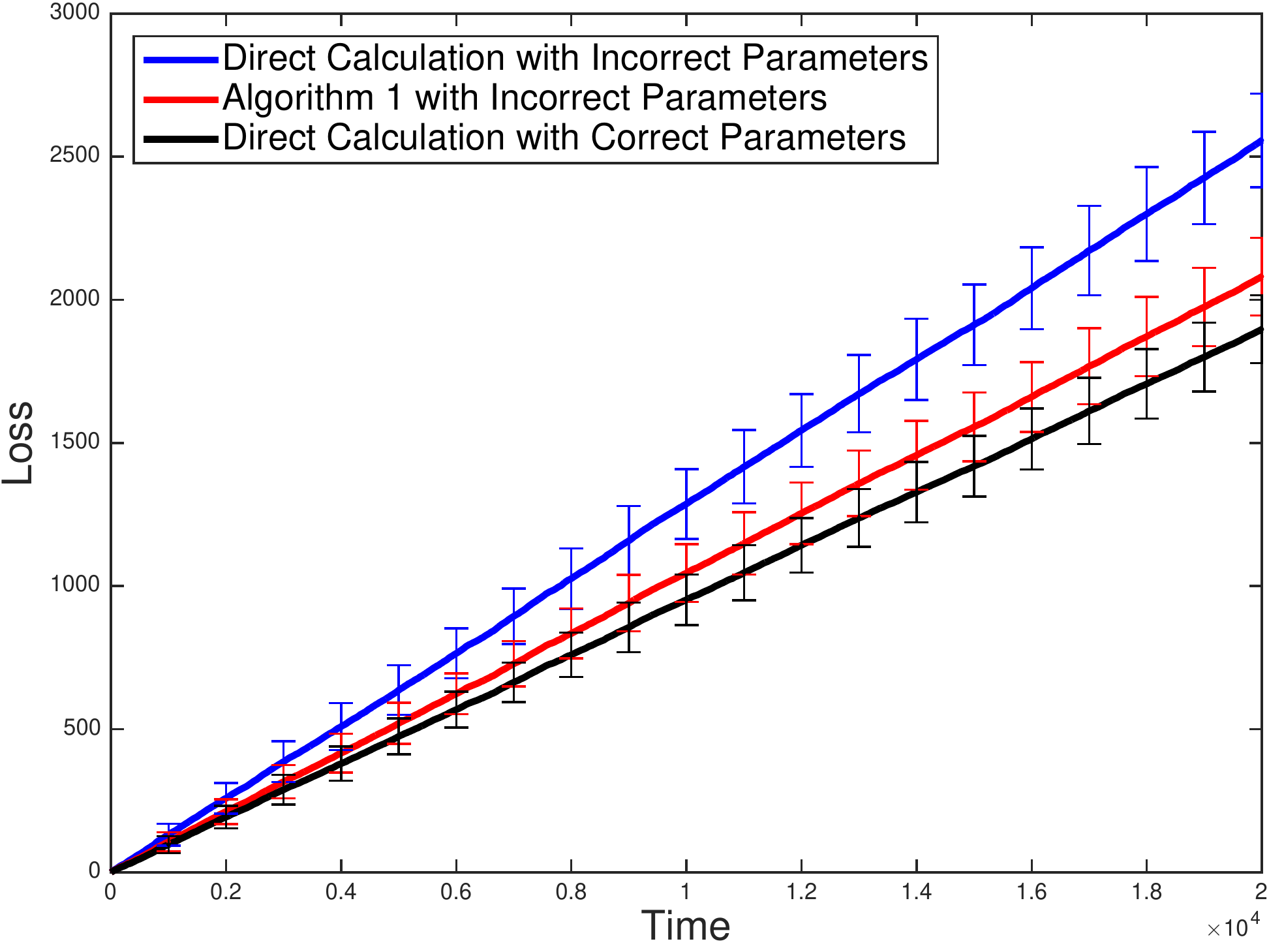}}~
\subfloat[Moving average loss with D = 250, averaged over 100 iterations.]{\includegraphics[height=1.75in]{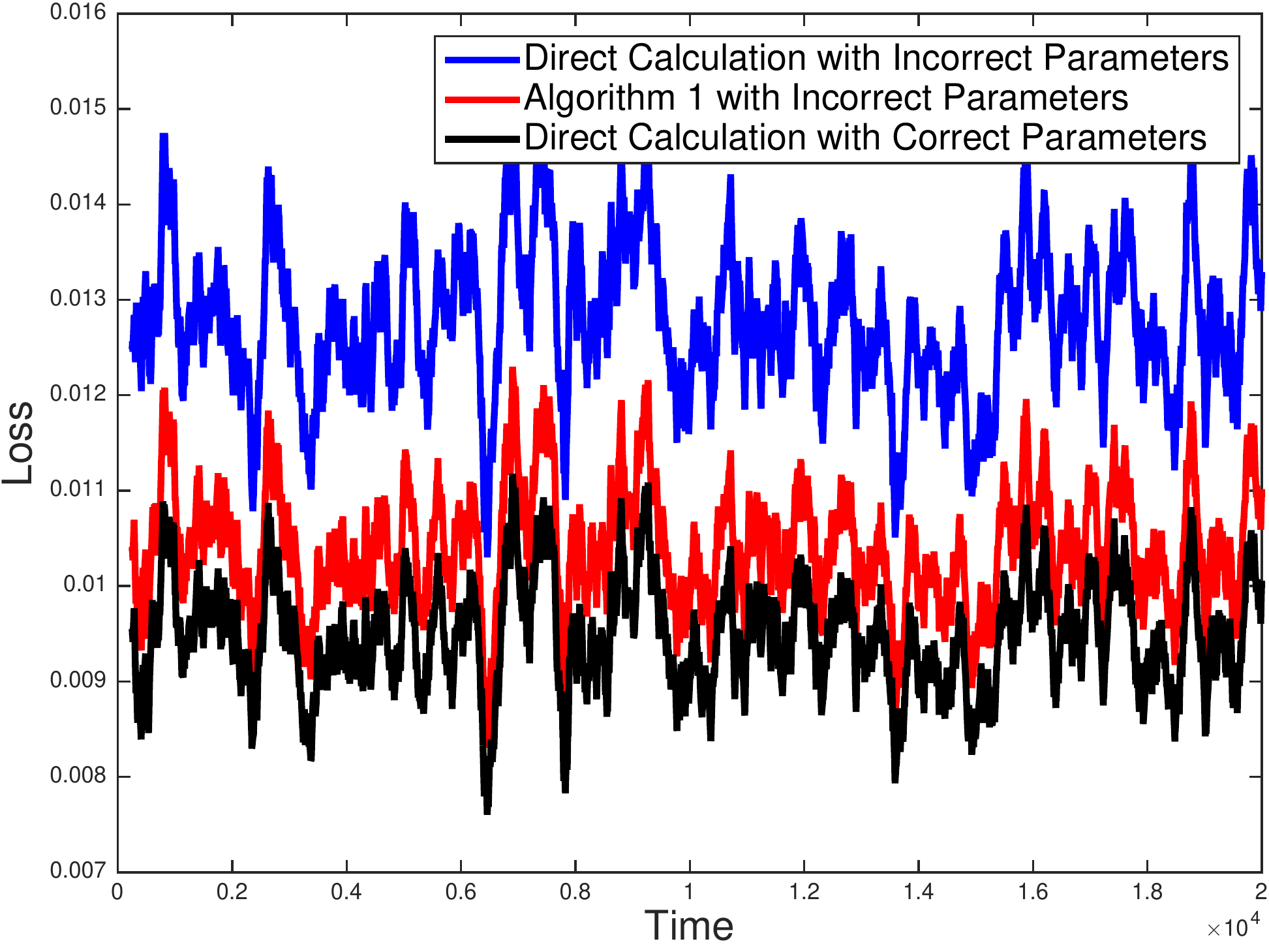}}~
\subfloat[Percentile of moving average difference between direct calculation and Algorithm \ref{alg:W_Known}.]{\includegraphics[height=1.75in]
{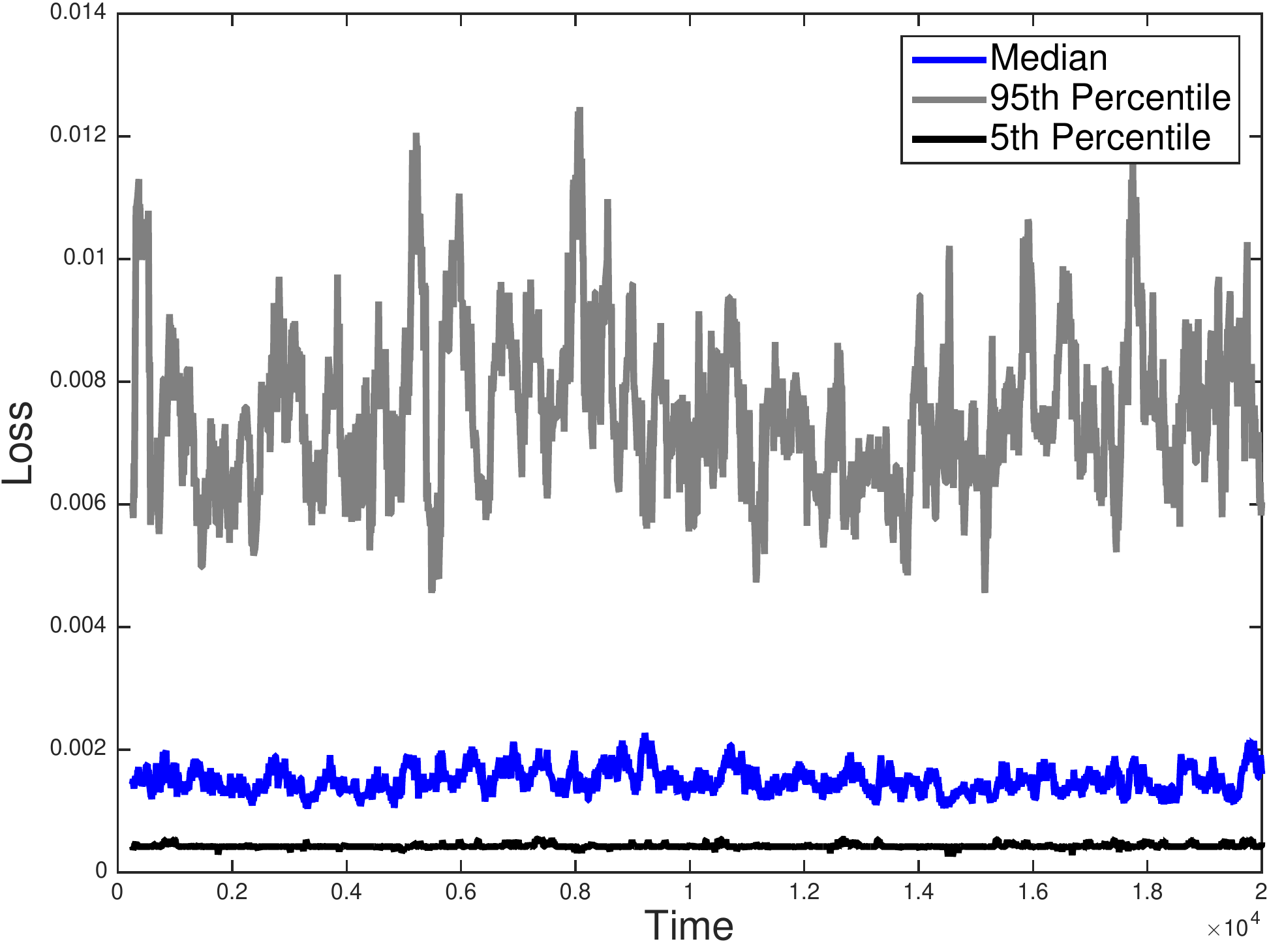}}
\caption{Performance of Algorithm \ref{alg:W_Known} using an incorrect exponential decay influence function.  Tracking the rate instead of just calculating it from known network values and the assumed influence function leads to better overall performance. The error bars on the left side of (a) may be overlapping, but if we look at the difference of individual trials we can see that the learning the rates gives consistently better performance than direct calculation on a per trial basis. The plot in (c) shows the percentile information of the difference between direct calculation and Algorithm \ref{alg:W_Known}, showing that our method adds robustness to the estimation procedure in over 95\% of the cases across time and data realizations.
}
\label{fig:Mismatch_exp}
\end{figure*}

In this section we present several experiments to demonstrate salient features of the proposed algorithms.  We first focus on the scenario where the network
influence matrix, $W$, is known.  In this scenario the important
observation is that our method is more robust to model mismatch than just
calculating the rate directly from the observations, assumed influence
function and matrix $W$. The next set of experiments demonstrates Algorithm \ref{alg:hawkes} for unknown $W$ on synthetic data and demonstrates how it can be used to learn both rates and the network of interest. Finally, we use Algorithm \ref{alg:hawkes} to analyze six months of Memetracker data corresponding to posts by a selection of well known news websites to try to determine what relationships exist amongst these organizations.

Throughout this section, we compare our method to the classical online learning
algorithm Online Gradient Descent (OGD) used to learn the network $W$. This method is described formally in Appendix \ref{app:OGD}.
One alternative algorithm to learn the network would be to simply count all the times one process has an event immediately after another process had an event, with larger counts corresponding to larger influence. OGD uses a current estimate of the network, and then uses the loss function and the assumed influence function, to update the network estimate in the direction of the most recent data point. Therefore the estimate is a weighted average of previously seen data, with more weight put on more recent data. In this way OGD is basically the same as the counting process but with more information put into the system. We compare against this method and show in several ways that our method performs comparatively when both methods know the correct influence function, but our method performs better when model information is misspecified.

One challenge of online methods is the tuning of the
  step-size parameters, in our case $\eta$ and $\rho$. In all of our
  experiments, we measure the effectiveness of step-size candidates based
  solely on accumulated loss on a small subset of the data. For
  instance, given a new dataset, we could run the method several times on the first
  5\% of the data with a range of step-size parameters, observe
  the total accumulated loss, and choose the parameters which minimize
  the loss over that time empirically. The basic setup of online
  learning protects us from over-fitting because setting step-sizes
  too large would push our estimates very close to the
  immediately preceding observation, which would cause large loss on
  the next observation. Conversely, very small step-sizes would not
  adapt or learn the parameters at all, also causing high accumulated
  loss. 
  
Throughout this section we plot several curves of interest to demonstrate the efficacy of our methods. The first metric is cumulative loss as defined at time $t$ as $\sum_{\tau=1}^t {\ell}_t(\lambda_t) $, for an estimator $\lambda_t$. This value will be plotted for values of $t=1$ to $T$.  Additionally, we plot a moving average curve of instantaneous loss, defined at time $t$ as $\frac{\delta}{D} \sum_{i=0}^{D/\delta-1} {\ell}_{t-i}(\lambda_{t-i})$, for a time window of width $D$. This curve gives an idea of the instantaneous loss, while not being so susceptible to noise as to be indecipherable.

\subsection{Model mismatch, $W$ known}

For the first experiment, data points were generated in a two-actor
network $(p=2)$, with $W$ an identity matrix scaled by $3/4$, the
influence function $h(t)=e^{-t} \ones_{[t>0]}$ and $\bar{\mu}$
was $[.005 .005]^\top$ for a time horizon of 20000, using the method of
\cite{expHawkesSimulation}.  The data was then processed in several ways.  The first was to calculate the discrete time
rates using an incorrect influence function, $\tilde{h}(t)=(2e)^{-t}
\ones_{[t>0]}$, without doing any learning.  In other words, a rate is estimated by plugging observed event times into Equation \ref{eq:timediscrete} using the assumed $W$, $\tilde{h}(\cdot)$, and $\bar{\mu}$.  We will call this method direct calculation.
  However, we expect suboptimal performance due to
the fact that $\tilde{h}(t)\neq h(t)$. This method is compared
to the output of Algorithm \ref{alg:W_Known} with the same, incorrect
$\tilde{h}(t)$ function, with $\delta=0.1$ and $\eta_t=10/\sqrt{T/\delta}$, to show that robustness to model mismatch has been added.  This overall setup was run separately on 100 different data realization, and the results are shown in Figure
\ref{fig:Mismatch_exp}.

\begin{figure*}[t]
\centering
\subfloat[Mean cumulative loss averaged over 100 iterations. Error bars show one standard deviation above and below mean.]{\includegraphics[height=1.75in]{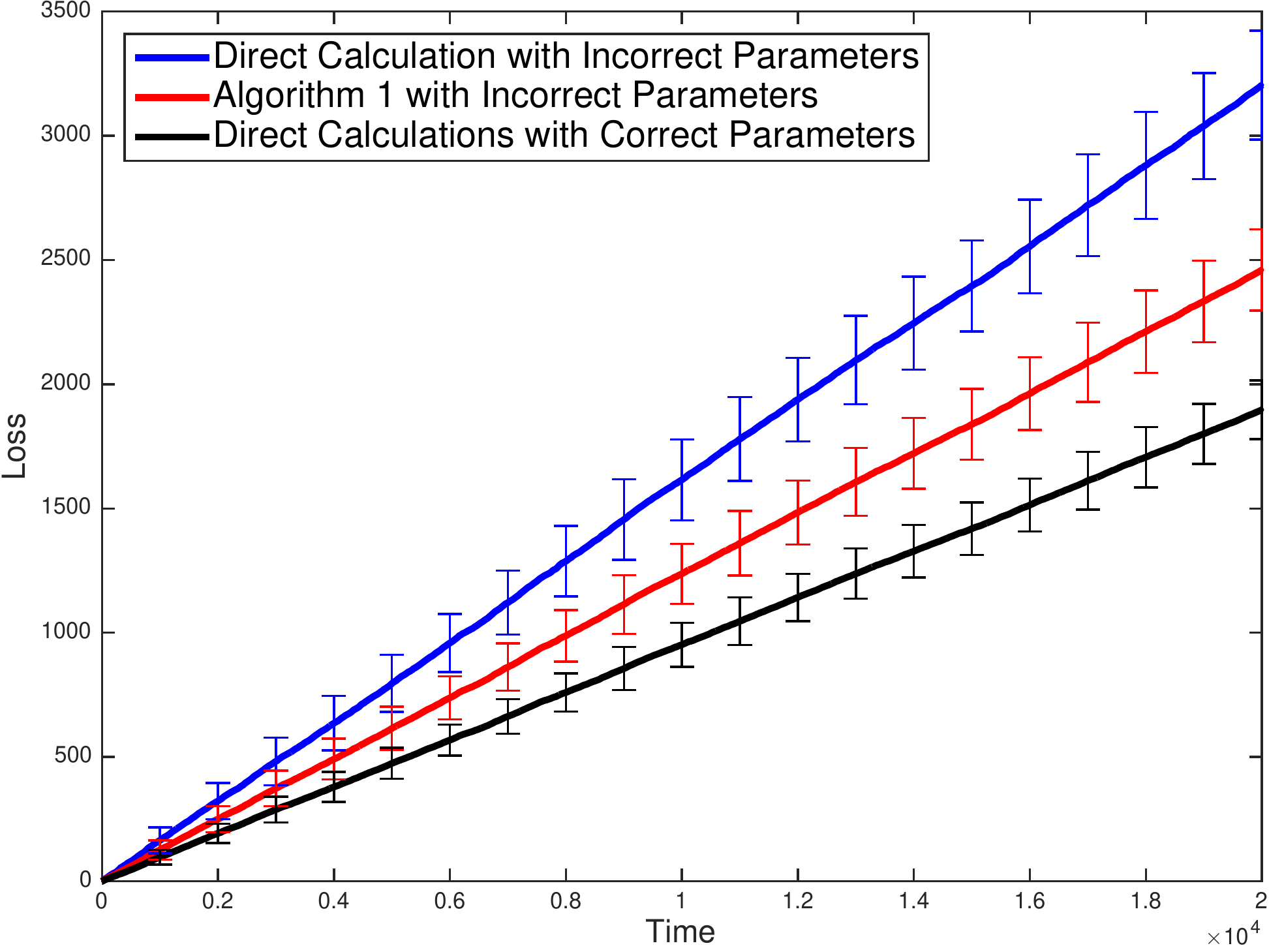}}~
\subfloat[Moving average loss with D = 250, averaged over 100 iterations.]{\includegraphics[height=1.75in]{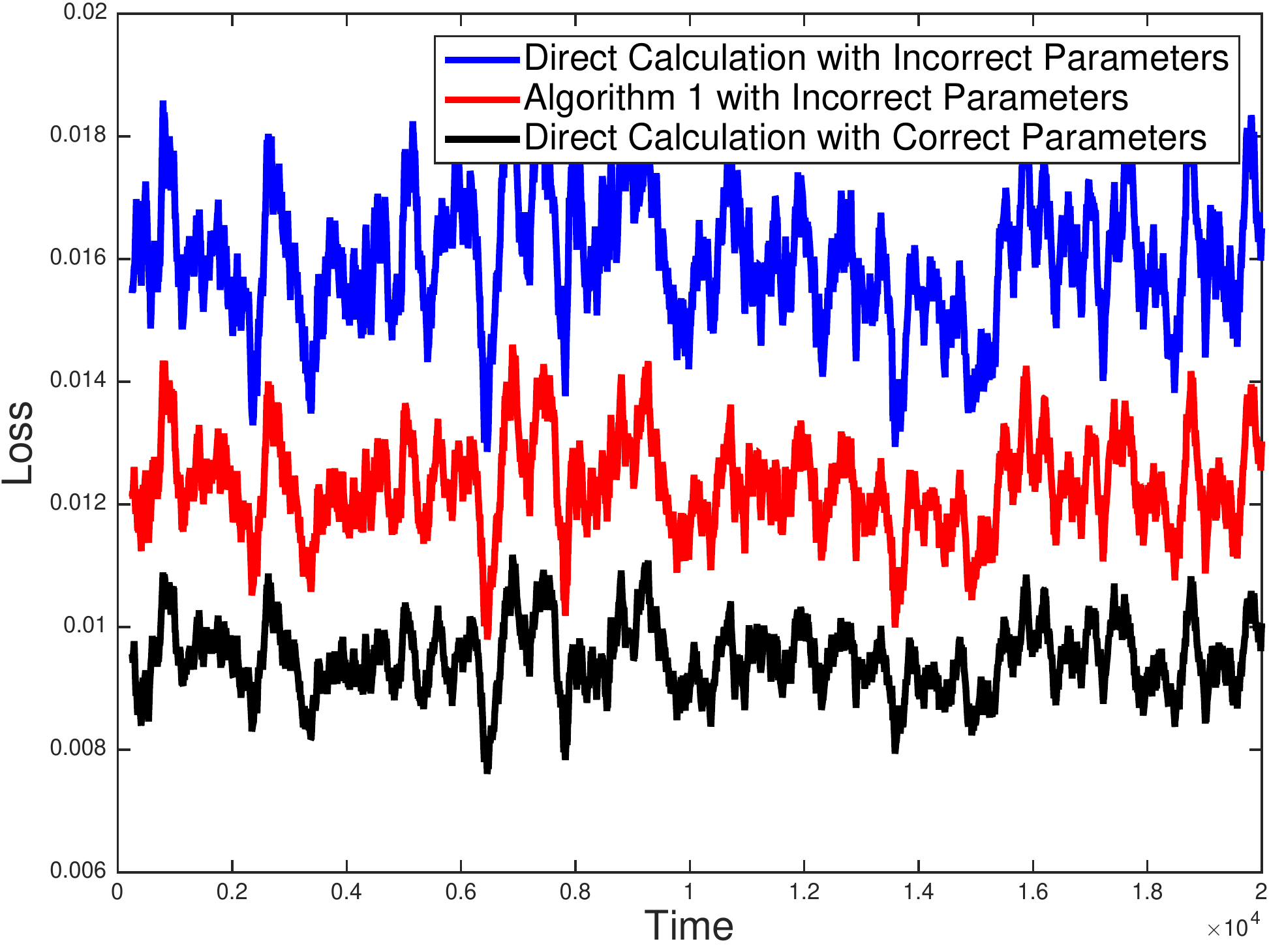}}~
\subfloat[Percentile of moving average difference between direct calculation and algorithm \ref{alg:W_Known}.]{\includegraphics[height=1.75in]
{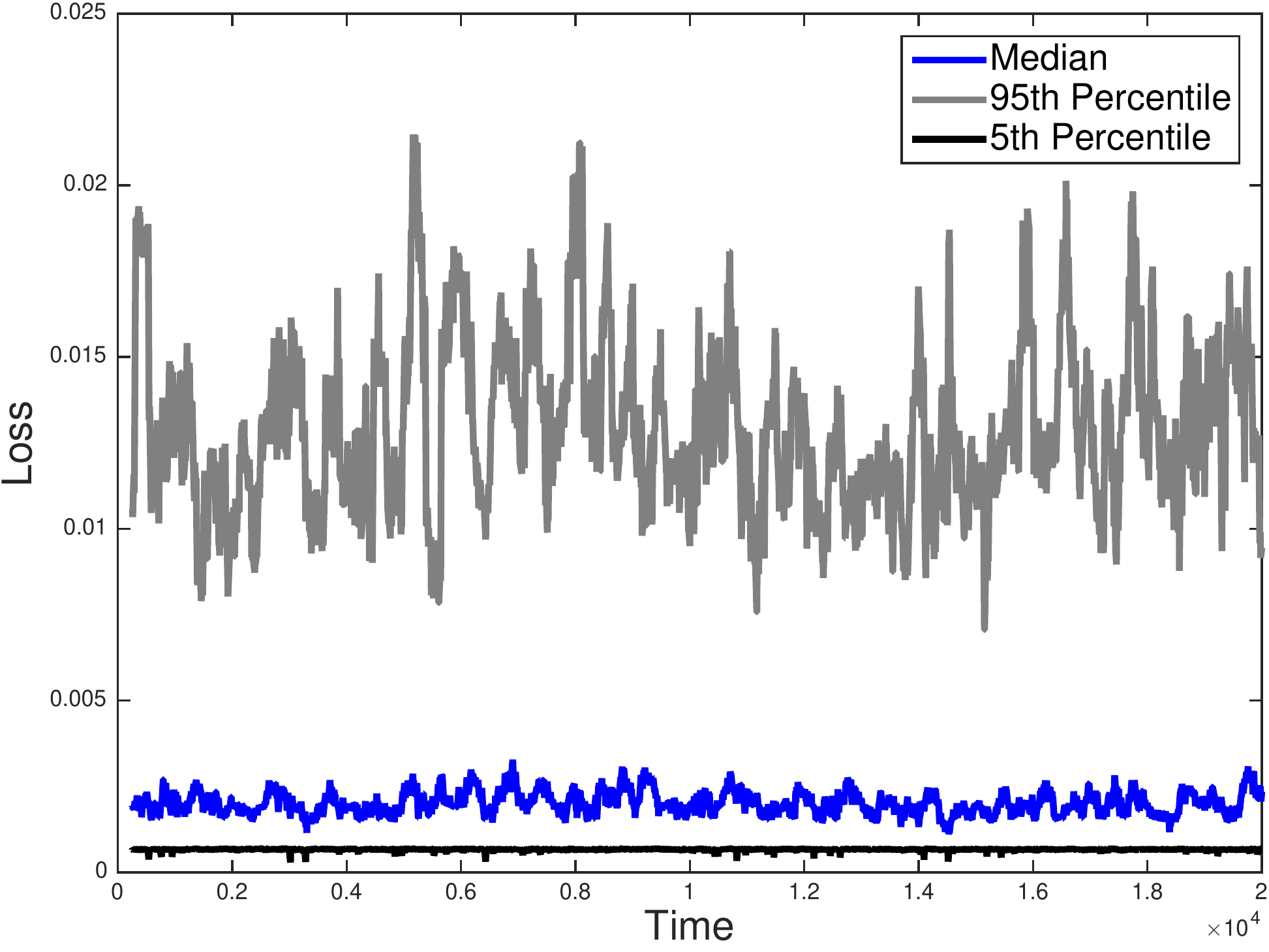}}
\caption{Performance of Algorithm \ref{alg:W_Known} using an incorrect rect influence function. 
Again, tracking the rate has added robustness to misspecified system parameters.}
\label{fig:Mismatch_sq}
\end{figure*}

Another experiment was run on the same data, generated using $h(t)=e^{-t} \ones_{[t>0]}$, and the same true $W$ matrix, but this time the influence function used to estimate the rates was $\tilde{h}(t)=\ones_{[0<t<5]}$ and all other parameters are kept the same. These results are shown in Figure \ref{fig:Mismatch_sq}.  Again, learning the rates instead of performing direct calculations using an incorrect influence function has added more robustness to model mismatch. In both Figures \ref{fig:Mismatch_exp} and \ref{fig:Mismatch_sq} plots (a) and (b) show that the proposed method accrues less loss on average than the direct calculation and is much closer to the loss incurred by the true rate. The plots (c) show that the difference between the loss incurred by direct calculation is not only higher on average than our method, but also is higher in almost every individual case, as the 5th percentile of the difference between direct calculation and our method is above zero.

\subsection{Learning $W$}

\begin{figure}[t]
\centering
\includegraphics[height=1.5 in]{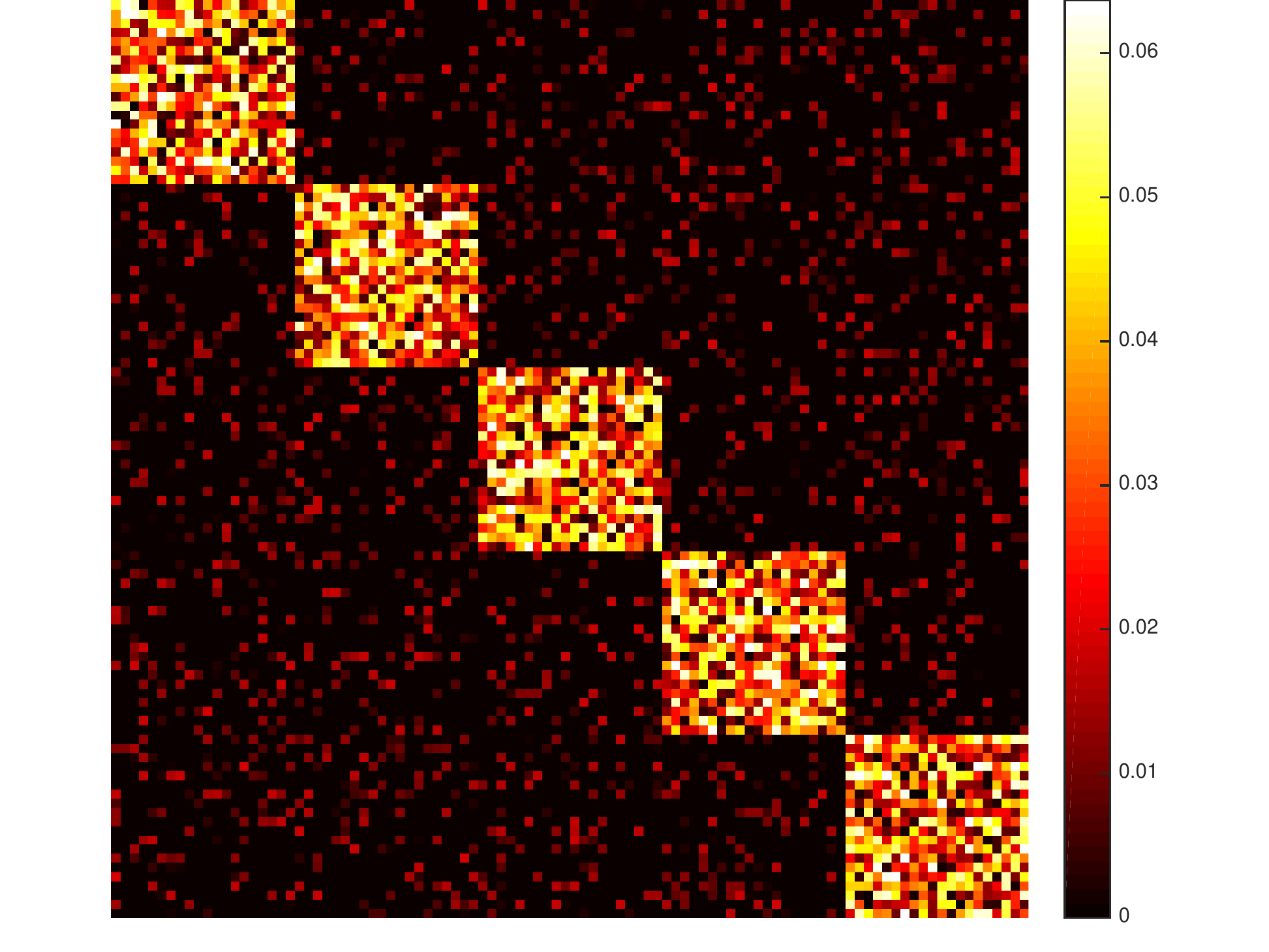}
\caption{True network used to generate event times.  Each pixel represents the influence the actor represented by the column has on the actor represented by the row, with lighter colors meaning more influence, and black meaning no influence.}
\label{fig:TrueNetwork}
\end{figure}

\begin{figure}[t!]
\centering
\subfloat[Moving average loss with time window $D=500$ averaged over 100 data realizations.]{\includegraphics[height=1.75in]{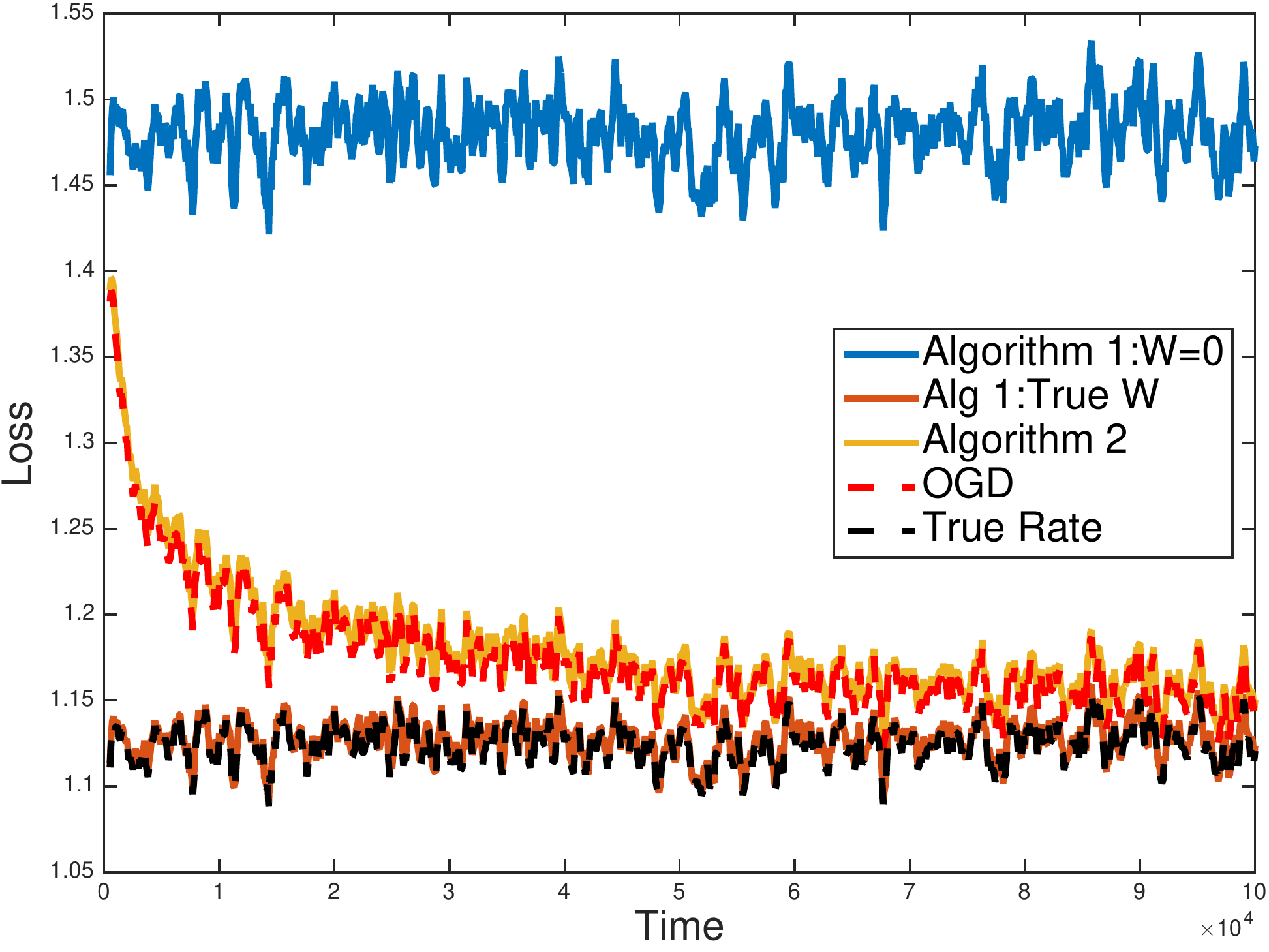}}~
\subfloat[Percentiles of difference between moving average losses of OGD and Alg \ref{alg:hawkes} estimates.]{\includegraphics[height=1.75in]{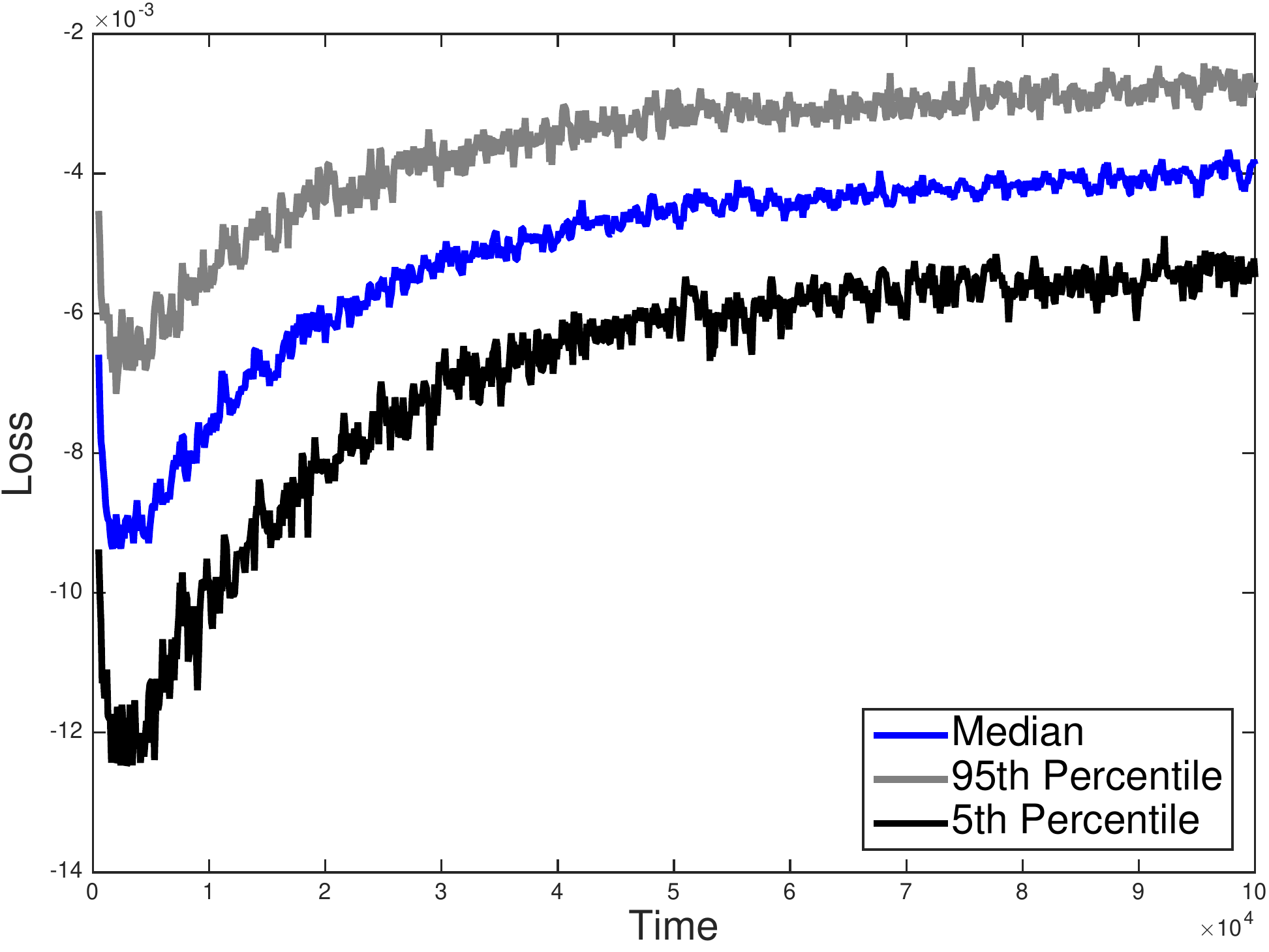}}
\caption{Performance of Algorithm \ref{alg:W_Known} with different values of $W$, compared to Algorithm \ref{alg:hawkes}.  Both our method and Online Gradient Descent (OGD) learn the structure of the true network and has performance that approaches Algorithm \ref{alg:W_Known} with the true value of $W$.
}
\label{fig:ResultsAlg2}
\end{figure}

In the next set of experiments, the ability of Algorithm \ref{alg:hawkes} to learn the network structure and rates from just event timing data is tested.  Networks of 100 actors $(p=100)$ is used with a true underlying network with small, densely connected subgraphs.  Each network, one example shown in Figure \ref{fig:TrueNetwork}, was generated by selecting the 20 $\times$ 20 blocks along to diagonal to have values chosen randomly on $[0,1]$, and all of the off diagonal elements being non-zero with probability 0.2, with strength randomly chosen on $[0, .3]$. Finally, each matrix was normalized such that it had a maximum singular value of 0.8 for stability.  100 such networks were generated, and one data realization was created for each network with a time horizon of $T=100000$. The value of $\bar{\mu}$ was uniformly generated on $[0.001, .01]^p$, and the influence function used was $h(t)=e^{-t}\ones_{[t>0]}$ and $\delta=.01$. Algorithm \ref{alg:hawkes} was then run with $\eta_t = 10/\sqrt{T/\delta}$ and $\rho = .01/\sqrt{T/\delta}.$ Additionally, the estimates of the network were regularized with the element-wise $\ell_1$ norm with regularization parameter $.001$ to encourage sparsity in the estimated networks.  Figure \ref{fig:ResultsAlg2} shows the results for Algorithm \ref{alg:W_Known} where the value of $W$ used was the generating value, and where $W$ was all 0s.  We compare these two to the result of Algorithm $\ref{alg:hawkes}$ and estimating $W$ using OGD with step size $\rho_t$, averaged over 100 data realizations.

\begin{figure}[t]
\centering
\subfloat[Moving average loss with time window $D=500$ with incorrect influence function averaged over 100 data realizations]{\includegraphics[height=1.75in]{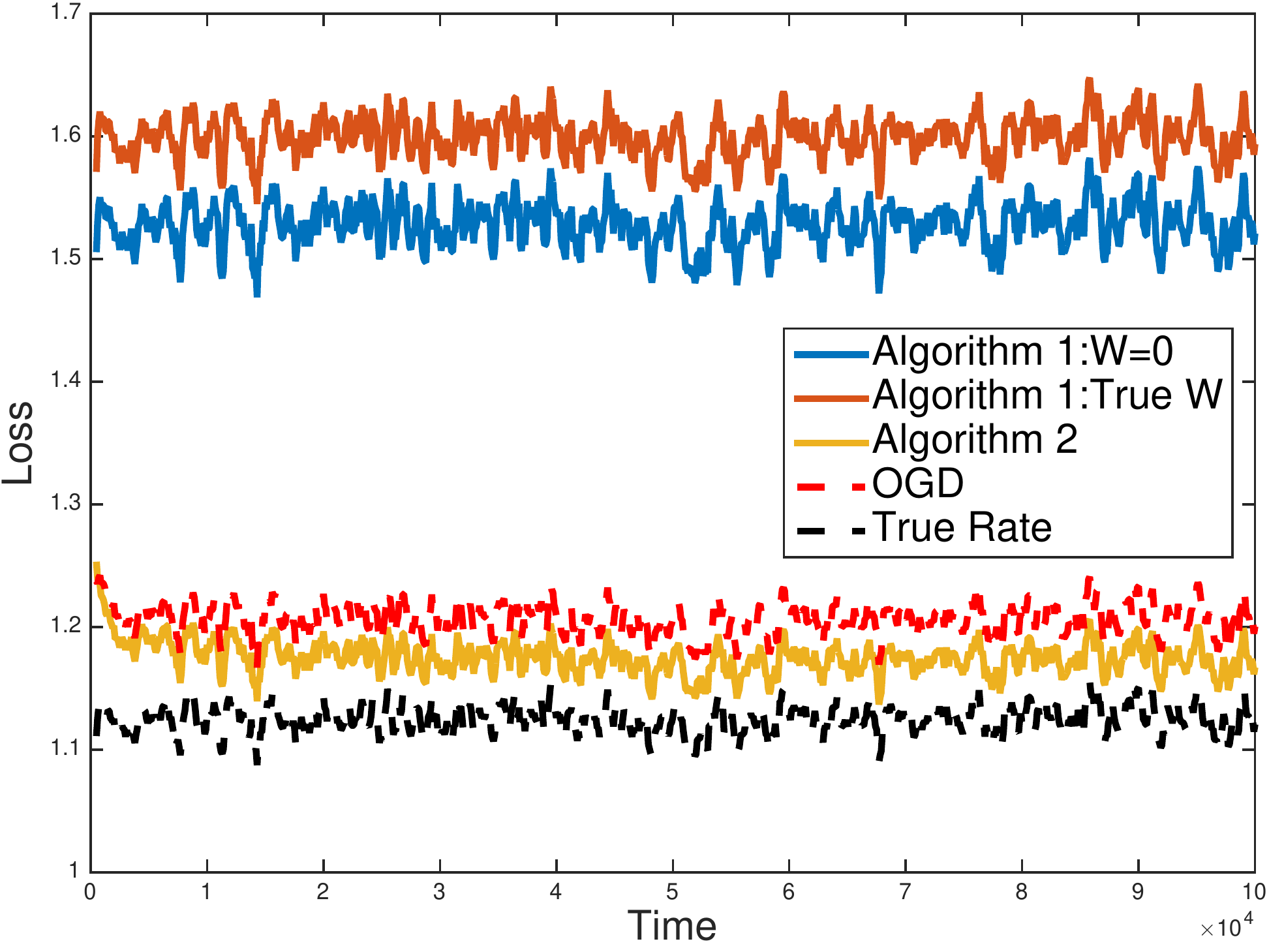}}~
\subfloat[Percentiles of difference between moving average losses of OGD and Alg \ref{alg:hawkes} estimates with incorrect influence function.]{\includegraphics[height=1.75in]
{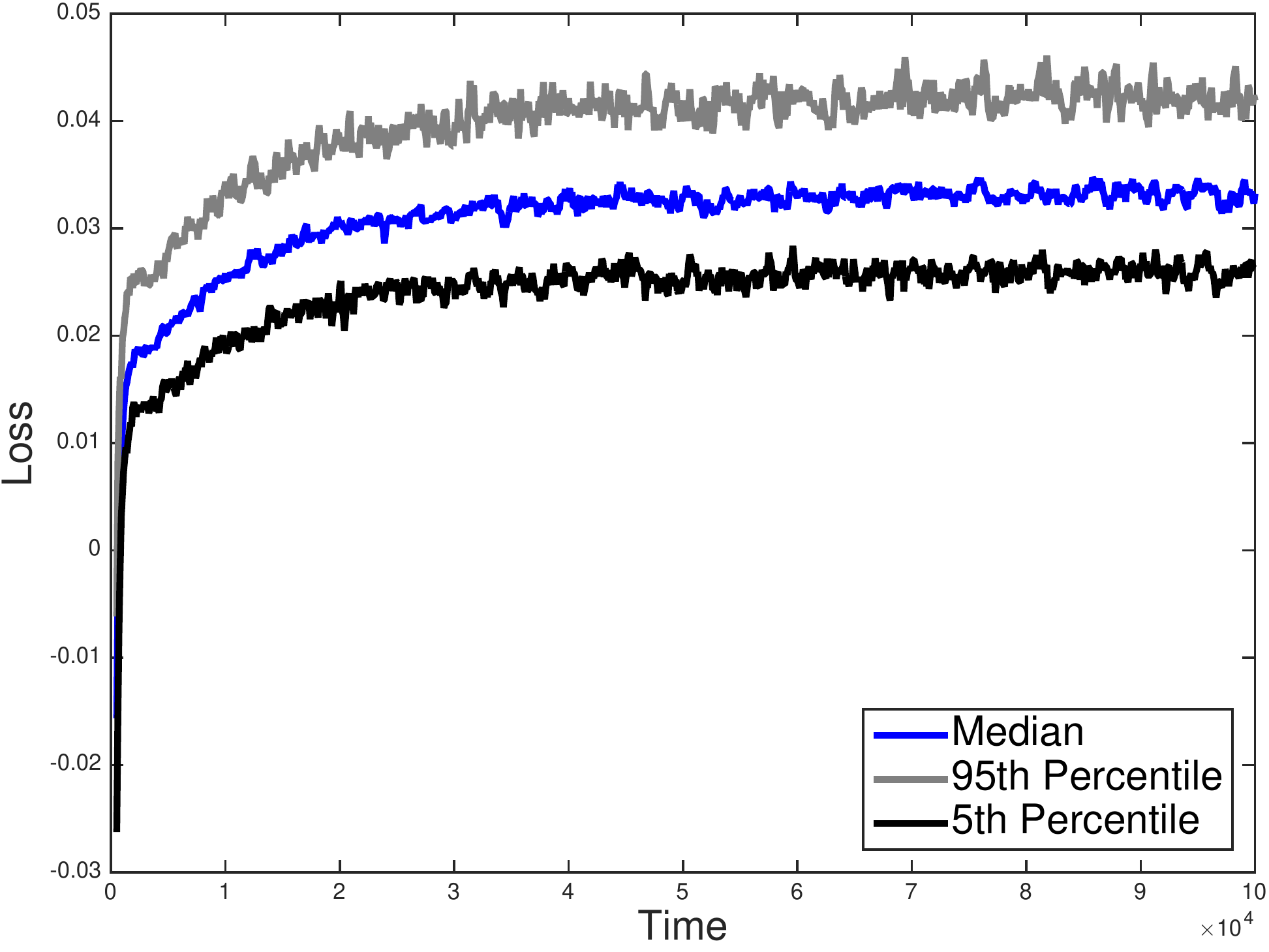}}
\caption{Performance of Algorithm \ref{alg:W_Known} with different values of $W$, compared to Algorithm \ref{alg:hawkes}. When the system parameters are misspecified, our method outperforms OGD on W in predicting likelihood of actors participating in events.  
}
\label{fig:ResultsAlg2_Mismatch}
\end{figure}

The important feature of Figure \ref{fig:ResultsAlg2} is that the results of Algorithm \ref{alg:hawkes} start poorly when the estimate of $W$ is bad, but as more and more data are revealed, the loss approaches the loss of the algorithm with full knowledge of the true matrix $W$ as predicted by Theorem \ref{thm:main}.  Additionally, the performance of Algorithm \ref{alg:hawkes} very closely mirrors the performance of using OGD to estimate $W$ directly and using that to get an estimate of the instantaneous rate using Equation \ref{eq:timediscrete}. This shows again that in the case where the influence function is known precisely, little is lost by tracking both rates and the network. Figure \ref{fig:ResultsAlg2}, plot (b), shows that the OGD algorithm almost always is incurring less loss than our method, but the gap is relatively small, on the order of $10^{-3}$ averaged over 500 time units.

\begin{figure}[t]
\centering
\subfloat[Algorithm \ref{alg:hawkes} estimate of network]{\includegraphics[height=1.35in]{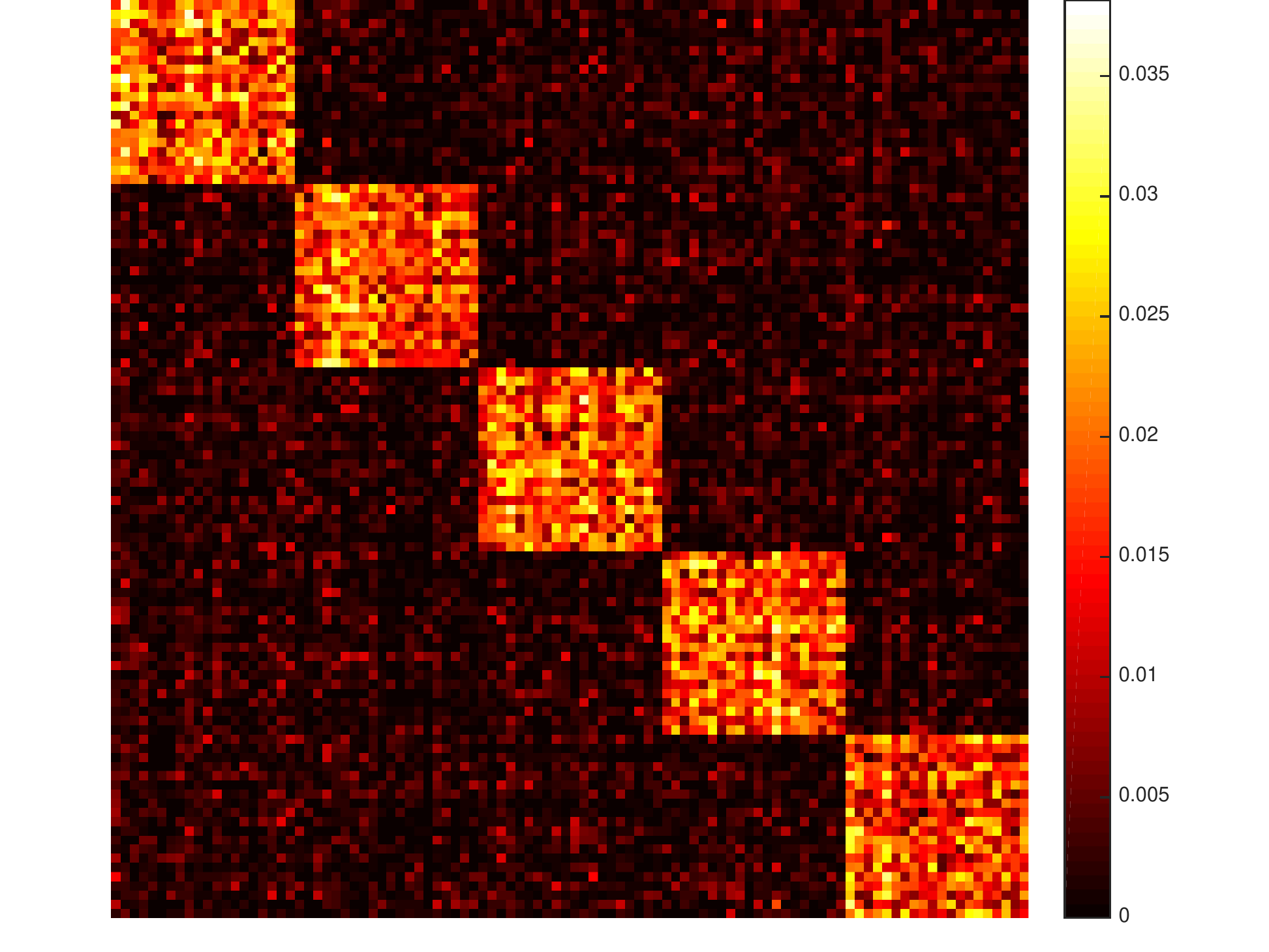}}~
\subfloat[OGD estimate of network]{\includegraphics[height=1.35in]
{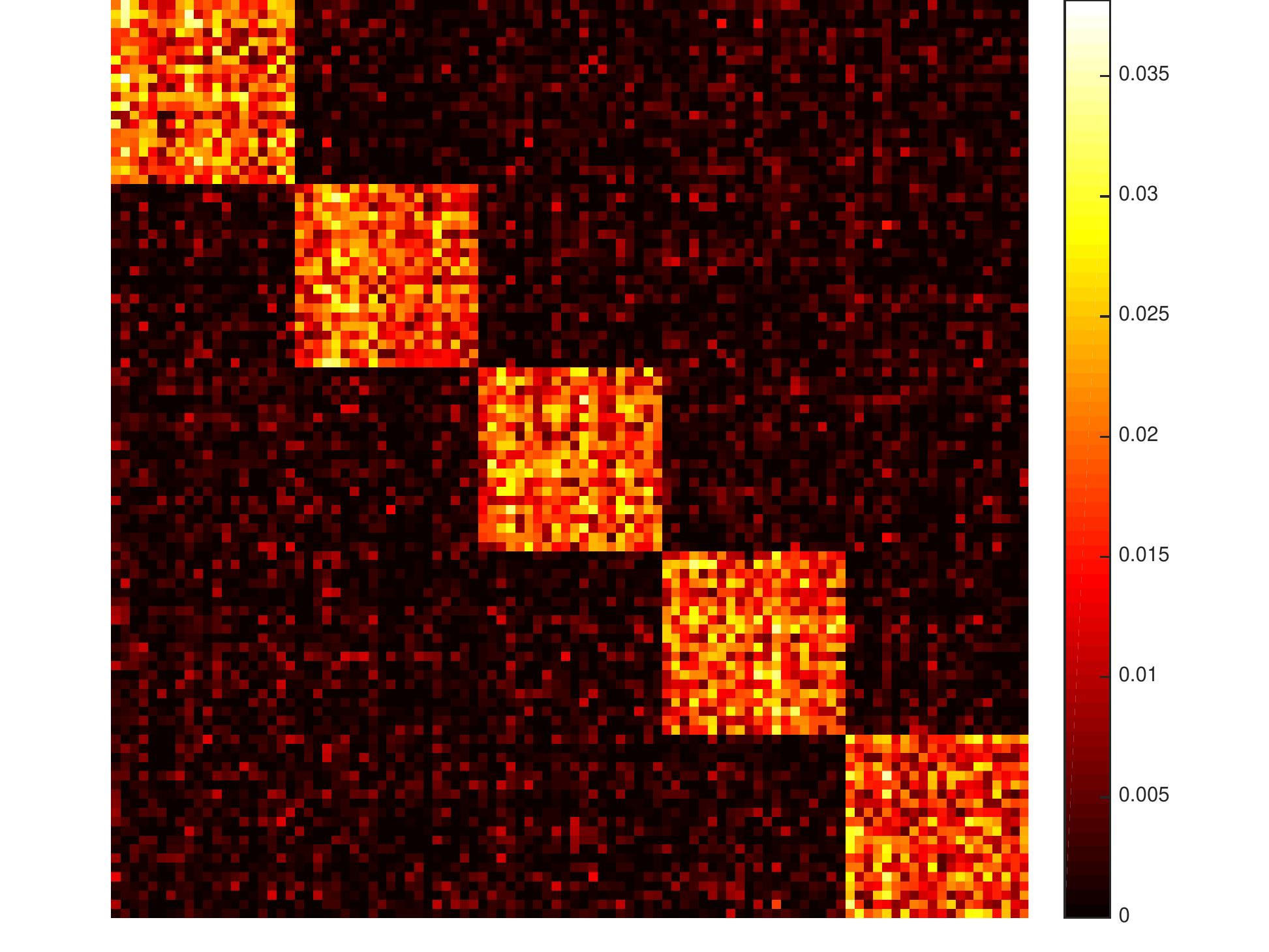}}\\
\subfloat[Algorithm \ref{alg:hawkes} estimate of network using \ref{alg:hawkes} with incorrect influence function]{\includegraphics[height=1.35 in]{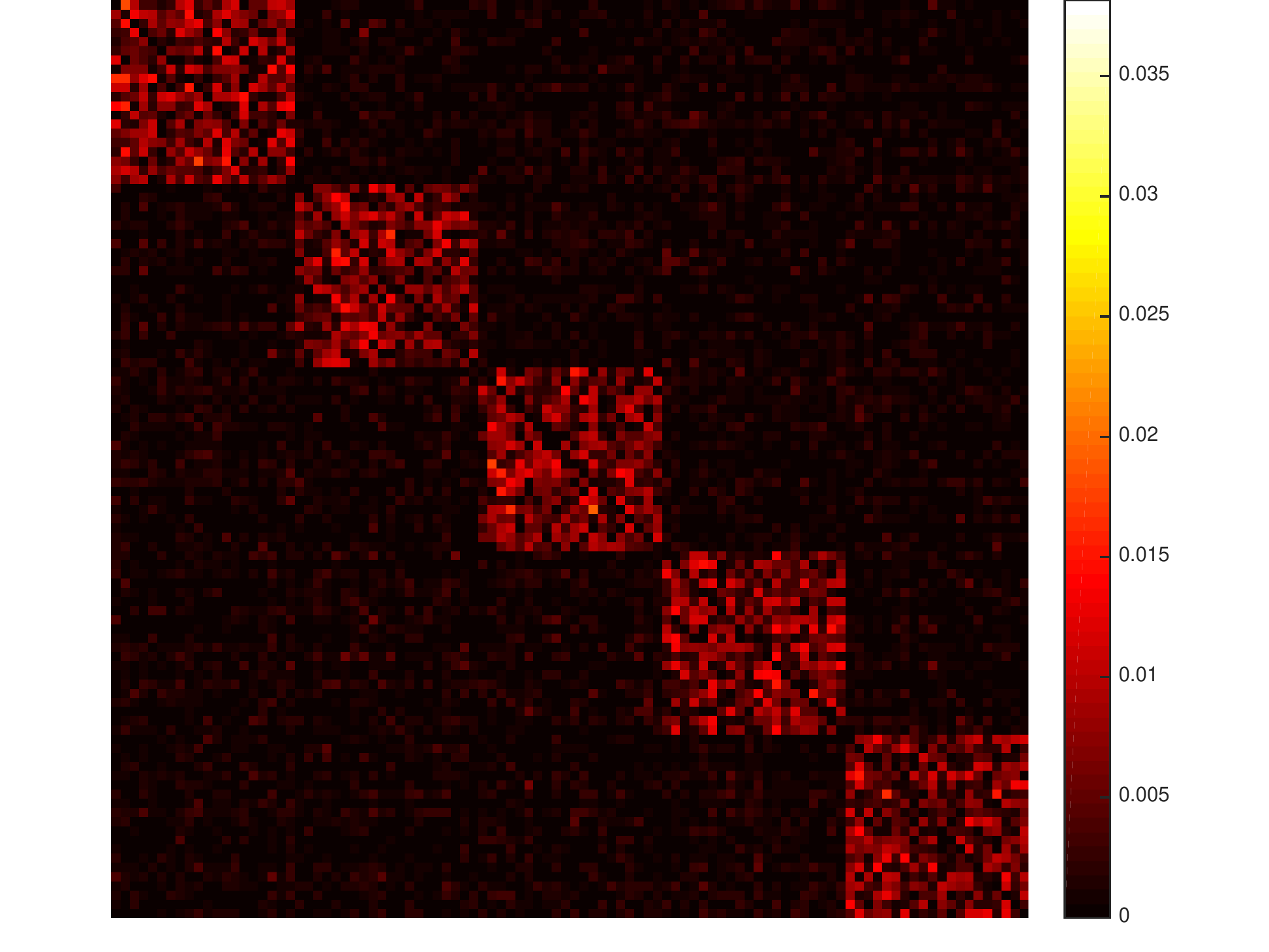}}~
\subfloat[OGD estimate of network using incorrect influence function]{\includegraphics[height=1.35 in]{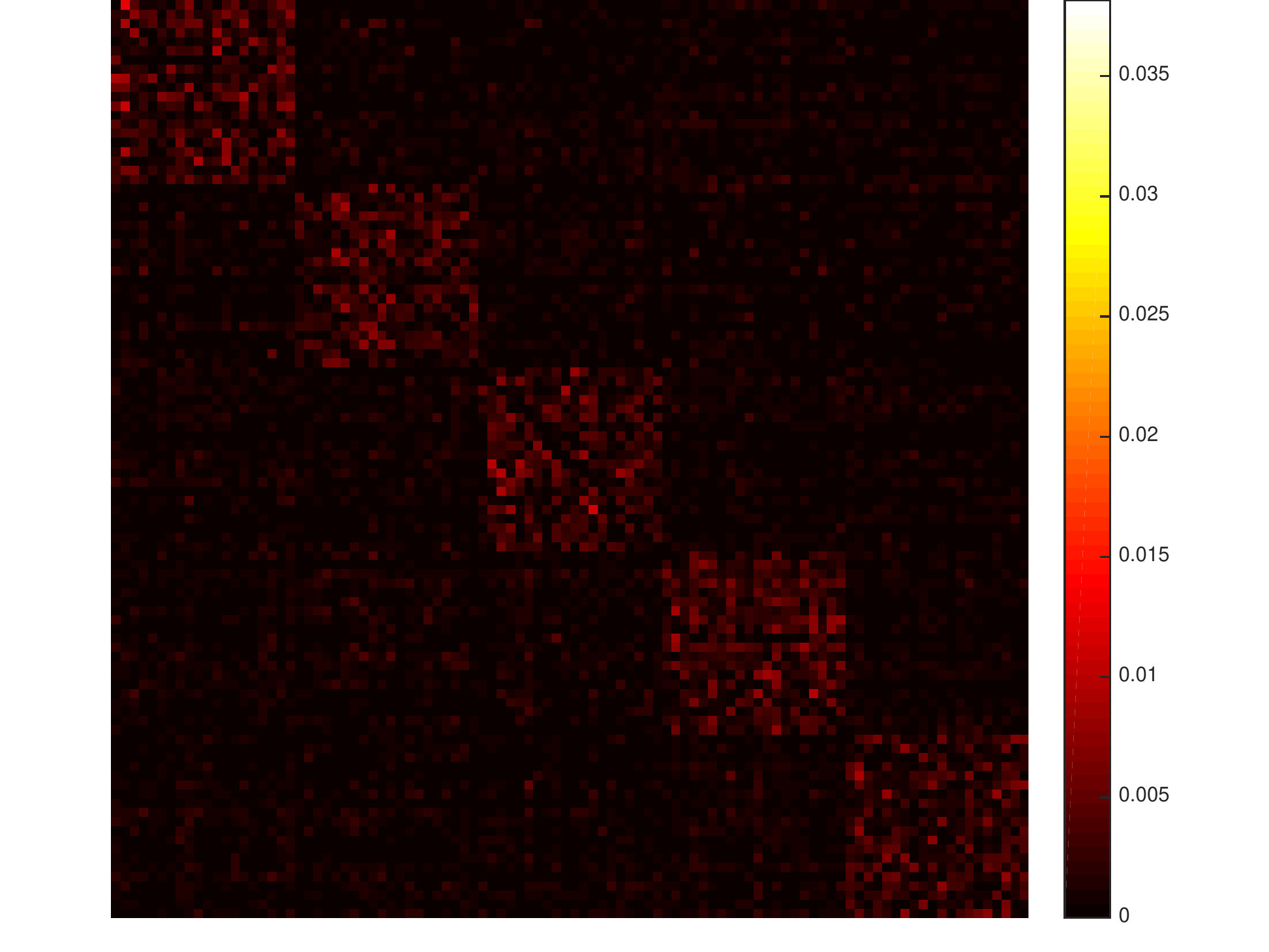}}~
\caption{Estimates of the underlying network using Algorithm \ref{alg:hawkes} and OGD with both correct (a,b) and incorrect (c,d) influence functions.  Our method captures more of the network structure when influence function is misspecified.  
}
\label{fig:ResultsAlg2_Networks}
\end{figure}

Using the same set of event times, another set of trials was run, this time using a mismatched influence function $\tilde{h}(t)=.9^{t}\ones_{[t>0]}$, and otherwise all the same parameters.  The results of these trials are shown in Figure \ref{fig:ResultsAlg2_Mismatch}.  In these results, the performance is not as accurate as when the ground truth influence function was known, but Algorithm \ref{alg:hawkes} steadily outperforms directly estimating $W$ using OGD, again demonstrating that our method has added robustness to poorly specified parameters. This time Figure \ref{fig:ResultsAlg2_Mismatch} (b) shows that our method almost always is out-performing OGD, with an average gain on the order of $10^{-1}$, thus the amount we have lost when the influence function is known is much less than the amount we have gained in the case of model mismatch.

Additionally, how well the networks have been estimated can be examined, both with the correct and incorrect influence function.  The final estimates of the networks for one data realization are shown in Figure \ref{fig:ResultsAlg2_Networks}. The estimates recover the block diagonal nature of the true network.  When the influence function is known correctly these structures are more pronounced, and the networks produced by Algorithm \ref{alg:hawkes} and OGD on $W$ are very similar.  However, when the influence function is misspecified, our method still recovers the strong clusters in the network whereas directly performing OGD on $W$ does not as obviously reveal the structure.  

\begin{figure}[t]
\centering
\subfloat[ROC Curve for Full Support]{\includegraphics[height=2.25 in]{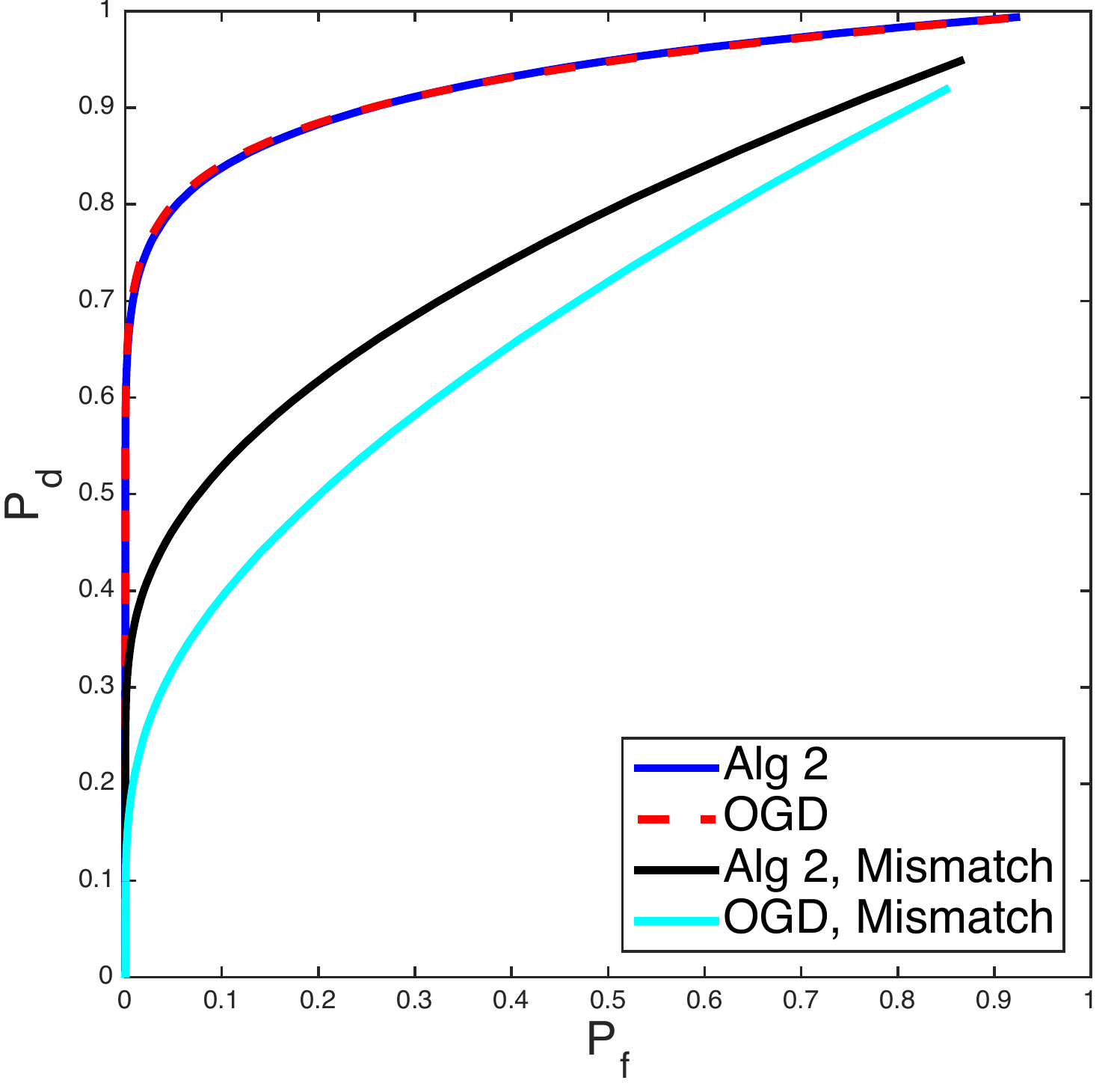}}~
\subfloat[ROC Curve for Largest 10\% Elements]{\includegraphics[height=2.25 in]{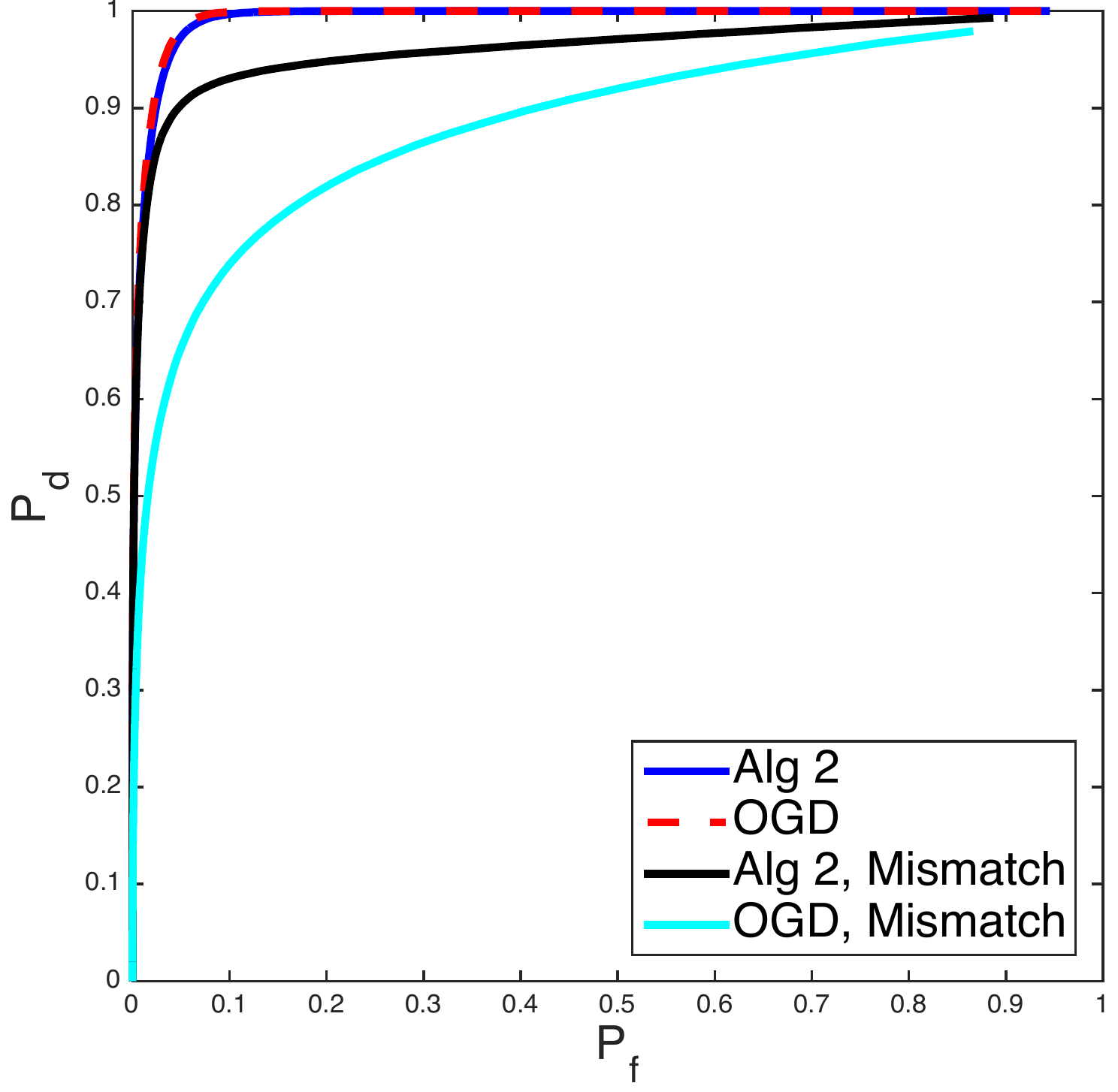}}
\caption{ROC Curves to demonstrate the method's abilities to find the significant relationships in the network.  Again, Algorithm \ref{alg:hawkes} and OGD on $W$ perform very similarly when the $h$ function is known, but Algorithm \ref{alg:hawkes} does better when it is misspecified}
\label{fig:ROC}
\end{figure}

We also observe how well the significant elements of the network are recovered by setting various thresholds and declaring all elements of the estimate above this threshold as significant relationships in the network.  These relationships are then compared with the true, non-zero elements of the network to generate ROC curves for each method.  This curve is computed both for the full $W$ matrix and just the largest 10\% elements of $W$ as baselines.  We choose to also focus on these largest edge weights as they represent the most important influences in the network.  These ROCs are shown in Figure \ref{fig:ROC}, which show our method's increased ability to find the important relations in the network compared to OGD.

\begin{figure}[t]
\centering
\includegraphics[height=2.5 in]{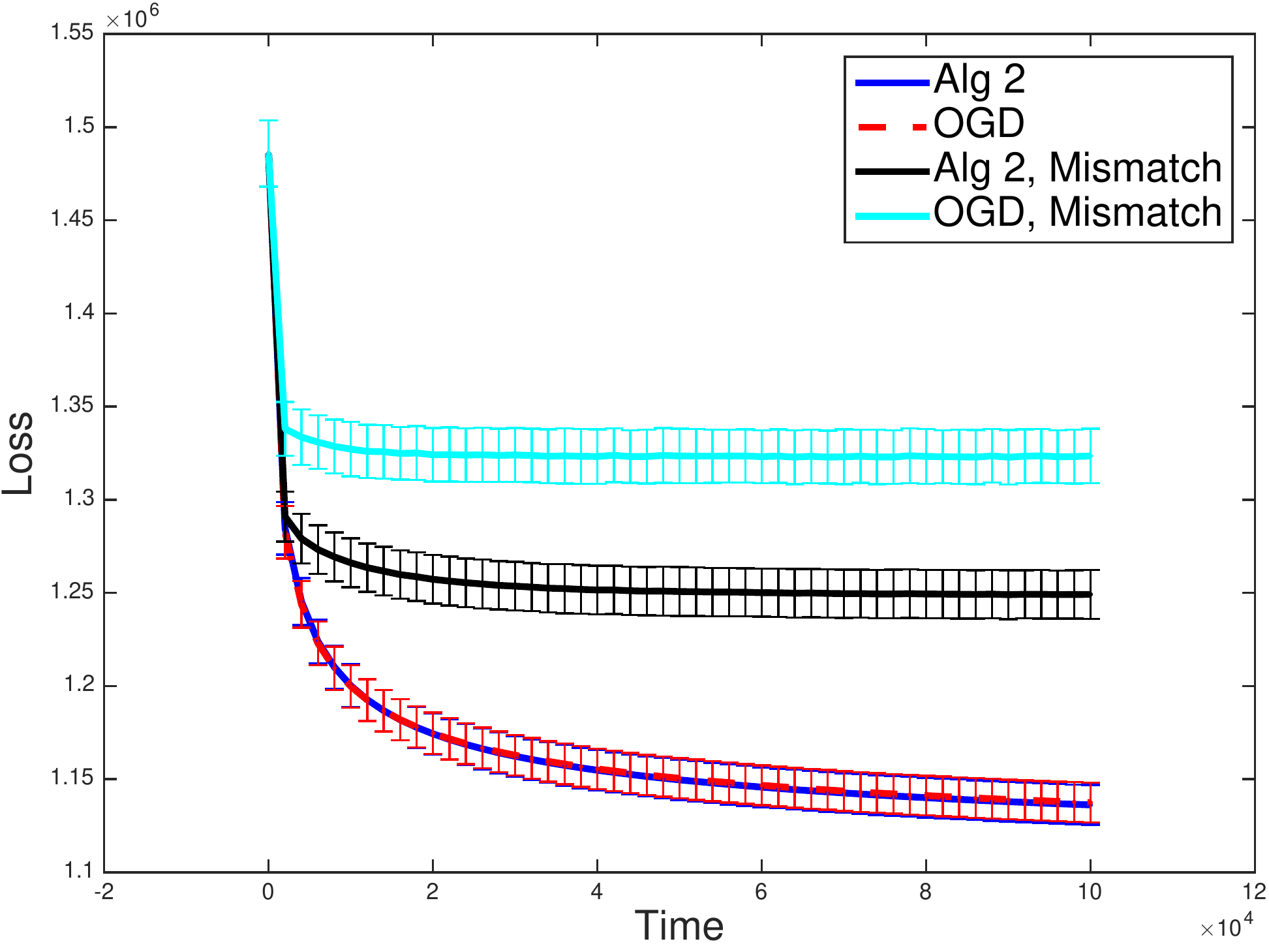}
\caption{Average of batch loss of network estimates with one standard deviation plotted in either direction.  As more data are revealed each method's performance improves.  When the influence function is known precisely our method and OGD on $W$ both perform well.  When the influence function is misspecified our method outperforms OGD.}
\label{fig:BatchLoss}
\end{figure}

As a final test of how well we are learning the matrix $W$, instantaneous estimates, $\wh{W}_t$, are used to compute the total loss of the entire data using this matrix as in Equation \ref{eq:HawkesLogLikeDisc}.  As more data are revealed, each estimate is produced with an increasing amount of training data and then tested on the full data set.  Each estimate approaches the cumulative loss of directly calculating the rates with the true matrix $W$.  These batch losses were all calculated using the true influence function, and are shown in Figure \ref{fig:BatchLoss}.  Our online method is decreasing the overall loss and approaches the same performance of knowing the true network.  Additionally, the estimation with the generative influence function is very similar for Algorithm \ref{alg:hawkes} and OGD on $W$, but our method performs better than OGD when the influence function is misspecified.

\subsection{Memetracker Data}
For the final set of experiments, we used the raw phrases Memetracker\cite{Memetracker} data set (http://www.memetracker.org/data.html) and extracted every post from websites analyzed by the authors as reporting a high percentage of important news (http://www.memetracker.org/lag.html).  These 217 distinct websites made up our network of interest.  We extracted the posts from these websites for a six month span from August 2008 through the end of January 2009, totaling over 3.5 million events.  The only information considered was what websites were posting and at what times, without using information about content or links.  The event times were on the resolution of 1 second intervals, and thus we chose $\delta=1$ sec.  

The first trial experiment was to compare the performance of our online algorithm, to the estimator created using a batch estimate, in both predictive performance and computational speed. In order to do this the first 100,000 event times were used and SpaRSA\cite{sparsa} was used to find the matrix $W$ which minimized the time averaged empirical loss plus an $\ell_1$ regularization term, using $\alpha = .99$ and a constant $\bar{\mu} = 2 \times 10^{-5}$, with regularization parameter equal to $1 \times 10^{-3}$. SpaRSA was run for a total of 60 outer iterations, with a line search inner step which performed 15 function evaluations in an inner loop. Algorithm \ref{alg:hawkes} was run with the same $\ell_1$ regularization parameter, and step-sizes $\eta_t = \frac{.01}{\sqrt{T/\delta}}$ and $\rho_t = \frac{5\times 10^{-8}}{\sqrt{T/\delta}}$. Each function evaluation in the batch setting took approximately 1.51 seconds and every gradient evaluation about 57.13 seconds, for an overall run time of approximately 80 minutes. In comparison the entire online method took 398.94 seconds due to only having to compute instantaneous losses with respect to a single second's worth of events, and only passing over each event one time. The experiments were performed using MATLAB 2015a on a laptop running Mac OSX v10.10.5 with an Intel i7 2.5 GHz processor with 16 GB of memory. Therefore the online method took only 8.3\% of the time as the computationally intensive batch method.

\begin{figure}[t]
\centering
\subfloat[Cumulative loss]{\includegraphics[height = 2.25in]{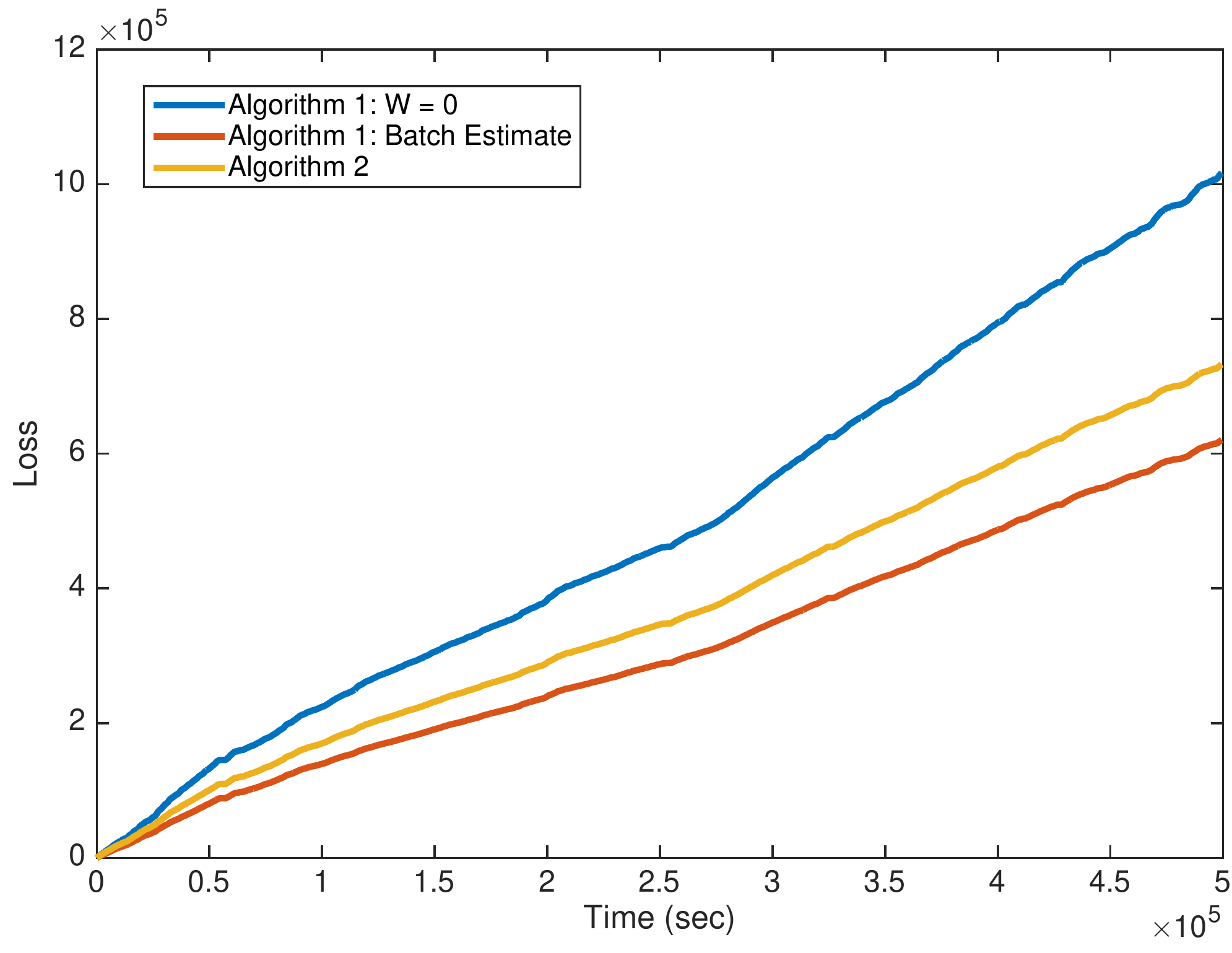}}~
\subfloat[Moving average loss with time window D = 15000 seconds]{\includegraphics[height = 2.25in]{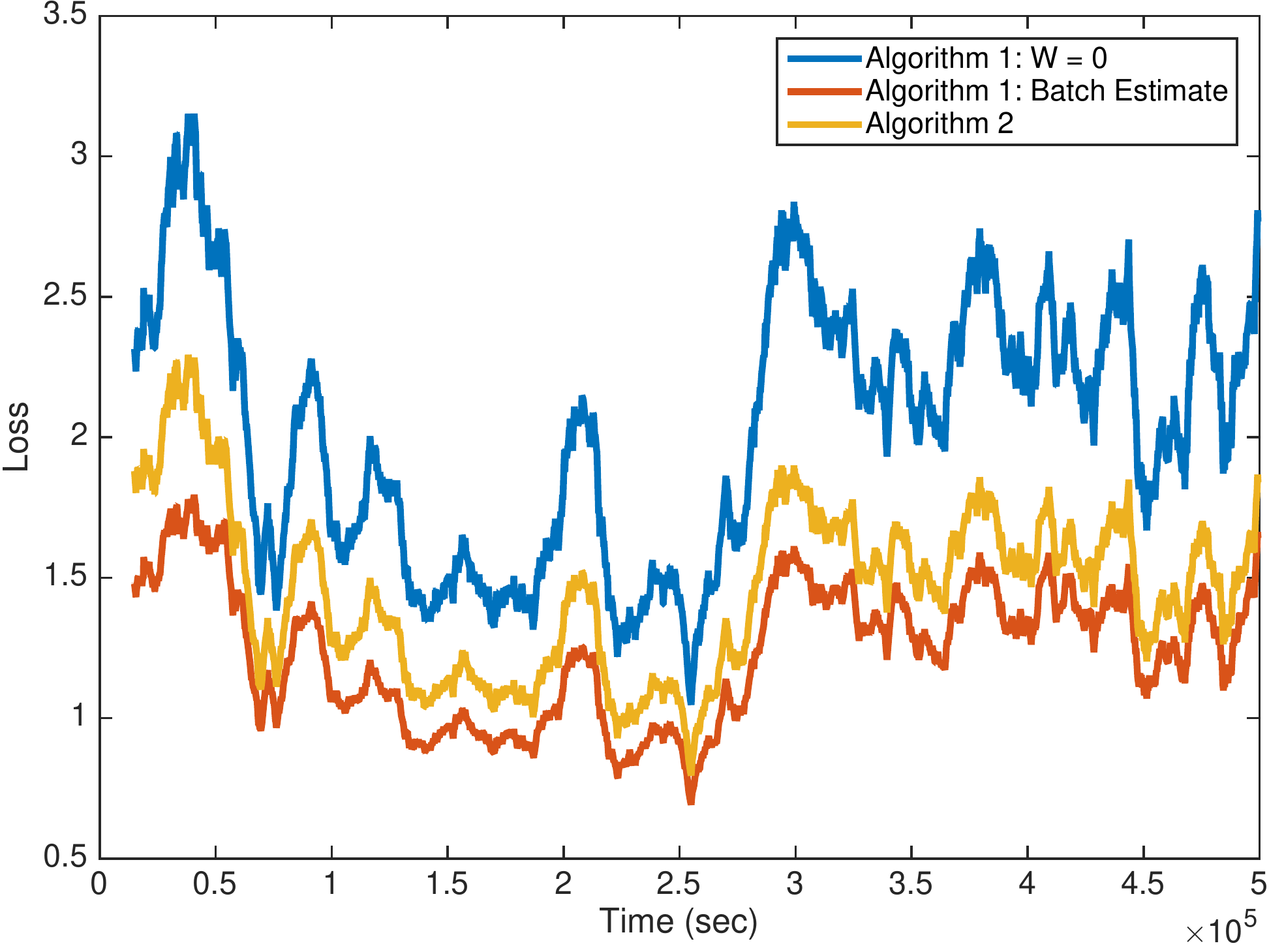}}
\caption{Comparison of Algorithm \ref{alg:hawkes} to the matrix produced using a computationally intensive batch estimate. Both the cumulative loss (a) and moving average (b) loss plots show the estimate produced by Algorithm \ref{alg:hawkes} starting poorly but rapidly approaching the performance of the batch estimate using only 8\% percent of computational time.}
\label{fig:BatchComparison}
\end{figure}

The comparison of the results of the experiment using our method and the batch method are show in Figure \ref{fig:BatchComparison}. Both the cumulative loss and moving average loss plots depict the predictive power of the online algorithm starting poorly, in relation to the batch estimate, but as more data are revealed the online algorithm approaches the performance of the batch estimate using only a fraction of the computational time. This is especially apparent in the moving average plot, which shows the initial performance of the online algorithm being close to assuming that there was no influence in the system ($W=0$) and by the end of the trial, only have a small difference from the batch estimate.

Algorithm \ref{alg:hawkes} was then run on the entire six months of data to learn influences within the network and validated with the delays discovered in the lag time of stories being reported. Running a batch method on this much data would be computationally intractable due to the poor time scaling of calculating the loss and gradient with increasing numbers of events. We used the data from the first half of the first month as a test set to tune parameters and found that $\alpha = .995, \eta=6.1\times 10^{-4}=, \rho=6.1\times 10^{-10}$ and $\bar{\mu} = 2\times 10^{-5}$ accumulated relatively low loss.  Using $\alpha=.995$, which corresponds to a reaction's ``half-life" being about 2 minutes and 20 seconds, is long enough time window for meaningful reactions to take place, but not long enough for many significantly different topics to be published.  It is worth noting this performance focuses on immediate dependencies, and thus we work on the scale of minutes, whereas the average lag times are reported on the scale of hours, and thus we are more likely to discover which organizations publish content faster, rather than finding websites which are likely using others as references.

\begin{figure}[t]
\centering
\includegraphics[height=2.5 in]{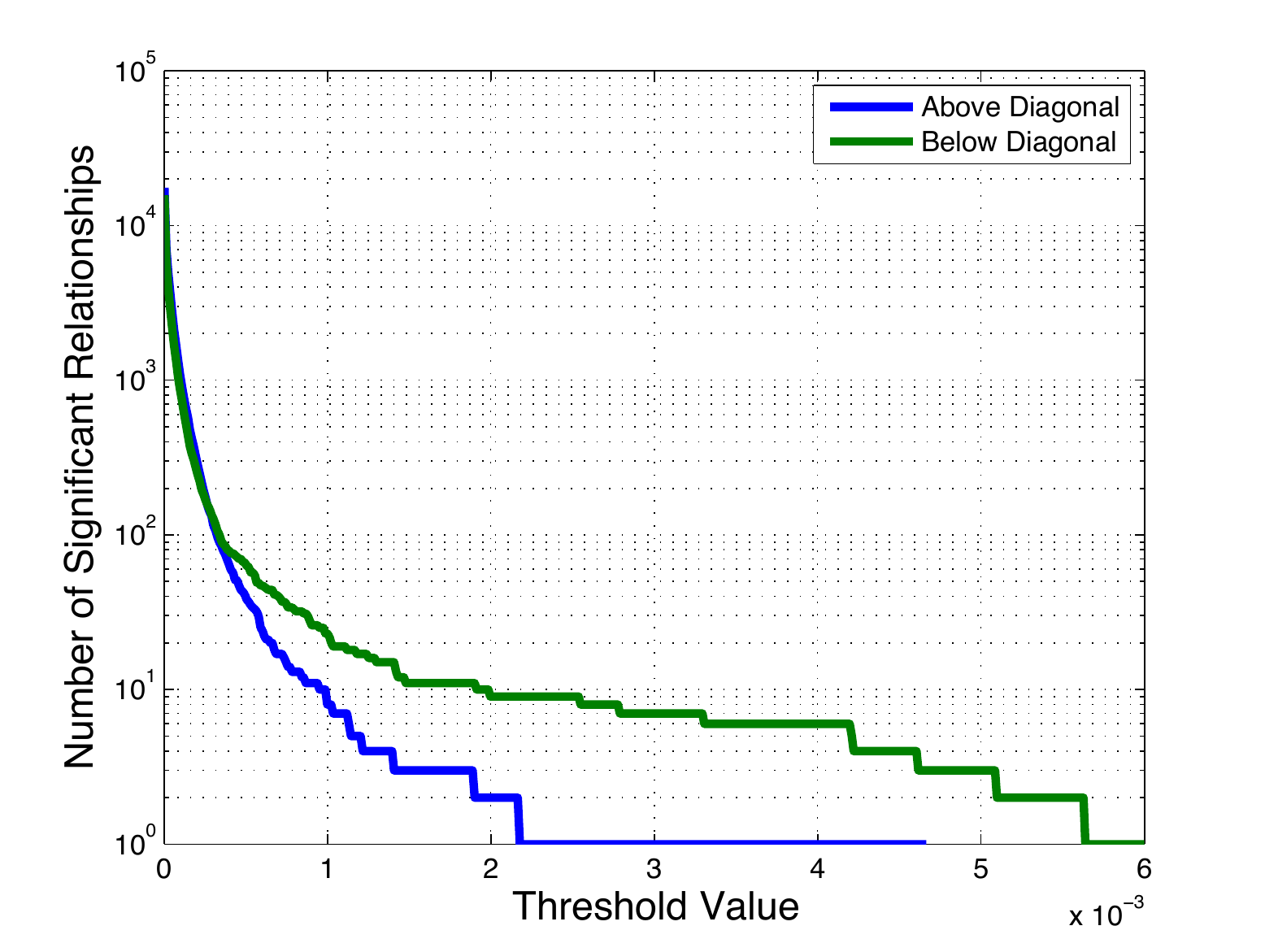}
\caption{Amount of relationships above a threshold for the network learned using the Memetracker dataset.  For most choices of the threshold, there are more relationships that are greater than the threshold above the diagonal than below.}
\label{fig:SigRels}
\end{figure}

The websites were ordered based on their average lag time described on the Memetracker results, from smallest to largest.  Therefore, we expected many of the significant relationships to be beneath the diagonal.  Recall $W_{i,j}$ reflects influence of actor $j$ on actor $i$.  Because the actors are ordered from smallest to largest lag time, we expect larger elements $W_{i,j}$ to have $i>j$, which corresponds to more significant relationships being beneath the diagonal. Relationships are declared ``significant" when $W_{i,j}$ is above a threshold.  Figure \ref{fig:SigRels} shows the number of significant relationships across a range of thresholds.  For most choices of threshold, more significant relationships are below the diagonal.  

We found that among the 20 most influential websites were dailyherald.com, washingtonpost.com, post-gazette.com, denverpost.com, news.bbc.co.uk, and cnn.com.  All of these are either local news organizations in major metropolitan areas or important national news organizations. Among the top dependencies were apnews.myway.com reacting to dailyherald.com, elections.foxnews.com reacting to washingtonpost.com and mcclatchydc.com reacting to washingtonpost.com . The first two pairs are examples of large national organizations being slower to respond than local sources.  Alternatively, the third pair is an example of two local news sources, where the organization with more journalists is able to publish faster than a competitor with less journalists.

\section{Conclusion}
In many real world applications, such as social, neural and financial
networks, actions by actors at one point in time will influence the
future actions of others in the network. In these applications it is
beneficial to estimate the likelihood of each actor acting at any
given point in time, given only timing information about previous
occurrences. This task is particularly challenging when data are
streaming at a rate that precludes traditional batch processing
methods.  We have proposed two online methods for tracking the time
varying rates at which actors participate in events; one method for
when the underlying network is known and the other for when it is
unknown. Relevant regret bounds for both methods scale with the
deviation of a comparator series of point process rate or intensity
functions, with no assumptions on the actual generative model of the
data. These methods were tested using both synthetic and real data to
show that they successfully track the intensities of interest, can
recover the underlying network structure, and are robust to model
mismatch.

\begin{appendices}

\section{Lemma \ref{lem:convexfunc}}
The proof of Lemma \ref{lem:disc_loss2}, which compares the loss of the continuous time and discrete time autoregressive rates, relies on the following relationship.
\begin{lemma} \label{lem:convexfunc} For $0<\alpha<1$ and $\delta>0$
  the following inequalities hold:
  \begin{align}
\frac{-1}{\log(\alpha)} - \frac{\delta}{2} \leq \frac{\delta \alpha^\delta}{1-\alpha^\delta} \leq \frac{-1}{\log(\alpha)}. \label{eq:lin_approx}
\end{align}
\end{lemma}
{\bf{Proof:}} The proof starts with the following observations:
\begin{align*}
\displaystyle\lim_{\delta \rightarrow 0} \frac{\delta \alpha^\delta}{1-\alpha^\delta}=&\frac{-1}{\log(\alpha)}\\
\frac{\partial}{\partial \delta} \frac{\delta \alpha^\delta}{1-\alpha^\delta}=&\frac{\alpha^\delta - \alpha^{2\delta} + \delta \alpha^\delta \log(\alpha)}{(1-\alpha^\delta)^2}.
\end{align*}
Because all three terms in Equation \ref{eq:lin_approx} are equal when $\delta=0$, showing that the derivative of $\frac{\delta \alpha^\delta}{1-\alpha^\delta}$ with respect to $\delta$ is between $-\frac{1}{2}$ and 0 suffices to prove the Lemma.  The upper bound stems from the following inequality:
\begin{align*}
&1+ \delta \log(\alpha) \leq \alpha^\delta \\
\implies& 1-\alpha^\delta +\delta \log(\alpha) \leq  0\\
\implies& \frac{\partial}{\partial \delta} \frac{\delta \alpha^\delta}{1-\alpha^\delta}=\frac{\alpha^\delta - \alpha^{2\delta} + \delta \alpha^\delta \log(\alpha)}{(1-\alpha^\delta)^2} \leq  0.
\end{align*}
The proof of the lower bound requires the analysis of another function, $\alpha^\delta - \alpha^{-\delta}$.
\begin{align*}
\frac{\partial}{\partial \delta} (\alpha^\delta - \alpha^{-\delta}) &= \log(\alpha)(\alpha^\delta + \alpha^{-\delta})\\
\frac{\partial^2}{\partial \delta^2} (\alpha^\delta - \alpha^{-\delta}) &=\log^2(\alpha) (\alpha^\delta - \alpha^{-\delta})\leq 0 {\text{ for }} \delta >0
\end{align*}
Because this function is concave for $\delta>0$, the following inequality holds:
\begin{align*}
\alpha^\delta - \alpha^{-\delta} \leq 2 \delta \log(\alpha)
\end{align*}
which simply says that for $\delta>0$ the function lies below its tangent line at $\delta=0$.  This inequality can be used to derive the desired lower bound.
\begin{align*}
&\alpha^\delta - \alpha^{-\delta} \leq 2 \delta \log(\alpha)\\
\implies& \frac{\alpha^{2\delta}-1}{2} \leq \delta \alpha^\delta \log(\alpha)\\
\implies& \frac{2\alpha^\delta - \alpha^{2\delta}-1}{2} \leq \alpha^\delta - \alpha^{2\delta}+\delta \alpha^\delta \log(\alpha)\\
\implies& -\frac{1}{2} \leq \frac{\alpha^\delta - \alpha^{2\delta} + \delta \alpha^\delta \log(\alpha)}{(1-\alpha^\delta)^2}=\frac{\partial}{\partial \delta} \frac{\delta \alpha^\delta}{1-\alpha^\delta}
\end{align*}
Therefore, the derivative of $\frac{\delta \alpha^\delta}{1-\alpha^\delta}$ is between $-\frac{1}{2}$ and 0 for $\delta>0$ and $0<\alpha<1$, which combined with the limit statement, proves the Lemma.

\section{Proof of Lemma \ref{lem:disc_loss2}}

This proof shows that the negative log likelihood of the true underlying Hawkes process (Equation \ref{eq:hawkesLogLike2}) given an excitation matrix $W$, differs by a factor proportional to $\delta$ from the discretized version (Equation \ref{eq:HawkesLogLikeDisc}).
\begin{align}
\Bigg| \sum_{k=1}^p &  \int_0^T \mu_k(\tau)d\tau - \sum_{n=1}^{N_T}\log\mu_{k_n}(\tau_n)
- \sum_{k=1}^p\Bigg(\sum_{t=1}^{T/\delta} \delta \lambda_{t,k} - 
x_{t,k} \log \lambda_{t,k}\Bigg) \Bigg| \nonumber \\
&\leq  \Bigg|\sum_{k=1}^p\Bigg( \int_0^T \mu_k(\tau) d\tau - \sum_{t=1}^{T/\delta} \delta \lambda_{t,k}\Bigg) \Bigg|
+ 
\Bigg|\sum_{k=1}^p \sum_{t=1}^{T/\delta}x_{t,k}\log\lambda_{t,k} - \sum_{n=1}^{N_t}\log \mu_{k_n}(\tau_n)\Bigg| 
\end{align}
The two absolute value terms will be bounded separately.  We start with the first term, involving the integral of the true rate, and the approximation to it.  

\begin{align*}
\Bigg|\sum_{k=1}^p\left( \int_0^T \mu_k(\tau) d\tau - \sum_{t=1}^{T/\delta} \delta \lambda_{t,k}\right) 
\leq&\sum_{k=1}^p\Bigg| \int_0^T \sum_{\tau_n<\tau} W_{k,k_n} \alpha^{\tau-\tau_n} d\tau-\delta\sum_{t=1}^{T/\delta} \sum_{\bar{\tau}_n < \delta t} W_{k,k_n} \alpha^{\delta t-\tau_n} \Bigg|\\
=&  \sum_{k=1}^p\Bigg| \sum_{n=1}^{N_T} W_{k,k_n} \int_{\tau_n}^T \alpha^{\tau-\tau_n} d\tau 
-W_{k,k_n}\delta \sum_{t=\bar{\tau}_n/\delta +1}^{T/\delta} \alpha^{\delta t- \tau_n}  \Bigg|\\
=& \sum_{k=1}^p \Bigg|\sum_{n=1}^{N_T} W_{k,k_n} \left(\frac{\alpha^{T-\tau_n}-1}{\log(\alpha)} -  
\delta \alpha^{\delta}\frac{\alpha^{\bar{\tau}_n-\tau_n}-\alpha^{T-\tau_n}}{1-\alpha^\delta} \right)\Bigg|\\
=& \sum_{k=1}^p \Bigg| \sum_{n=1}^{N_T}W_{k,k_n} \Bigg( \alpha^{T-\tau_n} \left(\frac{\delta \alpha^{\delta}}{1-\alpha^\delta} - \frac{-1}{\log(\alpha)}\right)\\
 &\hspace{1.5 in}+ \frac{-1}{\log(\alpha)} - \alpha^{\bar{\tau}_n-\tau_n} \frac{\delta \alpha^\delta}{1-\alpha^\delta} \Bigg) \Bigg| 
\end{align*}

These lines evaluate the value of the integral and the approximation to them using a geometric sum to get a closed form for their difference.  From this point, the use of Lemma \ref{lem:convexfunc} allows us to find an upper bound
\begin{align*}
 \alpha^{T-\tau_n} &\left(\frac{\delta \alpha^{\delta}}{1-\alpha^\delta} - \frac{-1}{\log(\alpha)}\right) + \frac{-1}{\log(\alpha)} - \alpha^{\bar{\tau}_n-\tau_n} \frac{\delta \alpha^\delta}{1-\alpha^\delta} \\
 \leq& \frac{-1}{\log(\alpha)} - \alpha^{\delta} \frac{\delta \alpha^\delta}{1-\alpha^\delta} \leq
 \frac{\delta}{2} + (1-\alpha^\delta)\frac{\delta \alpha^\delta}{1-\alpha^\delta} \leq \frac{3\delta}{2}.
\end{align*}
We can similarly use Lemma \ref{lem:convexfunc} to find the lower bound.
\begin{align*}
\alpha^{T-\tau_n} &\left(\frac{\delta \alpha^{\delta}}{1-\alpha^\delta} - \frac{-1}{\log(\alpha)}\right) + \frac{-1}{\log(\alpha)} - \alpha^{\bar{\tau}_n-\tau_n} \frac{\delta \alpha^\delta}{1-\alpha^\delta} \\
\geq & \frac{-\delta}{2} \alpha^{T-\tau_n} + \frac{-1}{\log(\alpha)} - \frac{\delta \alpha^\delta}{1-\alpha^\delta} \geq -\frac{\delta}{2}
\end{align*}
Combining these upper and lower bounds gives an overall bound on the integral approximation:
\begin{align}
 \Bigg|\sum_{k=1}^p\left( \int_0^T \mu_k(\tau) d\tau - \sum_{t=1}^{T/\delta} \delta \lambda_{t,k}\right) \Bigg| \leq pN_T W_{\max} \frac{3\delta}{2} \label{eq:int_approx}
 \end{align}
This shows that the integral terms deviate by no more than a linear factor of $\delta$ and can be controlled by setting $\delta$ small.  

Next the difference involving the log terms must be bounded.  We make the following observation:
\begin{align*}
\sum_{k=1}^p\sum_{t=1}^{T/\delta} x_{t,k} \log \lambda_{t,k} = \sum_{n=1}^{N_{T}} \log \lambda_{\bar{\tau}_n/\delta,k_n}
\end{align*}
This says that instead of going from time step to time step and incurring loss at every point, we just step through every event and incur loss at the following discrete time point $\bar{\tau}_n$.  Now the log terms can be compared as
\begin{align*}
\sum_{k=1}^p \sum_{t=1}^{T/\delta}x_{t,k}\log\lambda_{t,k} - \sum_{n=1}^{N_T}\log \mu_{k_n}(\tau_n)
=  \sum_{n=1}^{N_{T}} \log \left(\frac{\lambda_{\bar{\tau}_n/\delta,k_n}}{\mu_{k_n}(\tau_n)}\right).
\end{align*}

In order to bound this, we make the distinction between two classes of events.  The first set of events, $\mathcal{A}_n = \{i| \tau_i < \tau_n, \bar{\tau}_i < \bar{\tau}_n\}$, are events that happen before the $n^{th}$ event and are not in the same time window.  The second set, $\mathcal{B}_n = \{i| \tau_i < \tau_n, \bar{\tau}_i = \bar{\tau}_n\}$, are events that happen before the $n^{th}$ event but in the same time window. 
\begin{align*}
 &\sum_{n=1}^{N_{T}} \log \left(\frac{\lambda_{\bar{\tau}_n/\delta,k_n}}{\mu_{k_n}(\tau_n)}\right)\\=
 &\sum_{n=1}^{N_{T}} \log \left(\frac{\bar{\mu}_{k_n} + \displaystyle\sum_{i \in \mathcal{A}_n} W_{k_n,k_i}\alpha^{\bar{\tau}_n - \tau_i}}{\bar{\mu}_{k_n} + \displaystyle\sum_{i\in\mathcal{A}_n} W_{k_n,k_i}\alpha^{\tau_n - \tau_i} + \sum_{i \in \mathcal{B}_n} W_{k_n,k_i} \alpha^{\tau_n - \tau_i}} \right) 
\end{align*}
This term needs to be upper and lower bounded to get the final result. Because every element of $W\geq 0$ and $\alpha>0$, all the terms are positive so the terms in set $\mathcal{B}_n$ can be dropped, to get the following upper bound:
\begin{subequations}
\begin{align}
\sum_{n=1}^{N_{T}} \log &\left(\frac{\lambda_{\bar{\tau}_n/\delta,k_n}}{\mu_{k_n}(\tau_n)}\right)\leq \sum_{n=1}^{N_{T}}\log\left( \frac{\bar{\mu}_{k_n} + \displaystyle \sum_{i \in \mathcal{A}_n} W_{k_n,k_i} \alpha^{\bar{\tau}_n - \tau_i}}{\bar{\mu}_{k_n} + \displaystyle\sum_{i \in \mathcal{A}_n} W_{k_n,k_i} \alpha^{\tau_n - \tau_i}}\right)\label{eq:pf1}\\
&\leq \sum_{n=1}^{N_{k}} \log\left( \frac{\bar{\mu}_{k_n} + \displaystyle \sum_{i \in \mathcal{A}_n} W_{k_n,k_i} \alpha^{\tau_n-\tau_i}}{\bar{\mu}_{k_n} + \displaystyle\sum_{i \in \mathcal{A}_n} W_{k_n,k_i} \alpha^{\tau_n - \tau_i}}\right) = 0. \label{eq:pf2}
\end{align}
\end{subequations}
In Equation \ref{eq:pf1} positive terms have been removed from the denominator and in \ref{eq:pf2} the fact that $\bar{\tau}_n\geq \tau_n$ and therefore $\alpha^{\bar{\tau}_n} \leq \alpha^{\tau_n}$. Next, we lower bound $T_1$.  
\begin{subequations}
\begin{align}
&\sum_{n=1}^{N_{T}} \log \left(\frac{\lambda_{\bar{\tau}_n/\delta,k_n}}{\mu_{k_n}(\tau_n)}\right)\nonumber\\
\geq &\sum_{n=1}^{N_{T}}\log \left( \frac{\bar{\mu}_{k_n} + \displaystyle\sum_{i \in \mathcal{A}_n} W_{k_n,k_i} \alpha^{\bar{\tau}_n - \tau_i}}{\bar{\mu}_k + \displaystyle \sum_{i \in \mathcal{A}_n} W_{k_n,k_i}\alpha^{\bar{\tau}_n - \tau_i - \delta} + \displaystyle \sum_{i \in \mathcal{B}_n} W_{k_n,k_i} \alpha^{\tau_n-\tau_i}}\right)\label{eq:pf3}\\
 \geq& -\sum_{n=1}^{N_T} \log \left(\alpha^{-\delta} + \frac{|\mathcal{B}_n| W_{\max}}{\mu_{\min}} \right)\label{eq:pf4}\\
 \geq& -\sum_{n=1}^{N_T} \log \left(\alpha^{-\delta} + \frac{p x_{\max} \delta W_{\max}}{\mu_{\min}}\right)\label{eq:pf5}\\
 \geq & -N_T \left(\frac{W_{\max}x_{\max} p}{\mu_{\min}} - \log(\alpha)\right) \delta \label{eq:pf6}
\end{align}
\end{subequations}
Equation \ref{eq:pf3} comes from $\alpha^{\bar{\tau}_n - \delta} \geq \alpha^{\tau_n}$, Equation \ref{eq:pf4} uses $\alpha^{-\delta}>1$ and cancels out like terms from the numerator and denominator, Equation \ref{eq:pf5}
bounds the number of events in $\mathcal{B}_n$ by $p x_{\max} \delta$ which is the maximum number of events each actor can participate in a $\delta$ length time window, times the number of actors, and finally Equation \ref{eq:pf6} uses $\log(x+a) \leq \frac{x}{a} + \log(a)$ for $a>0$ which is a consequence of the concavity of the $\log$ function.  The upper and lower bound gives
$$ \left|\sum_{n=1}^{N_{T}} \log \left(\frac{\lambda_{\bar{\tau}_n/\delta,k_n}}{\mu_{k_n}(\tau_n)}\right)\right| \leq N_T \left(\frac{W_{\max} x_{\max} p}{\mu_{\min}} - \log(\alpha) \right) \delta$$
which when combined with Equation \ref{eq:int_approx} gives the result:
\begin{align*}
|L_T(\mu) -& L_T^{(\delta)}(\lambda)| \\
&\leq \left(\frac{3pW_{\max}}{2} + \frac{W_{\max}x_{\max}p}{\mu_{\min}} - \log(\alpha)\right) N_T \delta.
\end{align*}

\section{Dual Parameterization}
We introduce a change of variables which will allow us to more easily bound the regret of the proposed algorithms. Definine $\theta_t =[\theta_{t,1},...,\theta_{t,p}]\trans \triangleq \log(\delta \lambda_t).$  Using this change of variable gives a loss function in terms of $\theta$.
\begin{align}
\tilde{\ell}_t(\theta_t) = \ell_t\left(\frac{1}{\delta}\exp(\theta_t)\right) = -\langle \theta_t,x_t \rangle + \langle \exp(\theta_t),\ones \rangle \label{eq:losses2}.
\end{align}

It is important to note that this is a one-to-one relationship between $\lambda_t$ and $\theta_t$, and thus we may operate in either the $\lambda$ or $\theta$ space.  Notice that Equation \ref{eq:losses2} corresponds to the negative log-likelihood of an exponential family distribution of the form
$$p_t(\theta) = \exp\left\{\ave{\theta,x_t} - Z(\theta)\right\}$$
where we omit factors depending only on $x_t$ and not $\theta$, and where
$$Z(\theta) \deq \log \int \exp\{\ave{\theta,x} \} dx$$
is the so-called {\em log-partition function}. Important to our analysis is the dual of the log-partition function, $Z^*(\delta \lambda) \triangleq \displaystyle \sup_{\theta \in \THETA}\{ \langle \delta \lambda, \theta\rangle - Z(\theta)\}$. In our multivariate Hawkes setting, we have
$Z(\theta) = \ave{\exp(\theta),\ones}$.  Performing online optimization in the $\theta$ parameter space allows us to exploit several properties of exponential families, as described in general settings in \cite{raginsky_OCP} and in our specific context in Section \ref{sec:W_unknown}.  In particular, we will use the following facts:
\begin{align*}
\nabla Z(\theta) =& \exp(\theta)=\delta\lambda\\
\nabla Z^*(\delta\lambda) =& \log (\delta\lambda) = \theta
\end{align*}
An equivalent dynamical model from the $\lambda$ space can be defined in the $\theta$ parameter space as
\begin{equation}
\tilde{\Phi}_t(\theta,W) = \nabla Z^*(\delta\Phi_t(\nabla Z(\theta)/\delta,W)).
\label{eq:dyn2}
\end{equation}
which essentially converts the dual ($\delta\lambda$) to the primal ($\theta$), applies the dynamics, then converts back. Using the dual parameterization, we have the following equivalent to Algorithm \ref{alg:W_Known}, which uses the function $D(\theta_1 \| \theta_2) \deq Z(\theta_1) - Z(\theta_2) - \langle \grad Z(\theta_2), \theta_1 - \theta_2 \rangle$, which is the Bregman divergence induced by $Z(\theta)$. 
\begin{algorithm}
\caption{MV Hawkes Tracking - $W$ Known, dual parameterization}
\begin{algorithmic}[1] 
\label{alg:dual}
\STATE Initialize $\hat{\theta}_1=\log(\delta \bar{\mu})$
\FOR {$t=1,...,T/\delta$}
\STATE Observe $x_t$ and incur loss $\tilde{\ell}_t(\htheta_t) = \langle \ones \exp(\htheta_t) \rangle -  \langle x_t, \theta_t \rangle$
\STATE Set $\ttheta_{t+1} = \displaystyle \argmin_{\theta \in \Theta} \eta_t \langle \grad \tilde{\ell}_t(\htheta_t), \theta_t \rangle + D(\theta\|\htheta_t)$
\STATE Set $\htheta_{t+1} = \tilde{\Phi}_t(\ttheta_{t+1}, W)$
\ENDFOR
\end{algorithmic}
\end{algorithm}

We list some important properties of the loss function $\tilde{\ell}_t(\theta)$ that will be used in the proof of the regret bounds, in section \ref{sec:algorithms}.
\begin{itemize}
\item We assume a convex set of possible rate functions $\lambda \in [\lambda_{\min}, \lambda_{\max}]^p = \Lambda$ with $\lambda_{\min}>0$ and therefore the corresponding dual space $\Theta = [\log(\delta \lambda_{\min}), \log(\delta \lambda_{\max})]$ .
\item Because we assume that there is a maximum number of times each actor can act per unit time, $x_{\max}$, and the observations space $\cX$ is simply $[0,\delta x_{\max}]^p$, we additionally have that on the set $\Theta$:
\begin{align}
\|\grad \tilde{\ell}_t(\theta)\|_2 \leq& \|\exp(\theta)\|_2+\|x_t\|_2 \nonumber\\
\leq&\sqrt{p} \delta(\lambda_{\max}+ x_{\max})
\label{eq:gradLipschitz}
\end{align}
\item The function $Z(\theta)=\langle \ones, \exp(\theta)\rangle$ is strongly convex on $\Theta$ with strong convexity parameter $\delta \lambda_{\min}$ with respect to the $\ell_2$ norm.  Therefore the Bregman divergence induced by $Z$ obeys the following:
\begin{align}
D(\theta_1\|\theta_2) =& Z(\theta_1) - Z(\theta_2) - \langle \grad Z(\theta_2),\theta_1 - \theta_2\rangle \nonumber\\
\geq & \frac{\delta \lambda_{\min}}{2}\|\theta_1 - \theta_2 \|^2_2 \label{eq:BregStrong}
\end{align}
\item The Bregman divergence induced by $Z(\theta)=\langle\ones,\exp(\theta)\rangle$ is always non-negative and can be upper bounded for any $\theta_1,\theta_2 \in \Theta$:
\begin{align}
D(\theta_1\|\theta_2) =& Z(\theta_1) - Z(\theta_2) - \langle \grad Z(\theta_2),\theta_1-\theta_2 \rangle\nonumber\\
\leq & \langle \grad Z(\theta_1) - \grad Z(\theta_2), \theta_1-\theta_2 \rangle \nonumber\\
\leq & \|\grad Z(\theta_1) - \grad Z(\theta_2)\|_2 \|\theta_2 - \theta_2\|_2 \nonumber\\
= & \delta \| \lambda_1 - \lambda_2 \|_2 \left\|\log\left(\frac{\lambda_1}{\lambda_2}\right)\right\|_2 \nonumber\\
\leq & \frac{\delta \lambda_{\max}^2 p}{\lambda_{\min}} \label{eq:D_max}
\end{align}
\item A property of Bregman divergences and dual functions says the following:
\begin{align*}
D(\theta_1 \| \theta_2) = &Z(\theta_1) - Z(\theta_2) - \langle \grad Z(\theta_2), \theta_1 - \theta_2 \rangle \\
 =& Z^*(\delta \lambda_2) - Z^*(\delta \lambda_1) - \langle \grad Z^*(\delta \lambda_1), \delta \lambda_2 - \delta \lambda_1 \rangle\\ \triangleq &D^*(\delta \lambda_2\|\delta \lambda_1)
\end{align*}
Therefore, $\tilde{\Phi}$ is contractive with respect to $D$ if and only if $\Phi$ is contractive with respect to $D^*$.
\begin{align*}
D(\tPhi(\theta_1, W) \| \tPhi(\theta_2, W)) =& D^*( \delta\Phi_t ( \lambda_2, W) \|  \delta\Phi_t(\lambda_1, W))\\
 \leq& D^*(\delta \lambda_2 \| \delta \lambda_1) = D(\theta_1 \| \theta_2)
\end{align*}
\end{itemize}

\section{Proof of Lemma \ref{lem:contractive}}\label{app:contractive}
We prove the lemma by proving the result in the $\theta$ space described in the previous section, and by the properties of duals of Bregman divergences, the result holds for the $\lambda$ space as well. We start the proof with the following important observation about the Bregman divergence in question:
\begin{align*}
D(\theta_1\|\theta_2) =& \langle \exp(\theta_1)-\exp(\theta_2),\ones\rangle - \langle \exp(\theta_2),\theta_1-\theta_2 \rangle\\
= & \sum_{k=1}^p \exp(\theta_{1,k})-\exp(\theta_{2,k}) - \exp(\theta_{2,k})(\theta_{1,k}-\theta_{2,k})\\
=&\sum_{k=1}^p d(\theta_{1,k}\|\theta_{2,k}).
\end{align*}
Above, $\theta_{1,k}$ and $\theta_{2,k}$ denote the $k^{th}$ element of vectors $\theta_1$ and $\theta_2$ respectively, and $d$ is the scalar Bregman divergence induced by $\exp(\theta)$. This shows that the $p$-dimensional Bregman divergence can be broken into the sum of $p$ terms.  We will prove bounds for the one dimensional version and then combine them to show that overall the dynamics are contractive. Because $\Phi_t(\lambda,W) = A_t \lambda + b_t$, we therefore have $\tPhi(\theta,W) = \log ( A_t \exp(\theta) + \delta b_t)$.

We start by showing the result for when the $k^{th}$ diagonal element of the matrix $A_t$, denoted by $A_{t,k}$ is 0. We denote the $k^{th}$ element of the vector after the application of the dynamics as $[\Phi_t(\theta)]_k$ and the $k^{th}$ element of $b_t$ as $b_{t,k}$.  
\begin{align*}
d(&[\Phi_t(\theta_{1})]_k\|[\Phi_t(\theta_{2})]_k)= A_{t,k}( \exp(\theta_{1,k}) - \exp(\theta_{2,k})) \\
&- (A_{t,k}\exp(\theta_{2,k}) + \delta b_{t,k}) \log \left( \frac{A_{t,k}\exp(\theta_{1,k}) + \delta b_{t,k}}{A_{t,k}\exp(\theta_{2,k}) + \delta b_{t,k}} \right)\\
=&-\delta b_{t,k}\log \left( \frac{b_{t,k}}{b_{t,k}} \right) = 0 =A_{t,k} d(\theta_{1,k}\|\theta_{2,k})
\end{align*}
When $A_{t,k}=b_{t,k}=0$ we define $\log(\frac{0}{0})\triangleq 0$. Next, we show the result for the case when $A_{t,k}>0$ and $b_{t,k}=0$:
\begin{align*}
d([\Phi_t&(\theta_{1})]_k\|[\Phi_t(\theta_{2})]_k) = A_{t,k}(\exp(\theta_{1,k}) - \exp(\theta_{2,k}))\\
&- A_{t,k}\exp(\theta_{2,k}) \log\left(\frac{A_{t,k}\exp(\theta_{1,k})}{A_{t,k}\exp(\theta_{2,k})} \right)\\
=&A_{t,k}\left( \exp(\theta_{1,k})-\exp(\theta_{2,k}) - \exp(\theta_{2,k})(\theta_{1,k}-\theta_{2,k})\right)\\
=&A_{t,k} d(\theta_{1,k}\|\theta_{2,k}).
\end{align*}
Finally, we show that the one dimensional Bregman is non-increasing in $b_{t,k}$ for $A_{t,k}>0$.
\begin{align*}
&\frac{d([\Phi_t(\theta_{1})]_k\|[\Phi_t(\theta_{2})]_k)}{d b_{t,k}} =-\delta \log \left( \frac{A_{t,k}\exp(\theta_{1,k}) + \delta b_{t,k}}{A_{t,k}\exp(\theta_{2,k}) + \delta b_{t,k}} \right)\\
&-\delta \left(\frac{A_{t,k}\exp(\theta_{2,k})+\delta b_{t,k}}{A_{t,k}\exp(\theta_{1,k})+\delta b_{t,k}} - \frac{A_{t,k}\exp(\theta_{2,k})+\delta b_{t,k}}{A_{t,k}\exp(\theta_{2,k})+\delta b_{t,k}}\right)\\
=&\delta \log \left( \frac{A_{t,k}\exp(\theta_{2,k}) +\delta b_{t,k}}{A_{t,k}\exp(\theta_{1,k}) + \delta b_{t,k}} \right)\\
&+\delta \left(1 - \frac{A_{t,k}\exp(\theta_{2,k})+\delta b_{t,k}}{A_{t,k}\exp(\theta_{1,k})+\delta b_{t,k}}\right) \leq 0
\end{align*}
The final inequality comes from the fact that $1-x \leq -\log x$.  The result of this is that we have shown that when $b_{t,k} > 0$ the one dimensional Bregman divergence is less than if $b_{t,k}=0$.  Combining all the results with the assumption that all the elements of the diagonal matrix $A_t$ are upper bounded by one, gives the conclusion that these dynamics are contractive.
\begin{align*}
D(\Phi_t&(\theta_1)\|\Phi_t(\theta_2)) = \sum_{k=1}^p d([\Phi_t(\theta_{1})]_k\|[\Phi_t(\theta_{2})]_k)\\
\leq & \sum_{k=1}^p A_{t,k} d(\theta_{1,k}\|\theta_{2,k}) \leq
\sum_{k=1}^p d(\theta_{1,k}\|\theta_{2,k}) = D(\theta_1\|\theta_2)
\end{align*}

\section{Proof of Theorem \ref{thm:W_known}}\label{app:pf_W_Known}
The proof of Theorem \ref{thm:W_known} is based on the proof of Theorem 3 of \cite{DMD_journal}, specialized to the Hawkes process.  The strategy is to bound the excess loss at any given moment, and then add all of these bounds from $t=1$ to $T/\delta$.  Importantly we use the fact that $\ell_t(\wh{\lambda}_t) = \tilde{\ell}_t(\htheta_t)$ and $\ell_t(\lambda_t)=\tilde{\ell}_t(\theta_t)$.  We start with some important properties.  The first is the first order optimality condition of line 4 of Algorithm \ref{alg:dual}, which states, for any $\theta \in \Theta$ we have:
\begin{align*}
\langle \eta_t \grad \tilde{\ell}_t(\hat{\theta}_t) + \grad Z (\tilde{\theta}_{t+1}) - \grad Z(\hat{\theta}_t), \tilde{\theta}_{t+1} - \theta \rangle \leq 0.
\end{align*}
By rearranging terms we get the form that is used.
\begin{align}
\langle \tilde{\ell}_t (\hat{\theta}_t),\tilde{\theta}_{t+1}-\theta_t \rangle \leq \frac{1}{\eta_t} \langle Z (\hat{\theta}_t )- \grad Z (\tilde{\theta}_{t+1}),\tilde{\theta}_{t+1}-\theta_t \rangle \label{eq:opt}
\end{align}

The second important fact is that a Bregman divergence induced by a function $Z$ takes on the form $D(a\|b)=Z(a) - Z(b) - \langle \grad Z(b), a-b \rangle$, and therefore we have the following:
\begin{align}
D(a\|b) - D(a\|c) - D(c\|b) = \langle \grad Z(b) - \grad Z(c), c - a \rangle. \label{eq:BregTri}
\end{align}
Using these key facts, we start by bounding the excess loss at a single time point.
\begin{subequations}
\begin{align}
\tilde{\ell}_t(\htheta_t)-\tilde{\ell}_t(\theta_t)
\leq&\langle \grad \tilde{\ell}_t(\htheta_t),\htheta_t - \theta_t \rangle \label{eq:l_convex}\\
= & \langle \nabla \tilde{\ell}_t(\htheta_t),\ttheta_{t+1}-\theta_t\rangle + \langle \nabla \tilde{\ell}_t (\htheta_t),\htheta_t - \ttheta_{t+1} \rangle\nonumber\\
\leq & \frac{1}{\eta_t} \langle \grad Z(\htheta_t) - \grad Z(\ttheta_{t+1}),\ttheta_{t+1}-\theta_t \rangle \nonumber \\
&+\langle \grad \tilde{\ell}_t(\htheta_t),\htheta_t - \ttheta_{t+1}\rangle\label{eq:optimality}\\
=& \frac{1}{\eta_t} \left(D(\theta_t\|\htheta_t) - D(\theta_t\|\ttheta_{t+1}) - D(\ttheta_{t+1}\|\htheta_t)\right)\nonumber\\ 
&+\langle \grad \tilde{\ell}_t(\htheta_t),\htheta_t - \ttheta_{t+1}\rangle \label{eq:BregTriResult}
\end{align}
\end{subequations}
Equation \ref{eq:l_convex} is due the convexity of the function $\tilde{\ell}_t$, Equation \ref{eq:optimality} uses Equation \ref{eq:opt}, and Equation \ref{eq:BregTriResult} is the application of Equation \ref{eq:BregTri}.  We add and subtract necessary Bregman divergences, and bound differences separately.
\begin{align*}
\tilde{\ell}_t(\htheta_t) - \tilde{\ell}_t(\theta_t) 
\leq& \frac{1}{\eta_t} \left(D(\theta_t\|\htheta_t) - D(\theta_{t+1}\|\htheta_{t+1})\right)\\
& + \frac{1}{\eta_t}\left(D(\theta_{t+1}\|\htheta_{t+1}) - D(\tPhi_t(\theta_t,W)\|\htheta_{t+1})\right)\\
& + \frac{1}{\eta_t}\left( D(\tPhi_t(\theta_t,W)\|\htheta_{t+1}) - D(\theta_t \|\ttheta_{t+1}) \right)\\
& - \frac{1}{\eta_t} D(\ttheta_{t+1}\|\htheta_t) + \langle \grad \tilde{\ell}_t(\htheta_t),\htheta_t - \ttheta_{t+1}\rangle
\end{align*}
We will bound each of these lines separately.
\begin{subequations}
\begin{align}
D(&\theta_{t+1}\|\htheta_{t+1}) - D(\tPhi_t(\theta_t,W)\|\htheta_{t+1})\nonumber\\
= &Z(\theta_{t+1}) - Z(\tPhi_t(\theta_t,W)) + \langle \grad Z(\htheta_{t+1}),\tPhi_t(\theta_t,W) - \theta_{t+1} \rangle\label{eq:BregDef}\\
\leq & \langle \grad Z(\htheta_{t+1}) - \grad Z(\theta_{t+1}), \tPhi_t(\theta_t,W) - \theta_{t+1} \rangle\label{eq:convxZ}\\
\leq & \| \grad Z(\htheta_{t+1}) - \grad Z(\theta_{t+1})\|_2 \|\tPhi_t(\theta_t,W)-\theta_{t+1} \|_2\label{eq:CSIneq}\\
=&\| \delta (\hat{\lambda}_{t+1} - \lambda_{t+1}) \|_2 \|\log(\delta {\Phi}_t(\lambda_t,W)) - \log(\delta \lambda_{t+1}) \|_2\nonumber\\
\leq& \delta \sqrt{p} \lambda_{\max} \frac{1}{\lambda_{\min}} \|{\Phi}_t(\lambda_t,W) - \lambda_{t+1}\|_2 \label{eq:logLipschitz}
\end{align}
\end{subequations}
Equation \ref{eq:BregDef} uses the definition of the Bregman divergence, Equation \ref{eq:convxZ} the convexity of $Z$, Equation \ref{eq:CSIneq} the Cauchy-Schwarz inequality, and Equation \ref{eq:logLipschitz} uses the bounded domain of $\lambda \in [\lambda_{\min},\lambda_{\max}]^p$ with $\lambda_{\min}>0$ and the Lipschitz property of the natural logarithm on $[\lambda_{\min},\lambda_{\max}]$. 
The next term we bound by using the contractivity assumption on $\Phi_t$.
\begin{align*}
D(&\tPhi_t(\theta_t,W)\|\htheta_{t+1}) - D(\theta_t\|\ttheta_{t+1})\\
 =& D(\tPhi_t(\theta_t,W)\|\tPhi_t(\ttheta_{t+1},W)) - D(\theta_t\|\ttheta_{t+1})\leq 0
\end{align*}
To bound the final term, we use the strong convexity property of $Z(\theta)$ which implies that $D(\theta_1\|\theta_2) \geq \frac{\delta \lambda_{\min}}{2} \|\theta_1 - \theta_2 \|^2_2$ (Equation \ref{eq:BregStrong}). 
\begin{subequations}
\begin{align}
\langle\grad\tilde{\ell}_t&(\htheta_t),\htheta_t - \ttheta_{t+1} \rangle - \frac{1}{\eta_t} D(\ttheta_{t+1}\|\htheta_t)\nonumber\\
\leq& \|\grad\tilde{\ell}_t(\htheta_t)\|_2\|\htheta_t - \ttheta_{t+1}\|_2 - \frac{\delta \lambda_{\min}}{2\eta_t}\|\ttheta_{t+1}-\htheta_t\|_2^2\label{eq:CSIneq2}\\ 
\leq & \frac{\eta_t}{2\delta \lambda_{\min}} \|\grad \tilde{\ell}_t(\htheta_t)\|_2^2 + \frac{\delta \lambda_{\min}}{2 \eta_t} \|\ttheta_{t+1}-\htheta_t\|_2^2 \nonumber \\
&- \frac{\delta \lambda_{\min}}{2\eta_t}\|\ttheta_{t+1}-\htheta_t\|_2^2\label{eq:youngs}\\
\leq & \frac{\eta_t p \delta (\lambda_{\max}+x_{\max})^2}{2 \lambda_{\min}}  \label{eq:Lbound}
\end{align}
\end{subequations}
In the above, Equation \ref{eq:CSIneq2} uses the Cauchy-Schwarz inequality and Equation \ref{eq:BregStrong}, Equation \ref{eq:youngs} uses Young's inequality, and finally Equation \ref{eq:Lbound} applies Equation \ref{eq:gradLipschitz}.
Combining all the terms, we get an upper bound on the excess loss at any given time point of the following form:
\begin{align*}
 \ell_t(&\wh{\lambda}_t) - \ell_t(\lambda_t)
 \leq \frac{1}{\eta_t} \left(D(\theta_t\|\htheta_t) - D(\theta_{t+1}\|\htheta_{t+1})\right )\\
 &+ \frac{\delta \sqrt{p}{ \lambda_{\max}}}{\eta_t\lambda_{\min}}\|{\Phi}_t(\lambda_t,W) - \lambda_t\|_2
 + \frac{\eta_t p \delta (\lambda_{\max}+x_{\max})^2}{2  \lambda_{\min}}.
\end{align*}
To get the final bound, we must add these terms over the entire length of the optimization process from $t=1,...,T/\delta$.  To do this, we first show how the telescoping of the Bregman divergence terms happens.  In the following lines, we use the assumption that $\eta_t$ is positive and non-increasing in $t$ as well as the upper bound on the Bregman divergence from Equation \ref{eq:D_max}.  
\begin{align*}
\sum_{t=1}^{T/\delta} \frac{1}{\eta_t}\left( D(\theta_t\|\htheta_t) - D(\theta_{t+1}\|\htheta_{t+1})\right) 
= &\frac{1}{\eta_{1}}D(\theta_1\|\htheta_1) - \frac{1}{\eta_{T/\delta}}D(\theta_{T/\delta+1}\|\htheta_{T/\delta+1})+ \sum_{t=2}^{T/\delta} D(\theta_t\|\htheta_{t}) \left(\frac{1}{\eta_t} - \frac{1}{\eta_{t-1}}\right)\\
\leq  &\frac{1}{\eta_{1}}D(\theta_1\|\htheta_1) - \frac{1}{\eta_{T/\delta}}D(\theta_{T/\delta+1}\|\htheta_{T/\delta+1}) 
 + \frac{\delta \lambda_{\max}^2 p}{\lambda_{\min}} \left(\frac{1}{\eta_{T/\delta}}-\frac{1}{\eta_{1}}\right)\\
 \leq & \frac{\delta \lambda_{\max}^2 p }{\eta_{T/\delta}\lambda_{\min}}
\end{align*}
Using this, we combine all the terms to get the final bound.
\begin{align*}
\sum_{t=1}^{T/\delta} {\ell}_t (\wh{\lambda}_t) - {\ell}_t(\lambda_t) \leq \frac{\delta \lambda_{\max}^2 p}{\eta_{T/\delta}\lambda_{\min}}
+ \frac{p \delta (\lambda_{\max} + x_{\max})^2}{2 \lambda_{\min}} \sum_{t=1}^{T/\delta} \eta_t
+ \frac{\delta \sqrt{p}{ \lambda_{\max}}}{\lambda_{\min}} \sum_{t=1}^{T/\delta} \frac{1}{\eta_t}\|{\Phi}_t(\lambda_t,W)-\lambda_{t+1}\|_2\ 
\end{align*}
If the time horizon, $T$, is known, we choose $\eta_1=\eta_2=...=\eta_{T/\delta}$ to be a constant proportional to $\frac{1}{\sqrt{T/\delta}}$, or if $T$ is unknown, we choose $\eta_t$ to be proportional to $\frac{1}{\sqrt{t}}$.  For the former choice the regret bound becomes:
\begin{align*}
 \sqrt{\delta}\Bigg(\frac{ \lambda_{\max}^2 p}{\lambda_{\min}} + \frac{ \sqrt{p}{\lambda_{\max}}}{\lambda_{\min}} \sum_{t=1}^{T/\delta}\|{\Phi}_t(\lambda_t,W) - \lambda_{t+1} \|_2
 + \frac{p (\lambda_{\max}+x_{\max})^2}{2\lambda_{\min}}\Bigg)\sqrt{T}.
 \end{align*}
And for the later choice, we use the fact that $\sum_{t=1}^{T/\delta} \frac{1}{\sqrt{t}} \leq 1 + \int_{1}^{T/\delta} \frac{1}{\sqrt{t}}dt=2\sqrt{T/\delta}-1< 2\sqrt{T/\delta}$.  This brings the overall bound to
\begin{align*}
\sqrt{\delta}\Bigg(\frac{ \lambda_{\max}^2 p}{\lambda_{\min}}+ \frac{ \sqrt{p}{ \lambda_{\max}}}{\lambda_{\min}} \sum_{t=1}^{T/\delta}\|{\Phi}_t(\lambda_t,W) - \lambda_t\|_2 
+ \frac{p  (\lambda_{\max}+x_{\max})^2}{ \lambda_{\min}}\Bigg) \sqrt{T}.
\end{align*}
Both of these are order $\sqrt{T}$ proving the result.

\section{Proof of Lemma \ref{lem:track_W}}\label{app:track_W}
The proof is a simple inductive argument.  We start with the base scenario, at $t=1$.  Since Algorithm \ref{alg:W_Known} begins with $\hat{\lambda}_1=\bar{\mu}$, we have
$$\wh{\lambda}_1^{W_1} = \wh{\lambda}_1^{W_2} + (W_1-W_2)\boldsymbol{0} = \bar{\mu}.$$
Therefore the results hold for $t=1$, with $K_1=\boldsymbol{0}$. Now we show the inductive step.  
If we use the update form from Equation \ref{eq:update}, we can explicitly compute the difference for different values of $W$.
\begin{align*}
\wh{\lambda}_{t+1}^{W_1} - \wh{\lambda}_{t+1}^{W_2} = &(1-\eta_t)\alpha^\delta (\wh{\lambda}_t^{W_1}-\wh{\lambda}_t^{W_2}) + (W_1-W_2)y_t\\
=& (W_1-W_2)((1-\eta_t)\alpha^\delta K_t +y_t)\\
=& (W_1-W_2)K_{t+1}
\end{align*}
Here we assumed that $\wh{\lambda}_t^{W_1}=\wh{\lambda}_t^{W_2} + (W_1 - W_2) K_t$ and then proved that the next step holds true for $K_{t+1}=(1-\eta_t)\alpha^\delta K_t + y_t,$ as the Lemma states.

\section{Proof of Theorem \ref{thm:main}}\label{app:pf_main}
In order to bound the regret of this algorithm, we split the regret into two separate difference terms and bound them individually.
\begin{align*}
\sum_{t=1}^{T/\delta} \ell_t(\wh{\lambda}_t) - \sum_{t=1}^{T/\delta} \ell_t(\lambda_t)
=\sum_{t=1}^{T/\delta}\ell_t(\wh{\lambda}_t)-\sum_{t=1}^{T/\delta} \ell_t(\wh{\lambda}_t^W) + \sum_{t=1}^{T/\delta}\ell_t(\wh{\lambda}_t^W) - \sum_{t=1}^{T/\delta} \ell_t(\lambda_t)
\end{align*}
Here, $\wh{\lambda}_t$ represents the output of Algorithm \ref{alg:hawkes} at time $t$, and $\wh{\lambda}_t^W$ is the output of Algorithm \ref{alg:W_Known} had we used $W$ for any given $W$ in Algorithm \ref{alg:W_Known}.  We will show a bound which holds for all $W \in \mathcal{W}$.  The bound on the second difference follows directly from Theorem \ref{thm:W_known}.
\begin{align}
\sum_{t=1}^{T/\delta}\ell_t(\wh{\lambda}_t^W) - \sum_{t=1}^{T/\delta} \ell_t(\lambda_t) 
=C_1\left(1+\sum_{t=1}^{T/\delta}\|\tilde{\Phi}_t(\lambda_t,W)-\lambda_{t+1}\|_2\right) \sqrt{T}  \label{eq:hawkesTerm1}
\end{align}
In order to bound the first difference, we use the results of Lemma \ref{lem:track_W} to express the loss function in terms of $W$.
\begin{align*}
\ell_t(&\hat{\lambda}_t^W) = \langle \ones, \delta \wh{\lambda}_t^W\rangle - \langle x_t,\log(\delta \wh{\lambda}_t^W)\rangle\\
=& \langle \ones, \delta (\wh{\lambda}^0 + W K_t )\rangle - \langle x_t,\log(\delta (\wh{\lambda}_t^0 + W K_t ))\rangle = g_t(W)
\end{align*}
This loss function is convex in $W$ and therefore allows for searching amongst the outputs of the Algorithm \ref{alg:W_Known} for different values of $W$ at every time step, to find which $W$ would have produced the best estimate for the current data.  The important observation now is that every output, $\wh{\lambda}_t$ of Algorithm \ref{alg:hawkes} is the output of Algorithm \ref{alg:W_Known} for the specific value $\wh{W}_{t}$.  To see this, notice that if $\wh{\lambda}_t = \wh{\lambda}_t^{\wh{W}_t}$, then $\tilde{\lambda}_{t+1}$ is equivalent to the output of line 4 from Algorithm \ref{alg:W_Known} as long as $\tlambda_{t+1} \in \Int {\Lambda}$.  Then in lines 9 and 10 of Algorithm \ref{alg:hawkes} we apply the dynamics and Lemma \ref{lem:track_W} thus producing $\wh{\lambda}_{t+1}=\wh{\lambda}_{t+1}^{\wh{W}_{t+1}}$.  Since $\wh{\lambda}_1 = \bar{\mu} = \wh{\lambda}_1^{W}$ for any $W$, this shows that at any time $\wh{\lambda}_t = \wh{\lambda}_t^{\wh{W}_t}$.
\begin{align*}
\sum_{t=1}^{T/\delta}\ell_t(\wh{\lambda}_t)-\sum_{t=1}^{T/\delta} \ell_t(\wh{\lambda}_t^W)
= \sum_{t=1}^{T/\delta} g_t(\wh{W}_t) - \sum_{t=1}^{T/\delta} g_t(W)
\end{align*}
Since Algorithm \ref{alg:hawkes}, line 8, performs a gradient descent method \cite{BecTeb03} to produce estimates $\wh{W}_t$, we know that the difference is bounded as
\begin{align}
\sum_{t=1}^{T/\delta} g_t(\wh{W}_t) - \min_{W \in \mathcal{W}}\sum_{t=1}^{T/\delta} g_t(W) \leq C_2\sqrt{T/\delta} \label{eq:hawkesTerm2}.
\end{align}
Combining Equation \ref{eq:hawkesTerm1} taken with the $W = \displaystyle \argmin_{W \in \mathcal{W}}\sum_{t=1}^{T} \|\Phi(\lambda_t,W) - \lambda_{t+1}\|_2$ and Equation \ref{eq:hawkesTerm2} gives the result:
\begin{align*}
\sum_{t=1}^{T/\delta} &\ell_t (\wh{\lambda}_t) - \sum_{t=1}^{T/\delta}\ell_t(\lambda_t) 
\leq C \left(1+\min_{W\in\mathcal{W}}\sum_{t=1}^{T/\delta}\|\Phi_t(\lambda_t,W)-\lambda_{t+1}\|_2\right) \sqrt{T}.
\end{align*}

\section{Online Gradient Descent}\label{app:OGD}
As a comparison to our methods, we describe an implementation of Online Gradient Descent (OGD) that can be used to learn the network weights, $W$. In order to do this, we will take the rate, $\hat{\lambda}_t$ to be a direct function of the network estimate $\widehat{W}_t$ using the exponential influence function of Section \ref{sec:loss} and use the loss function described in Equation \ref{eq:losses1}.
\begin{align*}
\lambda_t(W) =& \bar{\mu} + \sum_{\tau=1}^{t-1} \alpha^{\delta(t-1-\tau)} W y_\tau = \bar{\mu} + W K_t \\
K_t \triangleq& \sum_{\tau =1}^{t-1} \alpha^{\delta(t-1-\tau)} y_\tau = \alpha^\delta K_{t-1} + y_{t-1}\\
g_t(W) =& \langle \delta \lambda_t(W), \ones \rangle - \langle \log(\delta \lambda_t(W)),x_t\rangle \\
\grad g_t(W) =& \delta \ones K_t^T - \diag(\lambda_t(W))^{-1} x_t K_t^T
\end{align*}
Using these values as a framework, we can derive an Online Gradient Descent algorithm for the learning the network in a Hawkes process.

\begin{algorithm}
\caption{Learning Network $W$ with Online Gradient Descent}
\label{alg:OGD}
\begin{algorithmic}[1] 
\STATE Initialize $\wh{W}_1 = W_0$, $K_1 = \zeros$
\FOR {$t=1,...,T/\delta$}
\STATE Observe $x_t$ and incur loss $g_t(W) = \langle \ones, \delta \wh{W}_tK_t \rangle - \langle x_t, \log \delta (\bar{\mu} + \wh{W}_t K_t ) \rangle$
\STATE Set $\grad g_t(W) = \delta \ones K_t \trans - \diag(\wh{\lambda}_t^{\wh{W}_t})^{-1}x_t K_t\trans$
\STATE Set 
$\wh{W}_{t+1} = \proj_{\mathcal{W}} \left(\wh{W}_{t} - \rho_t \grad g_t(\wh{W}_t)\right)$
\STATE Define $y_t \triangleq \displaystyle \sum_{\bar{\tau}_n = \delta t} e_{k_n} \alpha^{(\delta (t+1) - \tau_n)}$
\STATE Set $K_{t+1} = \alpha^\delta K_{t} + y_t$
\ENDFOR
\end{algorithmic}
\end{algorithm}

Comparing Algorithms \ref{alg:hawkes} and \ref{alg:OGD}, we can see how our proposed algorithm, is actually a generalization of OGD, in which instead of learning just the network weights and plugging them into the equation for the current rate, we are also allowed to slightly alter the value of the rate to deviate from the direct computation. Additionally, comparing the two shows how OGD is simply our algorithm with the parameter $\eta_t = 0$ for all time steps $t$.

\section{Notation Legend}\label{app:Notation}
\begin{xtabular}{p{.15\linewidth} p{.75 \linewidth} }
Variable & Meaning \\ \hline
$p$ & Number of actors in the network \\
$k_n$ & Actor involved in the $n^{th}$ event \\
$\tau_n$ & Time of the $n^{th}$ event \\
$\mu_k(\tau)$ & Continuous time rate of the $k^{th}$ actor at time $\tau$ \\
$N_{k,\tau}$ & Number of events involving actor $k$ up to and including time $\tau$ \\
$N_\tau$ & Total number of events up to and including time $\tau$ \\
$\mathcal{H}^T$ & All observed events (actors and times) up to and including time $T$\\
$\bar{\mu}_k$ & Baseline rate for actor $k$\\
$W$ & Weighted adjacency matrix for the network\\
$h(t)$ & Influence function describing how one event influences other actors over time\\
$T$ & Total sensing time of the process \\
$\delta$ & Length of time window used for discretization \\
$x_{t,k}$ & Number of times actor $k$ acts during in time window $ (\delta(t-1), \delta t] $\\
$\lambda_{t,k}$ & Discrete time rate of actor $k$ and time $\delta t$, approximating $\mu_k(\delta t)$\\
$\bar{\tau}_n$ & Discrete times related to $\tau_n$ by $\lceil \frac{\tau_n}{\delta} \rceil$\\
$L_t(\mu)$ & Negative log-likelihood of Hawkes process at time $T$ \\
$L_t^{(\delta)}$ & Discrete time approximation to $L_t(\mu)$\\
$\ell_t(\lambda)$ & Discrete time loss of $\lambda$ at time $t$\\
$\theta$ & Dual paramter to $\lambda$, defined as $\theta = \log(\delta \lambda)$\\
$\tilde{\ell}_t(\theta)$ & Instantaneous loss of $\theta$ at time $t$. $\tilde{\ell}_t(\theta) = \ell_t(\lambda)$\\
$\lambda_{\min}, \lambda_{\max}$ & Smallest and largest values $\lambda$ is allowed to take\\
$x_{\max}$ & Maximum times an actor can act per unit time\\
$\Phi_t$ & Dynamical model in the $\lambda$ space at time $t$\\
$\tilde{\Phi}_t$ & Dynamical model in the $\theta$ space at time $t$\\
$y_t$ & Instantaneous vector of actors that acted at time $\delta t$, weighted by the influence function\\
$D(\theta_1\|\theta_2)$ & Bregman divergence between $\theta_1$ and $\theta_2$\\
$\lambda_t^W$ & Instantaneous rate generated using network values $W$\\
$K_t$ & Vector used to convert estimates of Algorithm 1 generated with $W_1$ to estimates generated with $W_2$
\end{xtabular}

\end{appendices}
\bibliographystyle{IEEEbib}
\bibliography{WillettRefs,DynamicOCP,HawkesRefs}

\begin{thebibliography}{10}

\bibitem{raginsky_OCP}
M.~Raginsky, R.~Willett, C.~Horn, J.~Silva, and R.~Marcia,
\newblock ``Sequential anomaly detection in the presence of noise and limited
  feedback,''
\newblock {\em IEEE Transactions on Information Theory}, vol. 58, no. 8, pp.
  5544--5562, 2012.

\bibitem{silva:pami}
J.~Silva and R.~Willett,
\newblock ``Hypergraph-based anomaly detection in very large networks,''
\newblock {\em IEEE Transactions on Pattern Analysis and Machine Intelligence},
  vol. 31, no. 3, pp. 563--569, 2009,
\newblock
  \href{http://doi.ieeecomputersociety.org/10.1109/TPAMI.2008.232}{doi:10.1109/TPAMI.2008.232}.

\bibitem{BertozziHawkes}
A.~Stomakhin, M.~B. Short, and A.~Bertozzi,
\newblock ``Reconstruction of missing data in social networks based on temporal
  patterns of interactions,''
\newblock {\em Inverse Problems}, vol. 27, no. 11, 2011.

\bibitem{HellerHawkes}
C.~Blundell, K.~A. Heller, and J.~M. Beck,
\newblock ``Modelling reciprocating relationships with hawkes processes,''
\newblock in {\em Proc. NIPS}, 2012.

\bibitem{zhouZhaSongHawkes}
K.~Zhou, H.~Zha, and L.~Song,
\newblock ``Learning social infectivity in sparse low-rank networks using
  multi-dimensional hawkes processes,''
\newblock in {\em Proceedings of the 16th International Conference on
  Artificial Intelligence and Statistics (AISTATS)}, 2013.

\bibitem{brown2004multiple}
E.~N. Brown, R.~E. Kass, and P.~P. Mitra,
\newblock ``Multiple neural spike train data analysis: state-of-the-art and
  future challenges,''
\newblock {\em Nature Neuroscience}, vol. 7, no. 5, pp. 456--461, 2004.

\bibitem{colemanConvexPoint}
T.~P. Coleman and S.~Sarma,
\newblock ``Using convex optimization for nonparametric statistical analysis of
  point processes,''
\newblock in {\em Proc. ISIT}, 2007.

\bibitem{SmithBrownStateSpace}
A.~C. Smith and E.~N. Brown,
\newblock ``Estimating a state-space model from point process observations,''
\newblock {\em Neural Computation}, vol. 15, pp. 965--991, 2003.

\bibitem{HinneHeskes2012}
M.~Hinne, T.~Heskes, and M.~A.~J. {van Gerven},
\newblock ``Bayesian inference of whole-brain networks,''
\newblock {\em arXiv:1202.1696 [q-bio.NC]}, 2012.

\bibitem{DingSchroeder2011}
M.~Ding, {CE} Schroeder, and X.~Wen,
\newblock ``Analyzing coherent brain networks with {G}ranger causality,''
\newblock in {\em Conf. Proc. IEEE Eng. Med. Biol. Soc.}, 2011, pp. 5916--8.

\bibitem{spikesPillow}
J.~W. Pillow, J.~Shlens, L.~Paninski, A.~Sher, A.~M. Litke, E.~J. Chichilnisky,
  and E.~P. Simoncelli,
\newblock ``Spatio-temporal correlations and visual signalling in a complete
  neuronal population,''
\newblock {\em Nature}, vol. 454, pp. 995--999, 2008.

\bibitem{spikesMasud}
M.~S. Masud and R.~Borisyuk,
\newblock ``Statistical technique for analysing functional connectivity of
  multiple spike trains,''
\newblock {\em Journal of Neuroscience Methods}, vol. 196, no. 1, pp. 201--219,
  2011.

\bibitem{ait2010modeling}
Y.~A{\"\i}t-Sahalia, J.~Cacho-Diaz, and R.~J.~A. Laeven,
\newblock ``Modeling financial contagion using mutually exciting jump
  processes,''
\newblock Tech. {R}ep., National Bureau of Economic Research, 2010.

\bibitem{cameron2013regression}
A~Colin Cameron and Pravin~K Trivedi,
\newblock {\em Regression analysis of count data}, vol.~53,
\newblock Cambridge university press, 2013.

\bibitem{engle2001garch}
Robert Engle,
\newblock ``Garch 101: The use of arch/garch models in applied econometrics,''
\newblock {\em Journal of economic perspectives}, pp. 157--168, 2001.

\bibitem{kuperman2001small}
M.~Kuperman and G.~Abramson,
\newblock ``Small world effect in an epidemiological model,''
\newblock {\em Physical Review Letters}, vol. 86, no. 13, pp. 2909, 2001.

\bibitem{hawkesEarthquake}
D.~Vere-Jones and T.~Ozaki,
\newblock ``Some examples of statistical estimation applied to earthquake
  data,''
\newblock {\em Ann. Inst. Statist. Math.}, vol. 34, pp. 189--207, 1982.

\bibitem{ogata1999seismicity}
Y.~Ogata,
\newblock ``Seismicity analysis through point-process modeling: A review,''
\newblock {\em Pure and Applied Geophysics}, vol. 155, no. 2-4, pp. 471--507,
  1999.

\bibitem{Schoenberg13}
F.P. Schoenberg,
\newblock ``Facilitated estimation of etas,''
\newblock {\em Bulletin of the Seismological Society of America}, vol. 103, pp.
  601 -- 605, 2013.

\bibitem{hawkes1}
A.~G. Hawkes,
\newblock ``Point spectra of some self-exciting and mutually-exciting point
  processes,''
\newblock {\em Journal of the Royal Statistical Society. Series B
  (Methodological)}, vol. 58, pp. 83--90, 1971.

\bibitem{hawkes2}
A.~G. Hawkes,
\newblock ``Point spectra of some mutually-exciting point processes,''
\newblock {\em Journal of the Royal Statistical Society. Series B
  (Methodological)}, vol. 33, no. 3, pp. 438--443, 1971.

\bibitem{PointProcesses}
D.~J. Daley and D.~Vere-Jones,
\newblock {\em An introduction to the theory of point processes, Vol. I:
  Probability and its Applications},
\newblock Springer-Verlag, New York, second edition, 2003.

\bibitem{DMD_journal}
E.~C. Hall and R.~M. Willett,
\newblock ``Online convex optimization in dynamic environments,''
\newblock {\em IEEE Journal of Selected Topics in Signal Processing}, vol. 9,
  2015.

\bibitem{dynamicMirrorDescent}
E.~C. Hall and R.~M. Willett,
\newblock ``Dynamical models and tracking regret in online convex
  programming,''
\newblock in {\em Proc. International Conference on Machine Learning (ICML)},
  2013,
\newblock \href{http://arxiv.org/abs/1301.1254}{arXiv.org:1301.1254}.

\bibitem{SimmaJordan10}
A.~Simma and M.I. Jordan,
\newblock ``Modeling events with cascades of poisson processes,''
\newblock {\em Proceedings of the Twenty-Sixth Conference of Uncertainty in
  Artificial Intelligence (UAI2010)}, 2010.

\bibitem{Mohler13}
G.~Mohler,
\newblock ``Modeling and estimation of multi-source clustering in crime and
  security data,''
\newblock {\em Annals of Applied Statistics}, vol. 7, pp. 1525 -- 1539, 2013.

\bibitem{SornetteUtkin09}
D.~Sornette and S.~Utkin,
\newblock ``Limits of declustering methods for disentangling exogenous from
  endogenous events in time series with foreshocks, main shocks and
  aftershocks,''
\newblock {\em Physical Review E}, 2009.

\bibitem{VeenSchoenberg08}
A.~Veen and F.P. Schoenberg,
\newblock ``Estimation of space-time branching process models in seismology
  using an em-type algorithm,''
\newblock {\em Journal of the American Statistical Association}, 2008.

\bibitem{ryanAdamsHawkes}
Scott~W. Linderman and Ryan~P. Adams,
\newblock ``Discovering latent network structure in point process data,''
\newblock arXiv:1402.0914, 2014.

\bibitem{LindermanAdams15}
S.W. Linderman and R.P. Adams,
\newblock ``Scalable bayesian inference for excitatory point process
  networks,''
\newblock arXiv: 1507.03228v1, 2015.

\bibitem{patriciaHawkes}
P.~{Reynaud-Bouret} and S.~Schbath,
\newblock ``Adaptive estimation for {Hawkes} processes; application to genome
  analysis,''
\newblock {\em Annals of Statistics}, vol. 38, no. 5, pp. 2781--2822, 2010.

\bibitem{hansen2012lasso}
N.~R. Hansen, P.~Reynaud-Bouret, and V.~Rivoirard,
\newblock ``{LASSO} and probabilistic inequalities for multivariate point
  processes,''
\newblock {\em arXiv preprint arXiv:1208.0570}, 2012.

\bibitem{soloHawkesICASSP13}
S.~A. Pasha and V.~Solo,
\newblock ``Hawkes-{L}aguerre reduced rank model for point processes,''
\newblock in {\em ICASSP}, 2013.

\bibitem{expHawkesSimulation}
Angelos Dassios and Hongbiao Zhao,
\newblock ``Exact simulation of hawkes process with exponentially decaying
  intensity,''
\newblock {\em Electronic Communications in Probability}, vol. 18, pp. no. 62,
  1--13, 2013.

\bibitem{Zin03}
M.~Zinkevich,
\newblock ``Online convex programming and generalized infinitesimal gradient
  descent,''
\newblock in {\em Proc. Int. Conf. on Machine Learning (ICML)}, 2003, pp.
  928--936.

\bibitem{BecTeb03}
A.~Beck and M.~Teboulle,
\newblock ``Mirror descent and nonlinear projected subgradient methods for
  convex programming,''
\newblock {\em Operations Research Letters}, vol. 31, pp. 167--175, 2003.

\bibitem{COMD}
J.~Duchi, S.~Shalev-Shwartz, Y.~Singer, and A.~Tewari,
\newblock ``Composite objective mirror descent,''
\newblock in {\em Conf. on Learning Theory (COLT)}, 2010.

\bibitem{Rak12}
A.~Rakhlin and K.~Sridharan,
\newblock ``Online learning with predictable sequences,''
\newblock arXiv:1208.3728, 2012.

\bibitem{Memetracker}
J.~Leskovec, L.~Backstrom, and J.~Kleinberg,
\newblock ``Meme-tracking and the dynamics of the news cycle,''
\newblock {\em ACM SIGKDD International Conference on Knowledge Discovery and
  Data Mining (KDD)}, 2009.

\bibitem{sparsa}
S.~Wright, R.~Nowak, and M.~Figueiredo,
\newblock ``Sparse reconstruction by separable approximation,''
\newblock {\em IEEE Transactions Signal Processing}, vol. 57, pp. 2479--2493,
  2009.

\end{thebibliography}

\end{document}